%% file: main.tex
\documentclass[twocolumn]{article}

\usepackage{graphicx}
\usepackage[top=1.1in,bottom=1.1in,left=1in,right=1in]{geometry}	
\usepackage[export]{adjustbox} %
\usepackage{array}
\usepackage{makecell}
\usepackage{booktabs}
\usepackage[dvipsnames, table]{xcolor}
\usepackage{xspace}
\usepackage{pifont}

\usepackage[accsupp]{axessibility}  %
\usepackage{fancyhdr}
\usepackage{multirow}
\usepackage{graphicx}   %
\usepackage{array}
\usepackage{caption}
\usepackage{subcaption}

\usepackage{orcidlink}

\definecolor{best}{rgb}{1, 0.7, 0.7}
\definecolor{second}{rgb}{1, 0.85, 0.7}
\usepackage{natbib}

\usepackage[affil-it]{authblk} %

\usepackage{hyperref}
\usepackage[capitalize]{cleveref}

\title{3DTV: A Feedforward Interpolation Network for Real-Time View Synthesis}

\author{Stefan Schulz}
\author{Fernando Edelstein}
\author{Hannah Dröge}
\author{Matthias B. Hullin}
\author{Markus Plack}
\affil{University of Bonn \\ 
\texttt{\{sschulz, edelstein, droege, hullin, mplack\}@cs.uni-bonn.de}}

\date{} %

\begin{document}

\twocolumn[{%
  \begin{@twocolumnfalse}
    \vspace{-0.5in}
    \maketitle
    
    \begin{center}
      \centering
      \includegraphics[width=\textwidth]{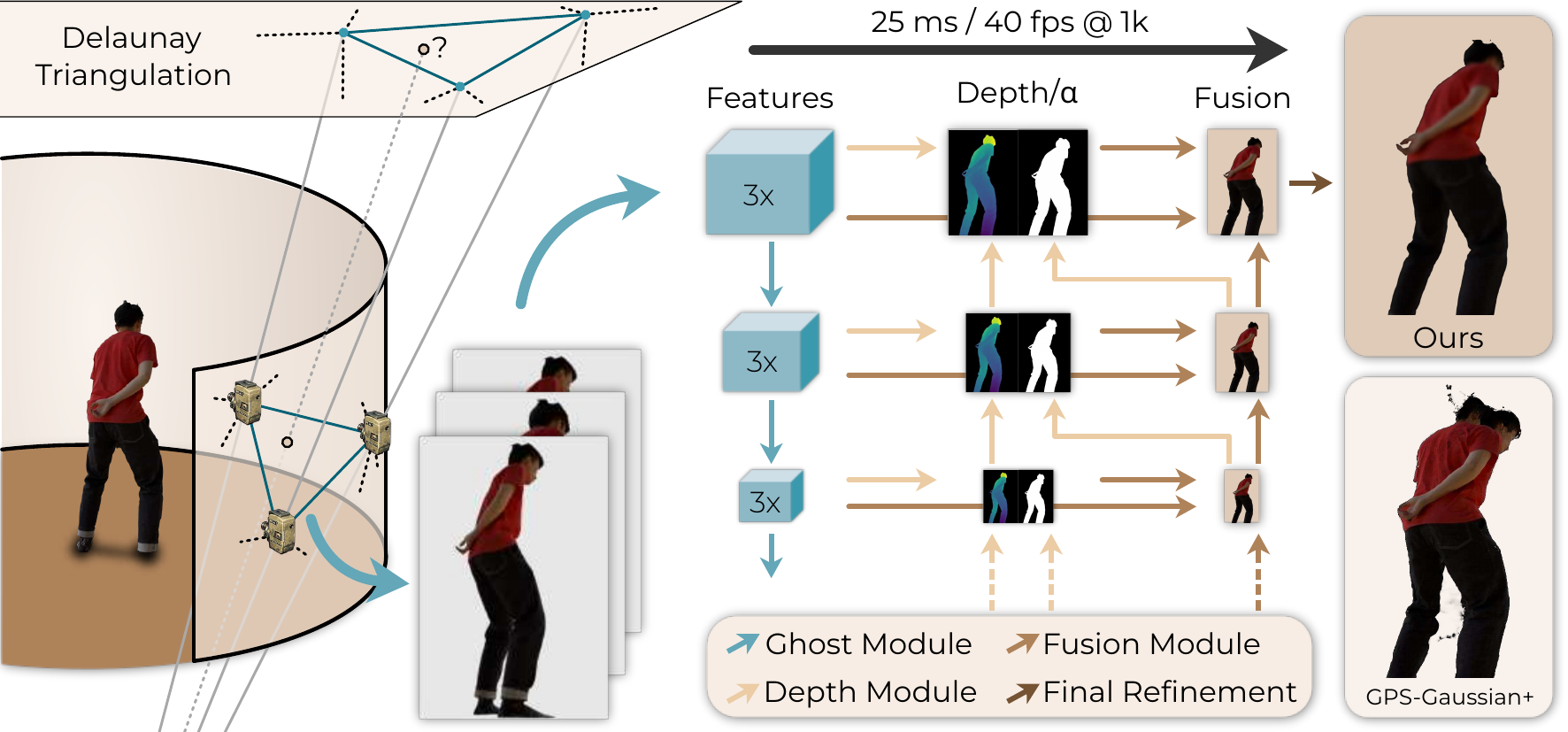}
      \captionof{figure}{
        We present 3DTV, a real-time method for novel view synthesis from sparse cameras. 
  It combines geometric view selection via Delaunay triangulation (left) with depth-guided feedforward synthesis and coarse-to-fine refinement (center), enabling novel views from only three input cameras at 40 FPS (1k resolution) without per-scene retraining. 
  Compared to recent sparse-view synthesis methods such as GPS-Gaus\-sian+~\cite{zhou2024gpsplus}, our approach is more stable and reduces artifacts (right).
      }
      \label{fig:teaser}
    \end{center}
    \vspace{1.5em} 
  \end{@twocolumnfalse}
}]

\input{sec/0_abstract}    
\input{sec/1_intro}
\input{sec/2_related}

\input{sec/3_method}

\input{sec/4_experiments}
\input{sec/5_conclusion}
\input{sec/6_acknowledgements}

\bibliographystyle{unsrt}
\bibliography{bibliography}

\clearpage

\appendix

\input{suppl/0_intro}

\input{suppl/1_triangulation}

\input{suppl/2_training_details}

\input{suppl/3_datasets}

\input{suppl/4_comparisons_failure}

\input{suppl/5_visual_ours}

\input{suppl/6_tables}

\end{document}

%% file: sec/0_abstract.tex
\begin{abstract}
    Real-time free-viewpoint rendering requires balancing multi-camera redundancy with the latency constraints of interactive applications.
    We address this challenge by combining lightweight geometry with learning and propose 3DTV, a feedforward network for real-time sparse-view interpolation. %
    A Delaunay-based triplet selection ensures angular coverage for each target view.
    Building on this, we introduce a pose-aware depth module that estimates a coarse-to-fine depth pyramid, enabling efficient feature reprojection and occlusion-aware blending.
    Unlike methods that require scene-specific optimization, 3DTV runs feedforward without retraining, making it practical for AR/VR, telepresence, and interactive applications.

    Our experiments on challenging multi-view video datasets demonstrate that 3DTV consistently achieves a strong balance of quality and efficiency, outperforming recent
    real-time novel-view baselines.
    Crucially, 3DTV avoids explicit proxies, enabling robust rendering across diverse scenes. 
    This makes it a practical solution for low-latency multi-view streaming and interactive rendering.
    
    \vspace{0.5em}

    \vspace{0.5em}
    \noindent \textbf{Project Page:} \\
    \texttt{\href{https://stefanmschulz.github.io/3DTV_webpage/}{https://stefanmschulz.github.io/3DTV\_webpage/}}
\end{abstract}

%% file: sec/1_intro.tex
\section{Introduction}
\label{sec:intro}

Real-time streaming and rendering multi-view video content poses a fundamental challenge: the raw capture produces massive amounts of data, while only a small subset of views are truly necessary for synthesizing novel viewpoints. This redundancy creates a tension between fidelity and efficiency, particularly in applications such as AR/VR, telepresence, and interactive video editing, where low-latency rendering is critical.

While dense view data can be well exploited by recent advances in novel view synthesis (NVS), which achieve striking photorealism through neural radiance fields, Gaussian splatting, or diffusion-based generation ~\cite{mildenhall2021nerf,muller2022instant,kerbl20233d,li2023validvariablelengthinputdiffusion, zhou2023sparsefusion},
these approaches typically incur heavy computational costs, require per-scene retraining, and fail to meet the demands of real-time or streaming scenarios.
In practice, existing approaches trade off latency, capture density, scene complexity (\textit{e.g.} multi-object environments) and processing requirements as summarized in \cref{tab:capabilities-limitations}.
In light of these trade-offs, classical image-based rendering methods demonstrate that not all views are equally valuable: selecting a geometrically meaningful subset can suffice for faithful interpolation.

Inspired by the seminal work of Chen and Williams on view interpolation~\cite{chen1993view}, we revisit this principle with modern learning-based techniques.
Specifically, we propose a sparse-view interpolation framework based on Delaunay triangulation of camera positions,
ensuring that each novel viewpoint is synthesized from a geometrically consistent triplet of source cameras.
This geometric conditioning reduces redundancy and yields balanced angular coverage, addressing one of the central limitations of heuristic nearest-neighbor selection.

\input{tables/capabilities}

Yet, novel view interpolation remains difficult in practice. Correspondence matching and occlusion handling are longstanding challenges, especially under wide baselines and complex scene geometry. We address these issues by adapting advances from the frame interpolation literature, replacing motion-centric flow reasoning with geometry-aware depth estimation.
Our network employs a coarse-to-fine pyramid architecture that progressively refines depth hypotheses across scales, enabling robust feature projection with significantly reduced computational overhead.

Together, these contributions yield a feedforward architecture capable of real-time novel view synthesis from only a small set of input cameras, without scene-specific retraining. By combining geometric view selection, depth-guided feature projection, and efficient hierarchical refinement, our approach bridges classical principles and modern neural rendering, advancing the goal of real-time free-viewpoint video synthesis.

\paragraph{{\bfseries\upshape Contributions.}}
We present 3DTV, a \textbf{3}-camera \textbf{D}elaunay \textbf{T}riangulation-based \textbf{V}iew interpolation framework
that combines principled geometric view selection with efficient depth-guided synthesis.
Our method achieves high-quality results under real-time constraints while requiring no per-scene retraining.
In summary, our contributions are as follows:
\begin{itemize}
    \item A Delaunay triangulation-based view selection strategy that identifies geometrically consistent camera triplets, enabling sparse-view interpolation from only three input cameras.
    \item A coarse-to-fine pyramid depth estimation and fusion architecture enabling efficient geometry-aware feature projection for real-time view synthesis.
    \item Extensive experiments demonstrating strong generalization to real-world data and competitive performance in both quality and runtime.
\end{itemize}
Code and trained models will be publicly released upon acceptance.

%% file: tables/capabilities.tex
\newcommand{\cmarkk}{\textcolor{ForestGreen}{\ding{52}}}%
\newcommand{\xmarkk}{\textcolor{BrickRed}{\ding{56}}}
\newcommand{\cmarkbraces}{\textcolor{GreenYellow}{(\ding{52})}}

\newcommand{\goodcameracount}[1]{\textcolor{ForestGreen}{\textbf{#1}}\hspace{12pt}}
\newcommand{\okaycameracount}[1]{\textcolor{GreenYellow}{\textbf{#1}}\hspace{12pt}}
\newcommand{\badcameracount}[1]{\textcolor{BrickRed}{\textbf{#1}}\hspace{12pt}}

\begin{table*}[htb]
\centering
\caption{Method capabilities and limitations overview.}
\label{tab:capabilities-limitations}
\scalebox{0.8}{
\begin{tabular}{@{}lr @{\hspace{5pt}} c @{\hspace{5pt}} c @{\hspace{5pt}} r >{\centering\arraybackslash}p{2.0cm} >{\centering\arraybackslash}p{2.0cm}@{}}
\toprule

&  &  & & & \multicolumn{2}{c}{Single GPU Reconstruction} \\

Method & Venue & Real-time & Multi-Object & \# Cameras & online & offline \\

\midrule

Nerfacto-big~\cite{nerfstudio} & SIG '23 & \xmarkk & \cmarkk & \badcameracount{$>$30} & \xmarkk & \xmarkk \\

Splatfacto-big~\cite{nerfstudio} & '24 & \xmarkk & \cmarkk & \badcameracount{$>$30} & \xmarkk & \xmarkk \\

RIFTCast~ \cite{zingsheim2025riftcast} & ACM MM '25 & \cmarkk & \cmarkk & \okaycameracount{$>$10} & \xmarkk & \cmarkk \\

ENeRF~\cite{lin2022efficient} & SA '22 & \cmarkbraces & \cmarkk & \goodcameracount{$>$2} & \cmarkk & \cmarkk \\

GPS-Gaussian~ \cite{zheng2024gpsgaussian} & CVPR '24 & \cmarkbraces & \cmarkbraces & \goodcameracount{2} & \cmarkk & \cmarkk \\

GPS-Gaussian+~ \cite{zhou2024gpsplus} & T-PAMI '25 & \cmarkbraces & \cmarkbraces & \goodcameracount{2} & \cmarkk & \cmarkk \\

FrugalNeRF~\cite{lin2024frugalnerf} & CVPR '25 & \xmarkk & \cmarkbraces & \goodcameracount{3} & \xmarkk & \xmarkk \\

\midrule

Ours & '26 & \cmarkk & \cmarkk & \goodcameracount{3} & \cmarkk & \cmarkk \\

\bottomrule
\end{tabular}
}
\end{table*}

%% file: sec/2_related.tex
\section{Related Work}
\label{sec:related}

\subsection{Frame Interpolation}

Frame interpolation is the task of generating intermediate frames from surrounding frames to ensure smooth and coherent transitions within a (video) sequence. Early classical approaches typically relied on explicit motion estimation as optical flow \cite{mahajan2009moving, sadek2012frame, raket2012motion} or phase-shifts \cite{meyer2015phase}
to guide pixel interpolation.
With the advent of data-driven techniques, convolutional neural networks (CNNs) became widely used for this task~\cite{long2016learning, jiang2018super, niklaus2017video, zhang2020flexible, park2020bmbc}. 
Several studies have addressed  large temporal gaps or large motions between consecutive frames  \cite{argaw2022longtermvideoframeinterpolation, reda2022filmframeinterpolationlarge, lu2022videoframeinterpolationtransformer}, e.g. by applying transformer architectures~\cite{lu2022videoframeinterpolationtransformer} or incorporating feature correlation into the loss function~\cite{reda2022filmframeinterpolationlarge}. 
Other approaches focused on improving spatial awareness and reducing computational cost~\cite{han2024motionawarevideoframeinterpolation}, while further research addressed the challenges such as non-uniform motion~\cite{seo2024bimvfidirectionalmotionfieldguided}, or lighting changes and motion blur using phase-based, data-driven methods~\cite{meyer2018phasenet}.
Recently, diffusion-based approaches have emerged as a powerful alternative, achieving high-fidelity results through probabilistic hierarchical sampling~\cite{hai2025hierarchicalflowdiffusionefficient, zhang2025eden, chavez2025timeadaptivevideoframeinterpolation}. Such methods include flow-guided diffusion~\cite{hai2025hierarchicalflowdiffusionefficient}, diffusion transformers for large motions~\cite{zhang2025eden}, and architectures specifically designed for animation videos~\cite{chavez2025timeadaptivevideoframeinterpolation}. {Many of these works \cite{sadek2012frame, raket2012motion, jiang2018super, bao2019depth, reda2022filmframeinterpolationlarge} rely on optical flow information to model temporal progression.}

Free-viewpoint interpolation can be interpreted as a form of frame interpolation. Inspired by Reda et al.~\cite{reda2022filmframeinterpolationlarge}, we replace central optical flow with depth estimation in the target camera to account for viewpoint changes rather than temporal motion.

\subsection{Multi-View Stereo}
Multi-view stereo (MVS) estimates dense 3D geometry from calibrated images by enforcing photo-consistency across views.
Traditional methods explored volumetric~\cite{kutulakos2000theory}, point-based~\cite{furukawa2009accurate}, and depth map-based approaches~\cite{campbell2008using,tola2012efficient}.
Recent learning-based approaches instead infer depth or 3D structure directly from images using learned priors.
MVSNet~\cite{yao2018mvsnet} constructs cost volumes from deep features and regularizes them with 3D CNNs.
To reduce memory consumption, follow-up work proposed a recurrent regularization scheme over learned features \cite{yao2019recurrent} or coarse-to-fine cost volume approaches \cite{gu2020cascadecostvolumehighresolution, yang2020cost, cheng2020deep}.
Building on the success of vision transformers in a wide range of computer vision tasks \cite{carion2020end, dosovitskiy2020image, han2022survey}, transformer-based architectures have recently been introduced to MVS, further enhancing reconstruction robustness even in textureless regions \cite{ding2022transmvsnet, liao2022wtmvsnetwindowbasedtransformersmultiview, zhu2021multiviewstereotransformer, wang2022mvsterepipolartransformerefficient, wang2024ctmvsnetefficientmultiviewstereo}.
In particular, TransMVSNet \cite{ding2022transmvsnet} was the first to use attention to capture long-range intra- and inter-view feature dependencies, followed by others that additionally incorporated epipolar constraints \cite{liao2022wtmvsnetwindowbasedtransformersmultiview, wang2022mvsterepipolartransformerefficient, wang2025etv}. Robustness was improved with multi-stage attention \cite{wang2024ctmvsnetefficientmultiviewstereo}, while non-Lambertian surfaces were addressed using a reference view synthesis loss \cite{chang2022rc}.

While transformer-based MVS focuses on long-range feature evaluation during reconstruction, our method instead uses triangulation-driven view selection with a lightweight, depth-oriented network to achieve real-time interpolation.

\subsection{Novel View Synthesis}

Novel view synthesis (NVS) focuses on generating realistic images from unseen camera viewpoints. Recently, Neural Radiance Fields (NeRF) \cite{mildenhall2021nerf}, which represent a scene with neural implicit functions, have become the dominant approach in this field. Many extensions \cite{bian2023nope, lin2024frugalnerf, chen2022hallucinated, fridovich2022plenoxels, martin2021nerf} have been proposed to improve reconstruction quality, generalization, and efficiency. Follow-up works such as Instant-NGP \cite{muller2022instant} significantly accelerate training and inference, while Kerbl \textit{et al.} \cite{kerbl20233d} introduce 3D Gaussian Splatting, which represents the scene with 3D Gaussians and further improves real-time rendering capabilities. Following Gaussian-based methods work on improvement for sparse inputs  \cite{jang2025comapgs, li2024dngaussian, paliwal2024coherentgs, chen2025dense, fan2024instantsplat, zhou2024gpsplus}, single input \cite{szymanowicz2024splatter_image} and  speed \cite{fan2024instantsplat, hanson2025speedy, navaneet2024compgs, zheng2024gpsgaussian}.
In parallel, diffusion models, first developed for image generation \cite{ho2020denoising, ho2022cascaded}, have recently been applied to novel view synthesis \cite{guizilini2025zeroshotnovelviewdepth, szymanowicz2025bolt3d}, from single \cite{watson2022novel, elata2024novel} or multiple views \cite{li2023validvariablelengthinputdiffusion, zhou2023sparsefusion}.
A related line of research targets real-time rendering/NVS without scene-specific retraining. 
PixelNeRF~\cite{yu2021pixelnerf} and IBRNet~\cite{wang2021ibrnet} infer radiance fields directly from input views, but remain slower than real-time.  FWD~\cite{cao2022fwdrealtimenovelview} achieves real-time rendering by combining forward warping with depth cues, while Guo \textit{et al.} ~\cite{guo2022fast} propose a category-agnostic method for real-time NVS. In the human NVS domain, GPS-Gaussian~\cite{zheng2024gpsgaussian} and GPSGaussian+~\cite{zhou2024gpsplus} enable real-time view synthesis without fine-tuning, whereas Snap-Snap~\cite{lu2025snapsnaptakingimagesreconstruct} achieves high-quality reconstructions but lacks real-time capability.

Our method combines principled geometric view selection with a lightweight depth-guided feedforward network, achieving real-time novel view synthesis without scene-specific retraining or optimization.

%% file: sec/3_method.tex
\section{Method}
\label{sec:method}

\begin{figure*}[tb]
  \centering
  \includegraphics[width=0.99\linewidth]{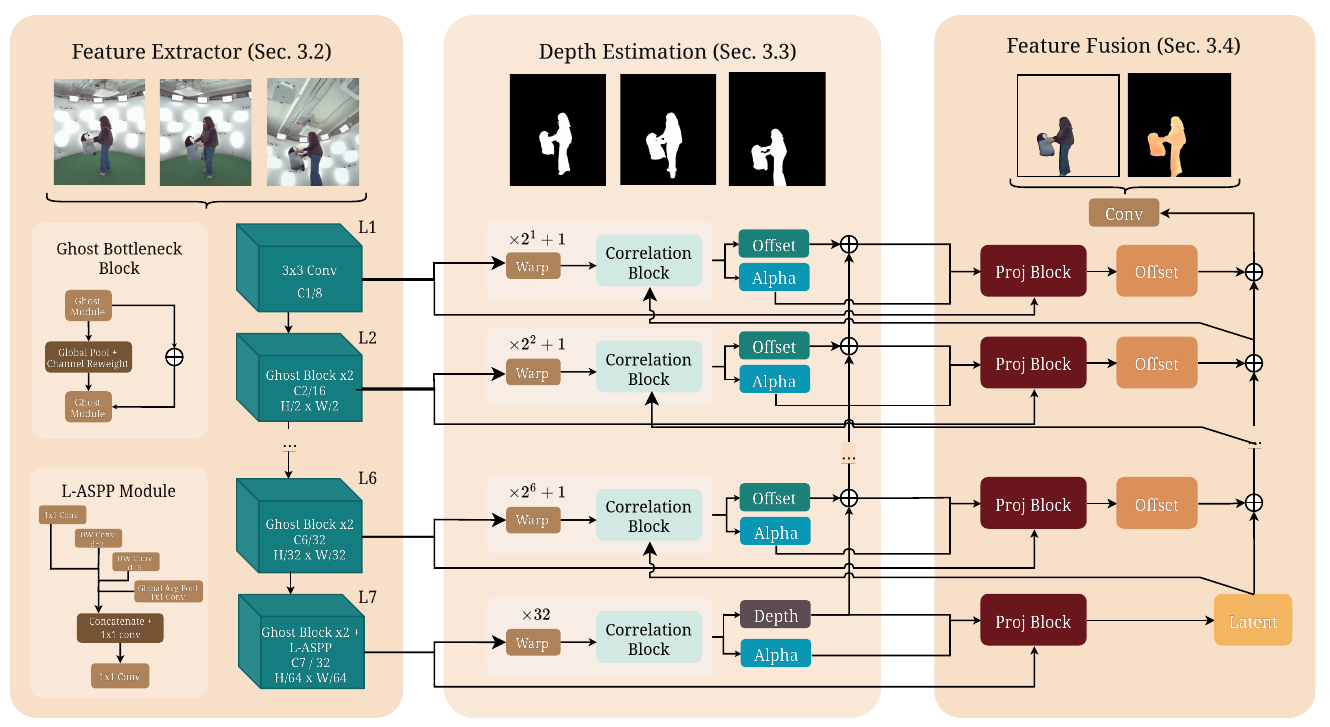}
  \caption{
  Overview of our real-time view interpolation framework.
  A lightweight Ghost-based backbone extracts multi-scale feature pyramids from the input images (left).
  A coarse-to-fine module then estimates depth and alpha for the target view using grouped correlations and residual refinement (center).
  Finally, a hierarchical fusion decoder aggregates projected features and synthesizes the novel RGB image (right).
  }
  \label{fig:network-overview}
\end{figure*}

We address bounded novel-view interpolation by combining geometric view selection with a lightweight depth-guided network.
Our pipeline first selects supporting cameras via Delaunay triangulation (\cref{sec:triangulation}), ensuring balanced angular coverage for each query view.
For each image, feature pyramids are extracted using an efficient network architecture (\cref{sec:architecture-extraction}).
From those, we regress depth and alpha for the target view in a coarse to fine fashion via a residual update network (\cref{sec:architecture-depth}).
The fusion network combines the features through differentiable projection, providing additional information for the upper depth estimation layers in a feedback loop, and synthesizes the final images (\cref{sec:architecture-projection}).

Each input view $I_i,i\in\{1,2,3\}$ is accompanied by a foreground mask $M_i$, which provides coarse subject segmentation.
During synthesis, the network predicts the target camera image $I_4$ as well as per-pixel latent features $Z^l$, depth $D^l$ and alpha maps $A^l$ at each pyramid level $l \in \{0,\dots,6\}$.

\subsection{Viewpoint Selection via Projected Delaunay Triangulation}
\label{sec:triangulation}

To synthesize a query viewpoint $\mathbf{q}$, we select a sparse set of three supporting input cameras. Standard $k$-Nearest Neighbor
selection in $\mathbb{R}^3$ often yields poorly conditioned configurations. We instead propose a projection-based strategy leveraging 2D Delaunay triangulation to ensure the query is spatially bracketed by its neighbors.
For inward-facing camera setups, we fit a cylinder to the camera centers $\{\mathbf{p}_i\}$, defined by axis $\mathbf{a}$, base $\mathbf{o}$, radius $r$, and height $h$. We perform a two-stage projection:
\begin{enumerate}
    \item \textbf{Radial Normalization:} Centers are projected onto the cylinder surface to remove depth bias:
    \begin{equation}
    \label{eq:cyl_proj}
    \begin{split}
        &\mathbf{p}_i^{\mathrm{cyl}} = \mathbf{o} + \langle \mathbf{p}_i - \mathbf{o}, \mathbf{a} \rangle \mathbf{a} + r \frac{\mathbf{v}_{i\perp}}{\|\mathbf{v}_{i\perp}\|}, \\
        &\text{where } \mathbf{v}_{i\perp} = (\mathbf{p}_i - \mathbf{o}) - \langle \mathbf{p}_i - \mathbf{o}, \mathbf{a} \rangle \mathbf{a}
    \end{split}
    \end{equation}
    \item \textbf{Perspective Mapping:} Points are mapped from 
    the origin $\mathbf{o}$
    onto a projection plane $\mathcal{P}$ at height $h$:
    \begin{equation}
    \label{eq:ray_plane}
        \mathbf{u}_i = \mathbf{o} + \frac{h}{\langle \mathbf{p}_i^{\mathrm{cyl}} - \mathbf{o}, \mathbf{a} \rangle}(\mathbf{p}_i^{\mathrm{cyl}} - \mathbf{o})
    \end{equation}
\end{enumerate}
We compute the Delaunay triangulation $\mathcal{T}$ over $\{\mathbf{u}_i\}$. For any query $\mathbf{q}$, we identify the enclosing triangle $\tau \in \mathcal{T}$ via the Müller-Trumbore ray intersection algorithm \cite{10.1145/1198555.1198746}, ensuring the selected triplet
$\tau$
provides a complete angular basis for interpolation.
Additional details and visualizations are provided in \cref{sec:triangulation}.

\subsection{Efficient Feature Extraction Backbone}
\label{sec:architecture-extraction}

To satisfy the computational constraints of real-time synthesis, we design a lightweight hierarchical backbone that extracts semantically meaningful features at a low channel count, thereby reducing the projection overhead described in \cref{sec:architecture-depth}. The architecture follows the efficiency principles of GhostNet~\cite{9157333} and GhostNetV2~\cite{10.5555/3600270.3600994}. Each block is built upon a Ghost module, where a subset of intrinsic feature maps is produced via standard convolution and the remaining channels are generated through inexpensive depthwise operations. This strategy mitigates redundancy in convolutional feature generation while preserving representational capacity.

For each source camera $i$, the backbone produces a seven-level feature pyramid $\{\mathcal{F}_i^l\}_{l=0}^{6}$, where $l=0$ denotes the highest spatial resolution and each subsequent level reduces the resolution by a factor of two. Every stage consists of a strided Ghost bottleneck for spatial downsampling followed by a second bottleneck for feature refinement. Within each bottleneck, we incorporate a lightweight channel attention mechanism based on global average pooling and channel-wise reweighting to enhance global context modeling at negligible computational cost. Residual shortcut connections are employed whenever the spatial resolution and channel dimensionality permit, ensuring stable optimization.

To compensate for the progressive loss of spatial detail induced by repeated downsampling, we append a lightweight Atrous Spatial Pyramid Pooling (L-ASPP) module at the deepest level. Inspired by MobileNetV3~\cite{howard2019searchingmobilenetv3}, and ultimately rooted in the atrous spatial pyramid design of DeepLab~\cite{chen2016semanticimagesegmentationdeep}, the module aggregates multi-scale context through a $1 \times 1$ convolution branch, multiple depthwise separable atrous convolution branches with distinct dilation rates, and a global average pooling branch, followed by feature fusion. This design enriches the coarsest representation with multi-scale contextual information while maintaining a low computational footprint.

\begin{figure*}[tb]
    \centering
    \includegraphics[width=0.99\linewidth]{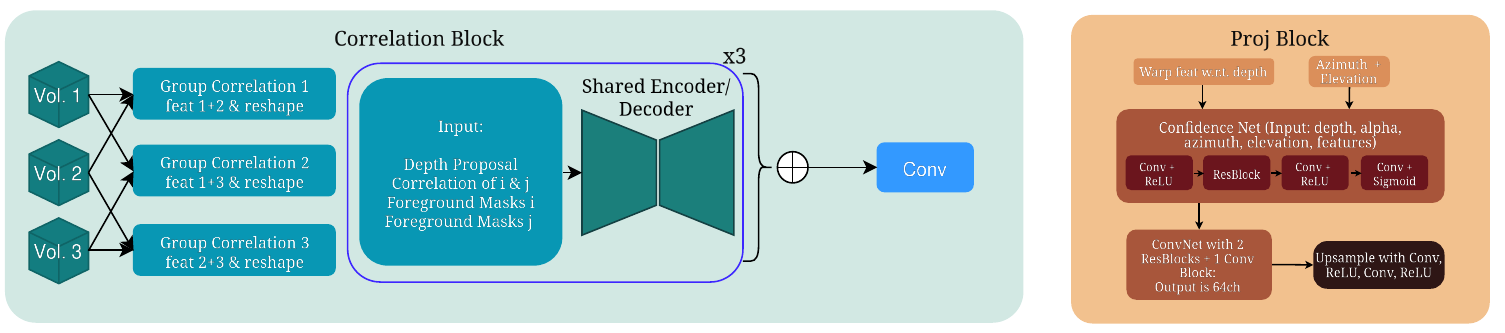}
    \caption{
    Details on the Correlation Block (left) and Proj Block (right) from \cref{fig:network-overview}.
    The correlation block computes pairwise group correlations from depth-projected features and masks, processes them with depth proposals in a shared encoder–decoder, and refines the fused latents with the upsampled depth.
    The projection block warps features using the current depth, predicts per-view weights, and aggregates them with a ConvNet into a latent representation.
    }
    \label{fig:network-details}
\end{figure*}

\subsection{Depth Estimation and Refinement}
\label{sec:architecture-depth}

Given source-view features $\{\mathcal{F}_i^l\}_{i=1}^{3}$, we estimate a dense depth map for the novel target view $I_4$ using a plane-sweep stereo formulation. At the coarsest level ($l=6$), we initialize 32 depth hypotheses $\mathcal{D}^6$ uniformly sampled in the range $[0.5, 8.5]$ meters following dataset statistics.
For finer levels $l<6$, the search space is refined using a local window around the upsampled depth prediction 
$D^{l+1}$
from the previous level:
\begin{equation}
\begin{split}
    \mathcal{D}^l &= \left\{ D^{l+1}_{\uparrow} + \delta \mid \delta \in \{-\epsilon_l, \dots, \epsilon_l\} \right\} \\
    \epsilon_l &= 2^{l-1} \times 0.02\text{m}
\end{split}
\end{equation}
This recursive refinement allows the model to achieve sub-pixel depth accuracy while maintaining a small number of depth planes at each resolution. For each level $l$ and depth hypothesis $d_k \in \mathcal{D}^l$, the source features are warped into the target camera frustum via homographies induced by the corresponding fronto-parallel plane,
\begin{equation}
    \mathcal{F}_{i \rightarrow 4}^{l}(d_k) = \mathcal{W}\!\left(\mathcal{F}_i^l, \mathbf{H}_{i \rightarrow 4}(d_k)\right),
    \label{eq:homography}
\end{equation}
where $\mathbf{H}_{i \rightarrow 4}(d_k)$ denotes the plane-induced homography and $\mathcal{W}(\cdot)$ the differentiable warping operator.

To preserve rich matching cues, we construct a grouped correlation volume following~\cite{guo2019group}. The $C$ feature channels are partitioned into $G$ groups, and for each source-view pair $(i,j)$ the group-wise correlation at depth $d_k$ is computed as
\begin{equation}
    \mathcal{C}_{ij,g}^l(d_k) = \frac{1}{C/G} 
    \sum_{c \in \mathcal{G}_g}
    \left(\mathcal{F}_{i \rightarrow 4}^{l}(d_k)\right)_c
    \left(\mathcal{F}_{j \rightarrow 4}^{l}(d_k)\right)_c,
\end{equation}
where $\mathcal{G}_g$ denotes the set of channels in group $g$. 

The resulting correlation volumes for all view pairs are concatenated along the channel dimension together with the proposed depths, the warped foreground masks and to enforce geometric consistency across scales, we introduce a latent propagation mechanism. The upscaled latent $Z^{l+1}$ from the feature projector is also passed as an additional conditioning input. A shared encoder processes the concatenated features to produce a latent representation. A decoder then regresses a depth residual $\Delta^l$ and an opacity map $A^l$, yielding the refined depth estimate $D^l = D^{l+1}_{\uparrow} + \Delta^l$.

\subsection{Hierarchical Feature Fusion and Image Synthesis}
\label{sec:architecture-projection}

The final stage synthesizes the target RGB image by warping and fusing source-view features under the estimated depth. For each level $l$ and source view $i$, the features are projected into the target frame using $D^l$, resulting in $\mathcal{F}_{i \rightarrow 4}^{l}(D^l)$. To account for occlusions and view-dependent effects, we introduce a confidence prediction network that produces per-view confidence weights $W_i^l \in [0,1]^{h\times w}$ from the warped features and geometric metadata like azimuth and elevation that we compute per camera. The fused representation is computed as
\begin{equation}
    \mathcal{F}_{\text{fused}}^l =
    \sum_{i=1}^{3} W_i^l \odot
    \mathcal{F}_{i \rightarrow 4}^{l}(D^l),
\end{equation}
where $\odot$ denotes element-wise multiplication.

Image synthesis is performed by a strictly hierarchical decoder. At each level $l$, the decoder receives the fused features $\mathcal{F}_{\text{fused}}^l$, the refined depth map $D^l$, the opacity map $A^l$, and the upsampled latent feature $Z_{\uparrow}^{l+1}$ from the coarser level
and outputs the new latents $Z^l$.

This coarse-to-fine formulation ensures that global structure estimated at lower resolutions regularizes high-frequency detail synthesis at finer scales. At the finest level, the latent feature $Z^0$ is mapped to the final RGB prediction $I_4$ via a lightweight refinement head.

\subsection{Loss Functions}
\label{sec:losses}

Our objective combines \textit{reconstruction losses} to ensure pixel-level accuracy, 
\textit{geometric losses} to supervise scene geometry, and 
\textit{perceptual losses} to encourage visual fidelity. 
All losses are applied hierarchically, providing supervision at multiple stages of the network.

\paragraph{Reconstruction Losses.}
We use an L1 \textbf{reconstruction loss} to penalize pixel-wise differences between the predicted image $I_4$ and the ground truth image $\hat{I}_4$:
\begin{equation}
    \mathcal{L}_1 = \| \hat{I}_4 - I_4 \|_1.
\label{eq:l1-loss}
\end{equation}
To enforce consistency in the RGB output we apply an L1 RGB  \textbf{pyramid loss} to the first 3 channels of the latent output on each pyramid level except for the uppermost compared to the bilinearly downsampled ground truth image:
\begin{equation}
    \mathcal{L}_{\text{RGB}} = \sum_{l>0} \| \hat{I}_{4\downarrow} - Z^l[:3] \|_1.
\label{eq:rgb-loss}
\end{equation}

\paragraph{Geometric Losses.}

To enforce consistency in the predicted depth maps $D^l$ across layers, we apply a masked L1 \textbf{depth loss} loss with respect to the ground truth depth $\hat{D}_4$ using the foreground mask $M_4$:
\begin{equation}
    \mathcal{L}_{\text{depth}} =
    \sum_{l} \frac{1}{\| M_{4\downarrow}\|_1}
    \left\|
      \left| \hat{D}_{4\downarrow} - D^l \right| \odot M_{4\downarrow}
    \right\|_1.
\label{eq:depth-loss}
\end{equation}
To further regularize the depth updates we apply an \textbf{offset loss} based on the specified window size $\epsilon_l$ at each level on the predicted residuals $\Delta^l$:
\begin{equation}
    \mathcal{L}_{\text{offset}} = \sum_{l}
    \frac{1}{\|  M_{4\downarrow} \|_1} \left\| \text{ReLU}(|\Delta^l| - \epsilon_l) \odot M_{4\downarrow} \right\|_1
\label{eq:offset-loss}
\end{equation}
The predicted alpha maps $A^l$ are supervised using an L2 \textbf{alpha loss} via the corresponding ground truth $\hat{A}_4$:

\begin{equation}
    \mathcal{L}_{\alpha} = \sum_{l} \| \hat{A}_{4\downarrow} - A^l \|_2.
\label{eq:alpha-loss}
\end{equation}

\paragraph{Perceptual Losses.}

To address the over-smoothing commonly introduced by pixel-wise losses, we incorporate a \textbf{VGG loss} based on the VGG-19 network, as introduced by Johnson et al.~\cite{johnson2016perceptuallossesrealtimestyle}:
\begin{equation}
    \mathcal{L}_{\text{vgg}} =  \sum_{j} \beta_j \left\| V_j(\hat{I}_4) - V_j(I_4) \right\|_1,
\label{eq:vgg-loss}
\end{equation}
where $V_j(\cdot)$ denotes the activation of the $j$-th VGG layer and $\beta_j$ is the corresponding layer weight.
To further capture texture and style information, we use a \textbf{style loss} following the approach of Gatys et al.~\cite{gatys2015neuralalgorithmartisticstyle}. It compares the Gram matrices $G_j(\cdot)$ of the predicted and ground truth features:
\begin{equation}
\begin{split}
    &\mathcal{L}_{\text{style}} = \sum_{j} \beta_j \left\| G_j(\hat{I}_4) - G_j(I_4) \right\|_2, \\
    &\text{where } G_j(\cdot) = (V_j(\cdot))^\mathsf{T} (V_j(\cdot)).
\end{split}
\label{eq:style-loss}
\end{equation}

\paragraph{Final Loss.}
The overall loss
is a weighted sum of the seven components:
\begin{equation}
\begin{split}
    \mathcal{L}_{\text{tot}} =
    \lambda_1 \mathcal{L}_1 &+ \lambda_2 \mathcal{L}_{\text{RGB}} +
    \lambda_3 \mathcal{L}_{\text{depth}} + \lambda_4 \mathcal{L}_{\text{offset}} + \lambda_5 \mathcal{L}_{\alpha} \\
    & + \lambda_6 \mathcal{L}_{\text{vgg}} + \lambda_7 \mathcal{L}_{\text{style}},
\end{split}
\label{eq:loss-composition}
\end{equation}
where $\lambda_i$ are hyperparameters controlling the relative contribution of each term.
You can find more details in \cref{sec:training-details}.

\subsection{Implementation Details}

We train on a synthetic dataset comprising 24,753 samples generated from 357 3D assets~\cite{polyhaven, sketchfab, humgen3d} and additional curated Blender scenes.
Each sample contains three source views and a target view, which is randomly sampled via barycentric coordinates with an additional $\pm 20$cm forward-axis jitter to ensure robustness.
The model is trained end-to-end on a single NVIDIA A40
and we target the NVIDIA RTX 4090 for inference, where our hardware-efficient profile enables real-time performance. Optimization details, augmentation scheme, and loss weights are provided in \cref{sec:training-details}.

%% file: sec/4_experiments.tex
\section{Experiments}
\label{sec:experimments}

\input{tables/quantitative}

We evaluate our method against a diverse set of state-of-the-art baselines. These include offline optimization methods: Nerfacto-big~\cite{nerfstudio}, Splatfacto-big~\cite{nerfstudio}, and FrugalNeRF~\cite{lin2024frugalnerf}; and feed-forward (online) methods: ENeRF~\cite{lin2022efficient}, GPS-Gaus\-sian~\cite{zheng2024gpsgaussian}, GPS-Gaussian+~\cite{zhou2024gpsplus}, and RIFTCast~\cite{zingsheim2025riftcast}. We report standard metrics: Peak Signal-To-Noise Ratio (PSNR), Structural Similarity (SSIM), and Learned Perceptual Image Patch Similarity (LPIPS).

Our evaluation spans six benchmarks covering diverse lighting, motion, and outfits: DNA Rendering~\cite{cheng2023dnarenderingdiverseneuralactor}, LLFF~\cite{mildenhall2020nerfrepresentingscenesneural}, MVHumanNet~\cite{xiong2024mvhumannet}, RIFTCast~\cite{zingsheim2025riftcast}, THuman2.1~\cite{tao2021function4d}, and ZJUMoCap~\cite{peng2021neural}.
Notably, LLFF is included as an out-of-distribution stress test to evaluate robustness on forward-facing real-world scenes with significantly larger depth ranges.
Methods requiring per-scene optimization (hereafter denoted with *) are evaluated on a representative subset.
Unless stated otherwise, experiments are performed at a resolution of $1024^2$.

\subsection{Qualitative and Quantitative Comparisons}

\begin{figure*}[t]
\centering

\newcolumntype{M}[1]{>{\centering\arraybackslash}m{#1}}

\setlength{\tabcolsep}{1pt}
\renewcommand{\arraystretch}{0.5}
\begin{tabular}{M{0.03\linewidth} *{5}{M{0.185\linewidth}}}

& GT & RIFTCast & ENeRF & GPS-Gaussian+ & Ours \\

\rotatebox{90}{\makecell{RIFTCast}} &
\includegraphics[width=\linewidth]{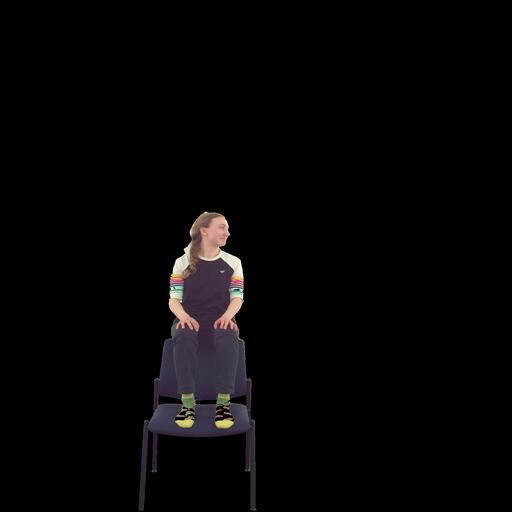} &
\includegraphics[width=\linewidth]{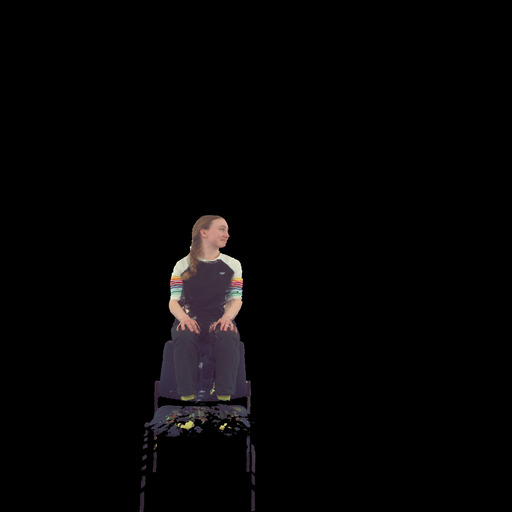} &
\includegraphics[width=\linewidth]{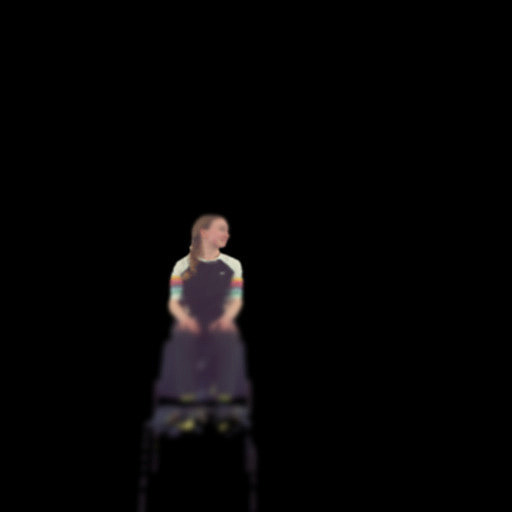} &
\includegraphics[width=\linewidth]{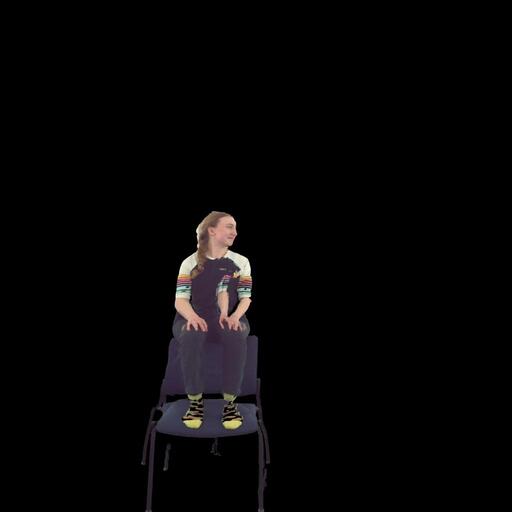} &
\includegraphics[width=\linewidth]{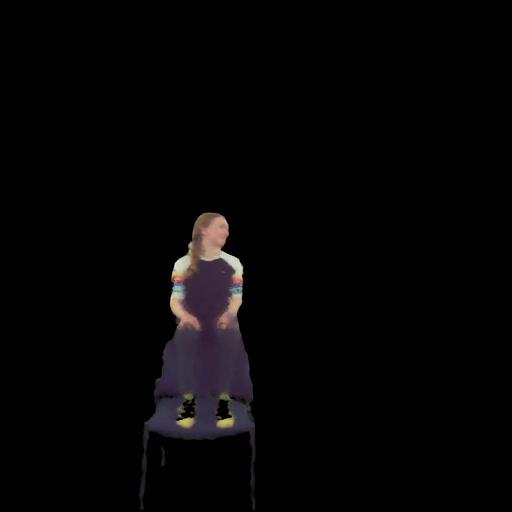} \\

\rotatebox{90}{\makecell{DNA}} &
\includegraphics[width=\linewidth]{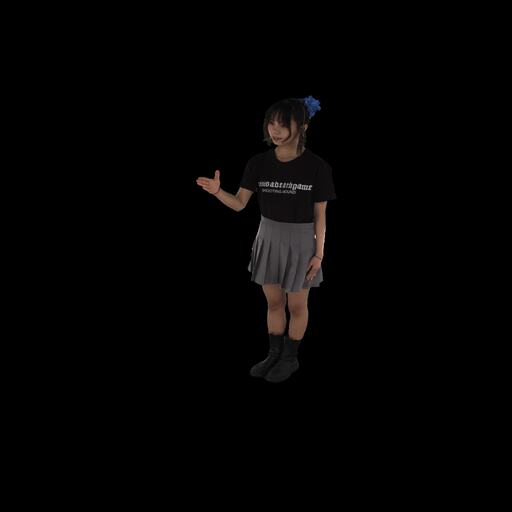} &
\includegraphics[width=\linewidth]{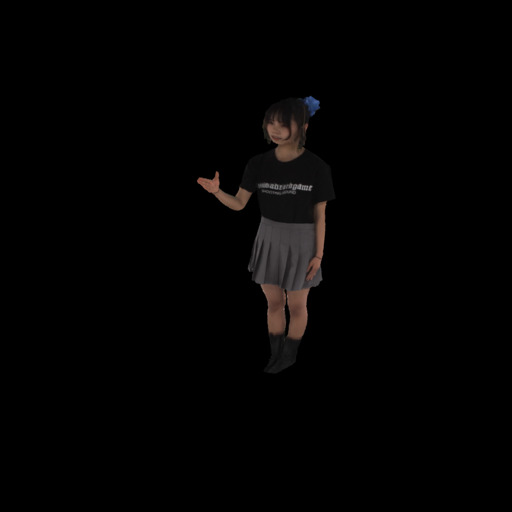} &
\includegraphics[width=\linewidth]{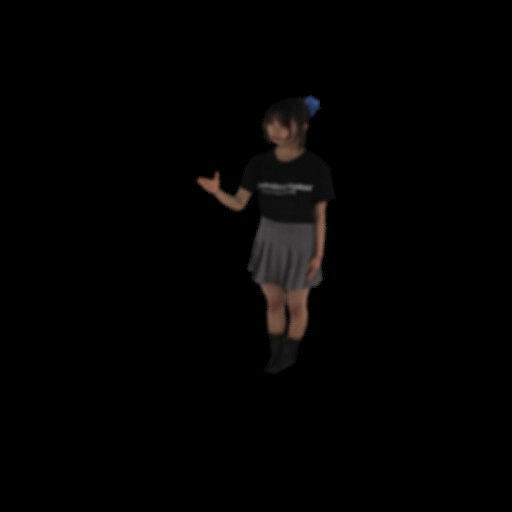} &
\includegraphics[width=\linewidth]{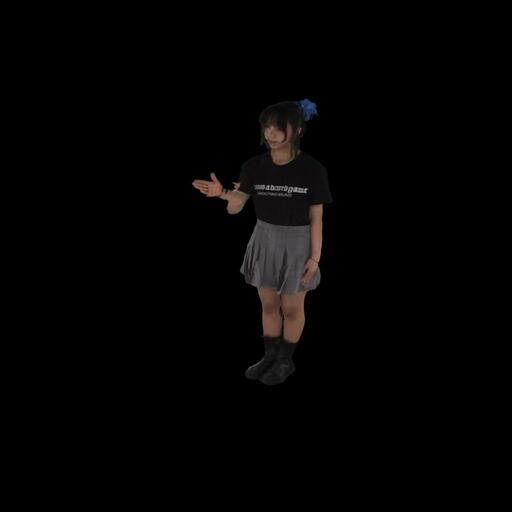} &
\includegraphics[width=\linewidth]{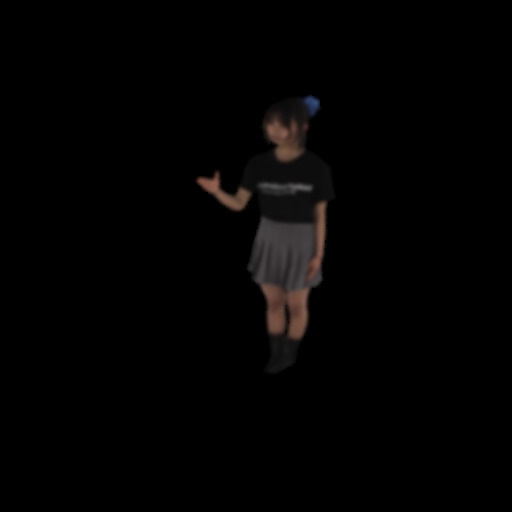} \\

\rotatebox{90}{\makecell{THuman2.1}} &
\includegraphics[width=\linewidth]{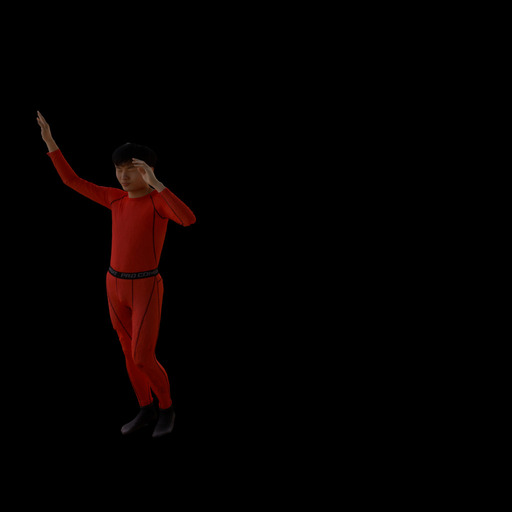} &
\includegraphics[width=\linewidth]{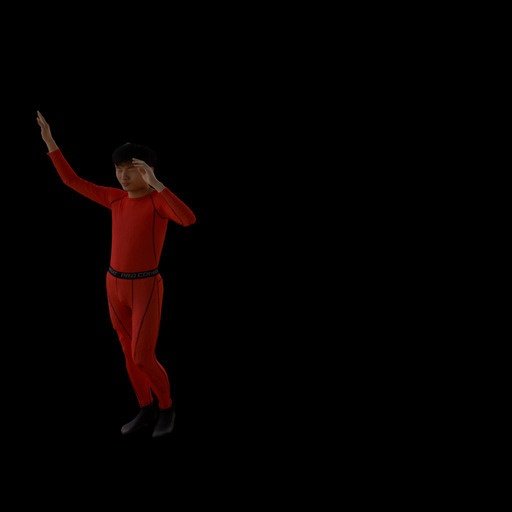} &
\includegraphics[width=\linewidth]{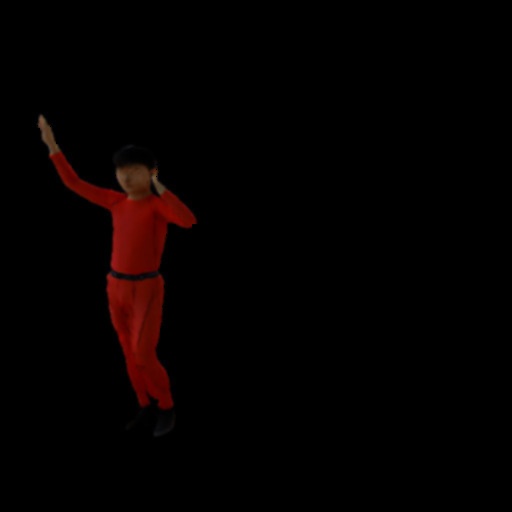} &
\includegraphics[width=\linewidth]{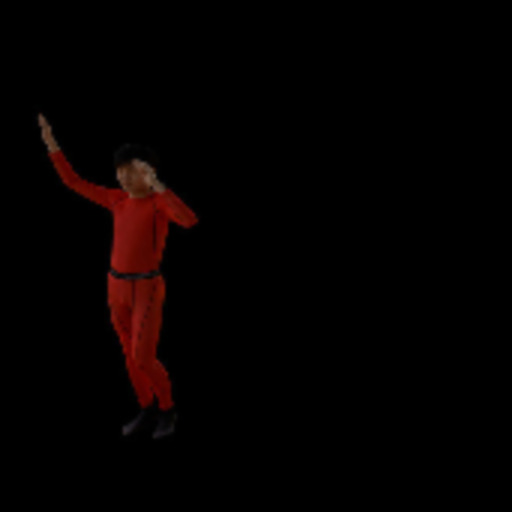} &
\includegraphics[width=\linewidth]{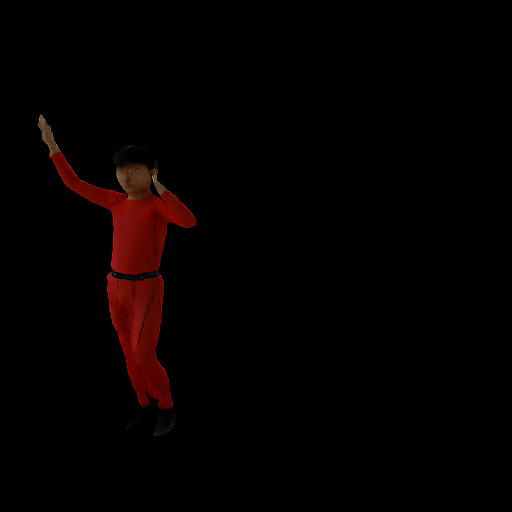} \\

\rotatebox{90}{\makecell{LLFF}} &
\includegraphics[width=\linewidth]{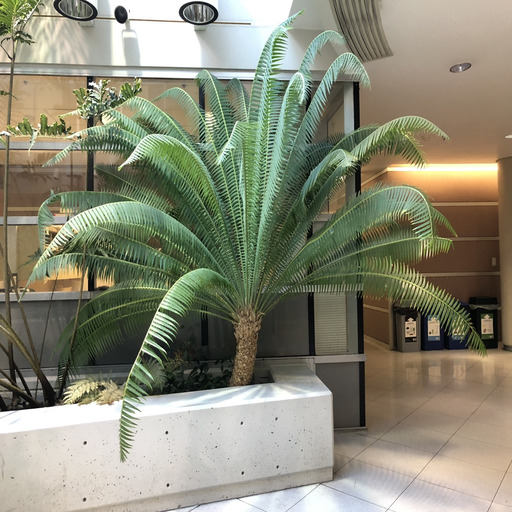} &
\begin{tikzpicture}
\fill[black] (0,0) rectangle (\linewidth,\linewidth);
\begin{scope}
\clip (0,0) rectangle (\linewidth,\linewidth);
\draw[gray!90, line width=4pt] (0,0) -- (\linewidth,\linewidth);
\draw[gray!90, line width=4pt] (0,\linewidth) -- (\linewidth,0);
\end{scope}
\node[text=white] at (0.5\linewidth,0.5\linewidth) {Not applicable};
\end{tikzpicture} &
\includegraphics[width=\linewidth]{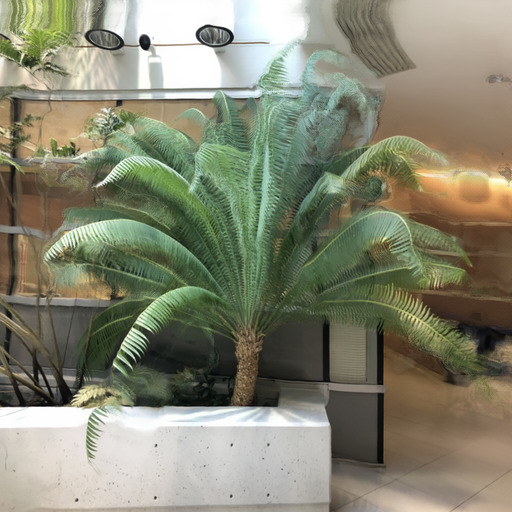} &
\includegraphics[width=\linewidth]{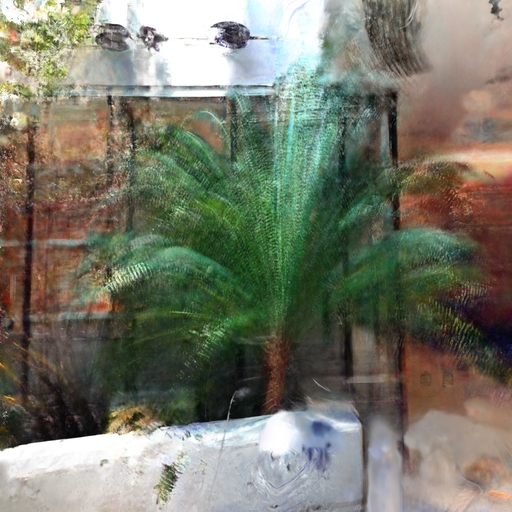} &
\includegraphics[width=\linewidth]{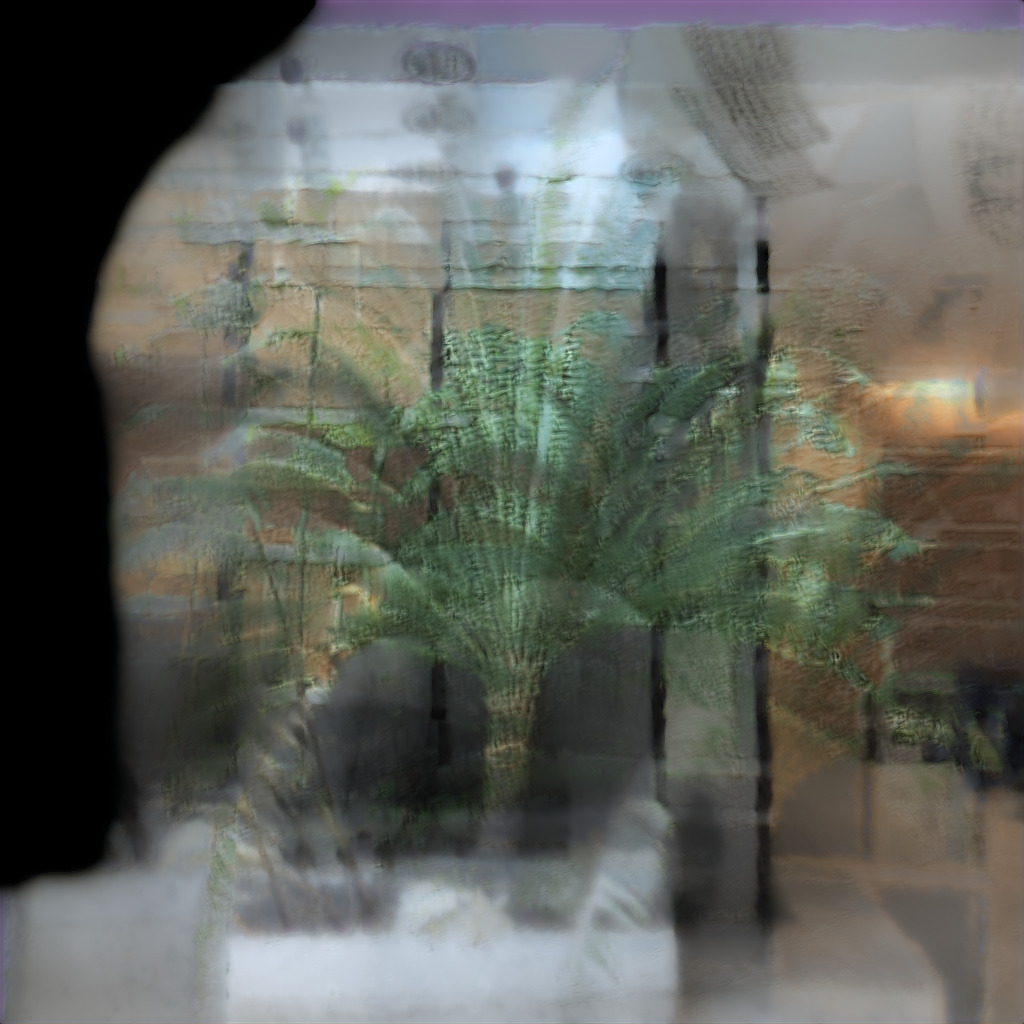} \\

\end{tabular}

\caption{Qualitative comparison on multiple datasets for human capture and non-human containing scenes with real-time capable methods.
While the first method, RIFTCast~\cite{zingsheim2025riftcast}, requires all views available and isolated scene content, the other methods can achieve novel view synthesis with 2 to 3 views.}

\label{fig:visual-comparison}
\end{figure*}
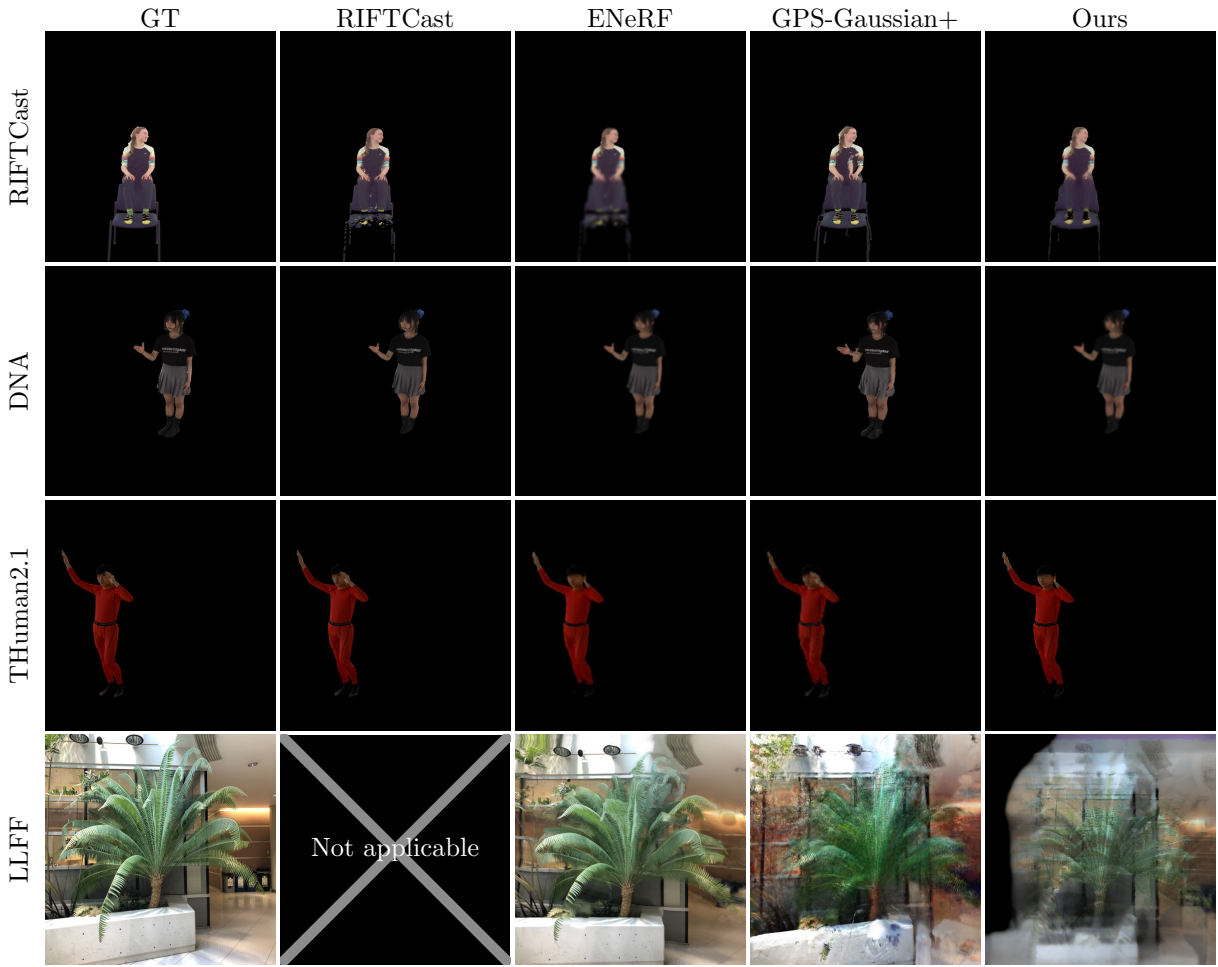

Fig.~\ref{fig:visual-comparison} compares our method against recent feed-forward and real-time capable baselines. Across human captures and general scenes, 2-view approaches such as GPS-Gaussian/\,+ frequently suffer from depth ambiguity under wide baselines, leading to floating structures, duplicated geometry, and temporal instability. By conditioning synthesis on a geometrically consistent 3-camera triplet and performing depth-guided feature projection, our method produces more stable geometry and preserves fine details such as faces and extremities more reliably than 2-view systems. In contrast to RIFTCast, which relies on processing all available views and full-view mask sets to construct a visual hull, our approach maintains consistent results using only the selected triplet.

Quantitative results in \cref{tab:metric-1-comparison} highlight the quality--efficiency trade-off. While per-scene optimization methods (e.g., Splatfacto-big, Nerfacto-big) achieve the strongest metrics, their multi-minute reconstruction overhead makes them impractical for interactive settings. Among sparse-view methods using only 2--3 inputs, our approach consistently achieves competitive or superior quality compared to prior feed-forward baselines on human-centric datasets (e.g., MVHumanNet and ZJUMoCap), indicating robustness to imperfect foreground masks and real capture artifacts.
Although our model is trained on synthetic indoor scenes with limited depth ranges, we evaluate on LLFF as an out-of-distribution stress test. The scenes violate several assumptions of our pipeline (large depth ranges, unbounded environments, and imperfect masks), yet the model still captures coarse geometry but lacks high-frequency detail.
This experiment highlights the robustness of the depth-guided projection, though the method is not optimized for such settings.

Overall, these results demonstrate that combining geometric triplet selection with lightweight depth-guided fusion improves stability and fidelity in sparse, wide-baseline regimes while retaining practical runtime characteristics.

\subsection{Runtime and Memory Analysis}

In Table~\ref{tab:runtime-memory}, we analyze the 
computational demands of the available approaches.
Optimization-based methods (Nerfacto, Splatfacto) provide superior results but fail to meet the requirements of real-time applications due to their heavy per-scene training requirements. 

Our method, particularly when optimized via TensorRT (Ours\textsuperscript{RT}), demonstrates a significant advantage in resource-constrained environments. At $1024 \times 1024$ resolution, our optimized model achieves 40 FPS with a peak memory footprint of only 2.2 GB, outperforming GPS-Gaussian and RIFTCast in terms of the quality-to-memory ratio.
While several recent methods report real-time inference through highly optimized deployment pipelines, these implementations are often not publicly available. In our work, we aim to ensure reproducibility by providing the full TensorRT deployment configuration alongside the model. Combined with our lightweight architecture and hierarchical feature projection, this enables a transparent and practical path to real-time performance.

\begin{table*}[th]
\centering
\footnotesize
\caption{
Memory and Speed comparisons of the tested methods on a NVIDIA RTX 4090 GPU using the official implementations.
We also provide a version of our method optimized with TensorRT (Ours\textsuperscript{RT}) to allow for real-time usage.
}
\label{tab:runtime-memory}
\scalebox{0.95}{
\begin{tabular}{@{}lccc ccc@{}}
\toprule
& \multicolumn{3}{c}{1024 x 1024} & \multicolumn{3}{c}{2048 x 2048} \\
\cmidrule(lr){2-4} \cmidrule(lr){5-7}
& Memory$\downarrow$ & Training time$\downarrow$ & Inference time$\downarrow$ 
& Memory$\downarrow$ & Training time$\downarrow$ & Inference time$\downarrow$ \\

\midrule

Nerfacto-big*~\cite{nerfstudio} (x)
& 12.1 GB & 100.3 min & 7.8 ms
& 12.8 GB & 108.3 min & 9.6 ms \\

Splatfacto-big*~\cite{nerfstudio} (x)
& \phantom{0}1.2 GB & 11.2 min & 1.3 ms
& \phantom{0}2.6 GB & 13.1 min & 1.8 ms \\

FrugalNeRF*~\cite{lin2024frugalnerf} (3)
& 20.2 GB & 12.3 min & 13.2 ms
& 20.4 GB & 16.2 min & 16.5 ms \\

\midrule

ENeRF~\cite{lin2022efficient} (3) 
& \phantom{0}8.1 GB & 0 ms & 97.3 ms
& 13.1 GB & 0 ms & 233.7 ms \\

GPS-Gaussian~ \cite{zheng2024gpsgaussian} (2) 
& \phantom{0}3.2 GB & 0 ms & 73.7 ms
& \phantom{0}8.1 GB & 0 ms & 119.2 ms \\

GPS-Gaussian+~ \cite{zhou2024gpsplus} (2)
& \phantom{0}3.4 GB & 0 ms & 72.4 ms
& \phantom{0}8.2 GB & 0 ms & 154.9 ms \\

RIFTCast~ \cite{zingsheim2025riftcast} (x)
& \phantom{0}5.7 GB & 0 ms & 47.3 ms
& \phantom{0}6.2 GB & 0 ms & 50.7 ms \\

\midrule

Ours (3)
& \phantom{0}7.1 GB & 0 ms & 117.1 ms
& 20.6 GB & 0 ms & 542.7 ms \\

Ours\textsuperscript{RT} (3)
& \phantom{0}2.2 GB & 0 ms & 24.5 ms
& \phantom{0}8.0 GB & 0 ms & 109.5 ms \\
\bottomrule
\end{tabular}
}
\end{table*}

\subsection{Ablations}

\begin{table*}
  \centering
  \caption{Quantitative comparison of different model variants with TensorRT optimization. Showcase of the improvements and sweet spot our different modules and choices bring with them at 1024 $\times$ 1024 resolution on DNA Rendering dataset.}
  \begin{tabular}{@{}lrrrrr@{}}
    \toprule
    Method \hspace{4cm} & PSNR$\uparrow$ & \hspace{3pt}SSIM$\uparrow$ & \hspace{3pt}LPIPS$\downarrow$ & \hspace{6pt}Memory$\downarrow$ & \hspace{3pt}FPS$\uparrow$ \\
    \midrule
    full model & 25.9 & 0.952 & 0.074 & 2.2 GB & 40.8 \\
    \specialrule{.01em}{0.1em}{0.15em} 
    2 views & 23.6 & 0.913 & 0.089 & 1.7 GB & 49.4 \\
    \specialrule{.01em}{0.1em}{0.15em}
    \# channels halved & 24.9 & 0.927 & 0.090 & 2.2 GB & 41.3 \\
    \# channels doubled & 26.1 & 0.952 & 0.072 & 2.6 GB & 32.6 \\
    \specialrule{.01em}{0.1em}{0.15em} 
    3-level pyramid & 22.8 & 0.893 & 0.112 & 2.2 GB & 42.4 \\
    6-level pyramid & 25.8 & 0.953 & 0.077 & 2.2 GB & 41.0 \\
    \specialrule{.01em}{0.1em}{0.15em} 
    w/o residual depth & 21.5 & 0.884 & 0.134 & 2.2 GB & 40.9 \\
    w/o residual projector addition & 25.2 & 0.947 & 0.079 & 2.2 GB & 40.9 \\
    w/o prior guided depth & 23.8 & 0.937 & 0.085 & 1.9 GB & 46.7 \\
    \bottomrule
  \end{tabular}
  \label{tab:ablations}
\end{table*}

We analyze the trade-offs between model complexity, synthesis quality, and runtime on our held-out test set (see \cref{tab:ablations}). Our final configuration ("full") provides the best balance for interactive performance.

\paragraph{Architectural Scalability.} 
Varying the channel width confirms that our base configuration provides a good trade-off between quality and efficiency. Halving the channels increases speed but noticeably degrades high-frequency details, while doubling them yields only minor improvements at substantially higher memory cost. We also find that the 7-level pyramid is important for wide-baseline interpolation. Shallower pyramids force the model to search over larger depth ranges at coarse scales, increasing memory usage and reducing depth stability.

\paragraph{Residual Learning and Stability.} 
Predicting depth updates ($\Delta^l$) and feature residuals allows the network to focus on local corrections rather than regressing absolute values at each scale~\cite{gu2020cascadecostvolumehighresolution, liu2021fcfrnetfeaturefusionbased}. Removing these residual connections leads to unstable training and visible "shimmering" artifacts. The residual hierarchy also helps propagate coarse geometric structure to finer levels.

\begin{figure*}[tb]
    \centering
    \begin{subfigure}[t]{0.23\linewidth}
        \centering
        \includegraphics[width=0.9\linewidth]{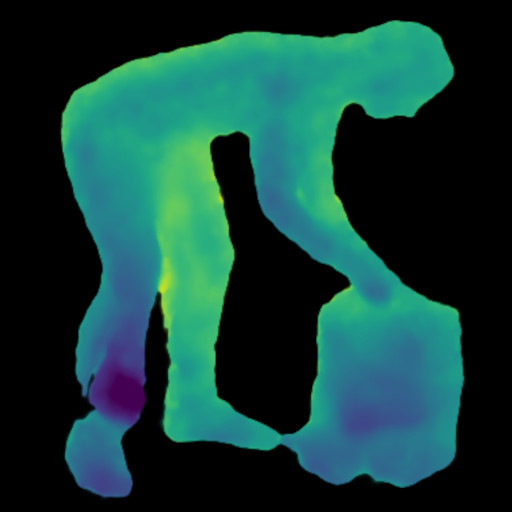}
        \caption{Depth Map generated by our method.}
        \label{fig:visual-depthmap}
    \end{subfigure}
    \hfill
    \begin{subfigure}[t]{0.23\linewidth}
        \centering
        \includegraphics[width=0.9\linewidth]{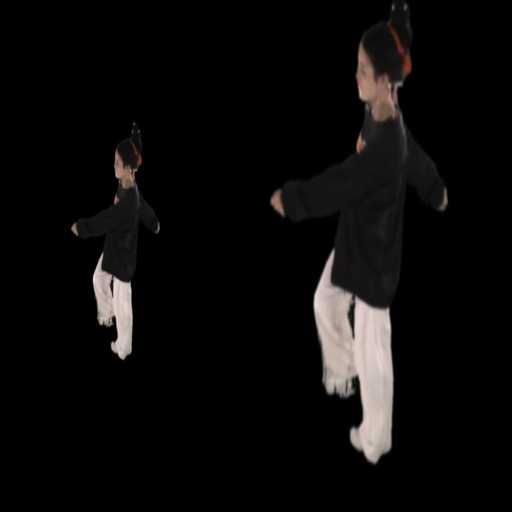}
        \caption{Generalization capabilities of our model to varying depth of the camera.}
        \label{fig:visual-camera-z-movement}
    \end{subfigure}
    \hfill
    \begin{subfigure}[t]{0.23\linewidth}
        \centering
        \includegraphics[width=0.9\linewidth]{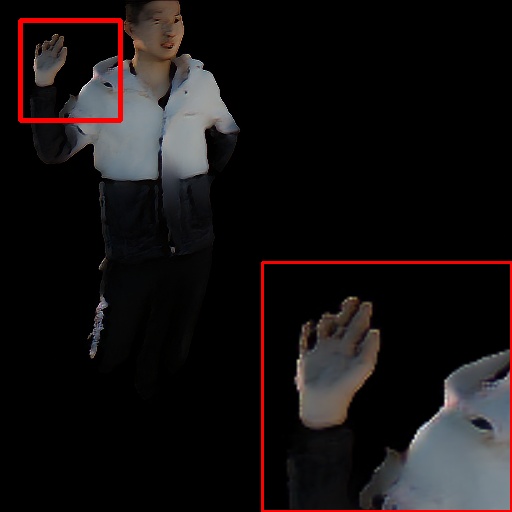}
        \caption{Generalization capabilities of our model to higher resolutions such as 2048 $\times$ 2048.}
        \label{fig:visual-fhd-generalization}
    \end{subfigure}
    \hfill
    \begin{subfigure}[t]{0.23\linewidth}
        \centering
        \includegraphics[width=0.9\linewidth]{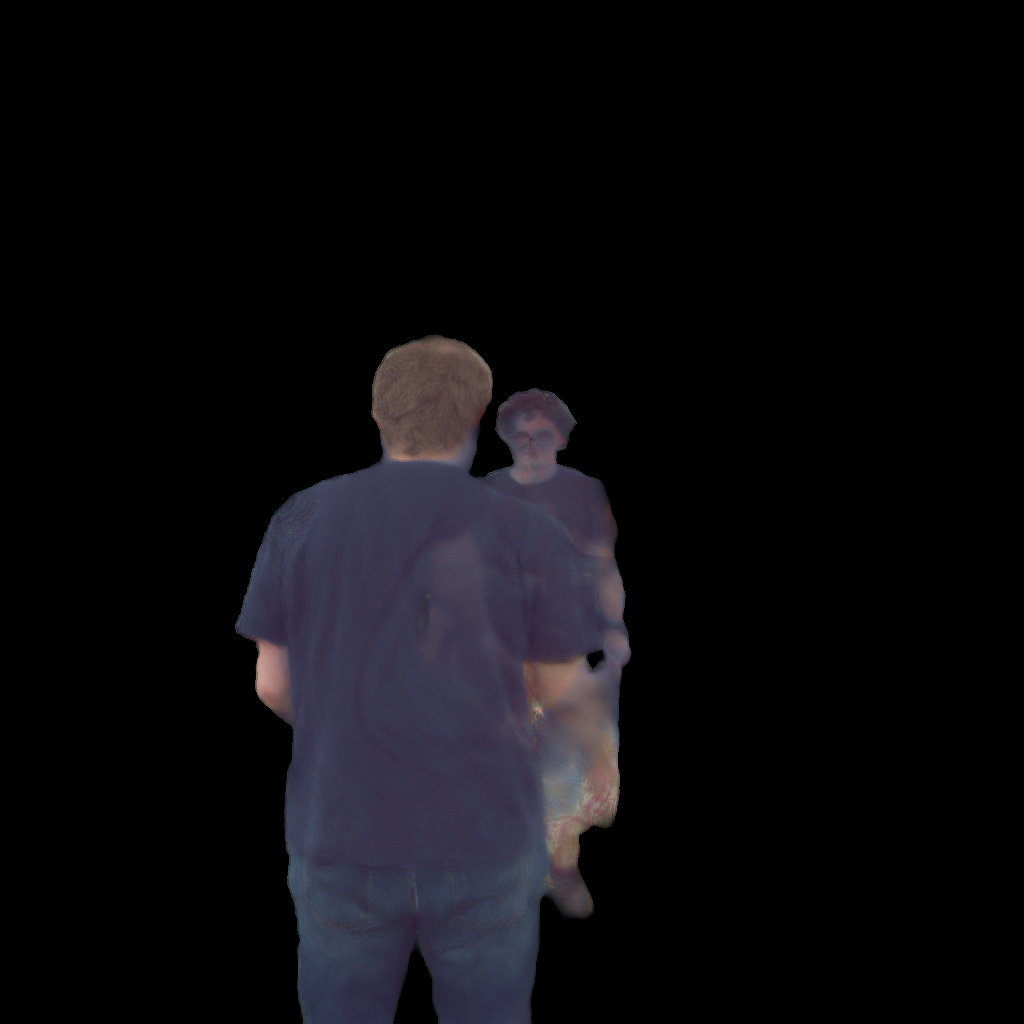}
        \caption{A failure case where two persons behind one another get washed out.}
        \label{fig:visual-failure}
    \end{subfigure}
    \caption{
      Generalization behavior of 3DTV across geometry, viewpoint, and resolution. Despite synthetic-only training, the model predicts coherent depth and stable synthesis under varying camera depths and scales to higher output resolutions without retraining.
    }
\end{figure*}

\paragraph{Generalization and Viewpoint Stability.}
Unlike methods such as Gaussian Splatting~\cite{kerbl20233dgaussiansplattingrealtime}, which often exhibit "popping" or blurring when moving outside trained bounds, our geometry-aware projection maintains high viewpoint stability. By learning a robust relationship between feature warping and metric depth, our model generalizes effectively to novel views and higher resolutions. Notably, our method scales from $1024^2$ to $2048^2$ without finetuning. As shown in \cref{fig:visual-fhd-generalization}, our depth-guided projection remains resolution-agnostic, preserving sharp textures where coordinate-based networks typically suffer from upsampling blur.

\paragraph{} Further details on results and ablations can be found in \cref{sec:quantitative}.

\subsection{Limitations}
While our Ghost-based backbone is efficient, achieving true real-time performance ($>60$ FPS) on would require further optimization such as weight quantization.
Furthermore, the model is currently constrained to three-view inputs and indoor capture stages with limited applicability to large scale scenes.
This limitation is reflected in our LLFF stress test, where large scene scales exceed the training distribution.

Additionally, our method targets bounded interpolation around the selected camera triplet, handling large extrapolation beyond the camera hull remains future work.
Finally, blurriness in complex regions suggests that increasing feature map density or adopting temporal feature propagation could better resolve high-frequency details and disocclusions.

%% file: tables/quantitative.tex
\begin{table*}[t]
\centering
\footnotesize
\caption{Quantitative comparison across multiple datasets for human capture and scenes without humans.
Values in parentheses indicate the number of source views used for novel view synthesis, where (x) denotes all available views.
Colored boxes show the \colorbox{best}{best} and \colorbox{second}{second-best} results per metric of sparse-view methods (2-3 images).
LLFF represents a challenging out-of-distribution scenario with complex scene geometry, which lies outside the training distribution of our model.
}
\label{tab:metric-1-comparison}
\scalebox{0.95}{
\begin{tabular}{@{}llccc ccc ccc@{}}
\toprule
& & \multicolumn{3}{c}{RIFTCast~\cite{zingsheim2025riftcast}} & \multicolumn{3}{c}{DNA Rendering~\cite{cheng2023dnarenderingdiverseneuralactor}} & \multicolumn{3}{c}{LLFF~\cite{mildenhall2020nerfrepresentingscenesneural}} \\
\cmidrule(lr){3-5} \cmidrule(lr){6-8} \cmidrule(lr){9-11}
&
& PSNR$\uparrow$ & SSIM$\uparrow$ & LPIPS$\downarrow$ 
& PSNR$\uparrow$ & SSIM$\uparrow$ & LPIPS$\downarrow$ 
& PSNR$\uparrow$ & SSIM$\uparrow$ & LPIPS$\downarrow$ \\

\midrule

\multirow{3}{*}{\rotatebox{90}{Dense}\hspace{4pt}} & Nerfacto-big*~\cite{nerfstudio} (x)
& 26.2 & 0.891 & 0.078
& 29.5 & 0.920 & 0.061
& (11.7) & (0.308) & (0.706) \\

& Splatfacto-big*~\cite{nerfstudio} (x)
& 30.7 & 0.920 & 0.058
& 34.5 & 0.976 & 0.028
& (13.7) & (0.438) & (0.669) \\

& RIFTCast~\cite{zingsheim2025riftcast} (x)
& 29.1 & 0.953 & 0.033
& 29.9 & 0.932 & 0.030
& n/a & n/a & n/a \\

\midrule

\multirow{5}{*}{\rotatebox{90}{Sparse}\hspace{4pt}} & FrugalNeRF*~\cite{lin2024frugalnerf} (3)
& 23.0 & 0.894 & 0.114
& 24.2 & 0.928 & 0.098
& \cellcolor{second}19.1 & \cellcolor{second}0.609 & \cellcolor{second}0.273 \\

& ENeRF~\cite{lin2022efficient} (3)
& \cellcolor{second}24.9 & \cellcolor{second}0.939 & \cellcolor{best}0.072
& \cellcolor{second}26.0 & 0.949 & \cellcolor{second}0.073
& \cellcolor{best}21.3 & \cellcolor{best}0.712 & \cellcolor{best}0.209 \\

& GPS-Gaussian~\cite{zheng2024gpsgaussian} (2)
& 21.8 & 0.916 & 0.087
& \cellcolor{best}26.1 & \cellcolor{second}0.951 & \cellcolor{best}0.071 
& n/a & n/a & n/a \\

& GPS-Gaussian+~\cite{zhou2024gpsplus} (2)
& 22.0 & 0.907 & 0.084
& 24.9 & 0.973 & 0.081
& 11.8 & 0.685 & 0.327 \\

\cmidrule(lr){2-11}

& Ours Best (3)
& \cellcolor{best}25.7 & \cellcolor{best}0.941 & \cellcolor{second}0.073
& 25.9 & \cellcolor{best}0.952 & 0.074
& 10.3 & 0.676 & 0.317 \\

\specialrule{.1em}{0.3em}{0.1em} 

\specialrule{.1em}{0.1em}{0.3em} 

& & \multicolumn{3}{c}{MVHuman~\cite{xiong2024mvhumannet}} & \multicolumn{3}{c}{THuman2.1~\cite{tao2021function4d}} & \multicolumn{3}{c}{ZJUMoCap~\cite{peng2021neural}} \\
\cmidrule(lr){3-5} \cmidrule(lr){6-8} \cmidrule(lr){9-11}
&
& PSNR$\uparrow$ & SSIM$\uparrow$ & LPIPS$\downarrow$ 
& PSNR$\uparrow$ & SSIM$\uparrow$ & LPIPS$\downarrow$
& PSNR$\uparrow$ & SSIM$\uparrow$ & LPIPS$\downarrow$ \\

\midrule

\multirow{3}{*}{\rotatebox{90}{Dense}\hspace{4pt}} & Nerfacto-big*~\cite{nerfstudio} (x)
& 26.3 & 0.948 & 0.082
& 25.9 & 0.937 & 0.098
& 24.1 & 0.921 & 0.110 \\

& Splatfacto-big*~\cite{nerfstudio} (x)
& 30.1 & 0.971 & 0.058
& 29.8 & 0.958 & 0.063
& 26.2 & 0.933 & 0.079 \\

& RIFTCast~\cite{zingsheim2025riftcast} (x)
& 28.3 & 0.965 & 0.066
& 28.1 & 0.955 & 0.068
& 25.7 & 0.933 & 0.080 \\

\midrule

\multirow{5}{*}{\rotatebox{90}{Sparse}\hspace{4pt}} & FrugalNeRF*~\cite{lin2024frugalnerf} (3)
& 22.6 & 0.887 & 0.102
& 23.2 & 0.900 & 0.096
& 22.1 & 0.881 & 0.107 \\

& ENeRF~\cite{lin2022efficient} (3)
& \cellcolor{second}25.2 & \cellcolor{best}0.945 & \cellcolor{second}0.078 
& \cellcolor{second}26.1 & \cellcolor{best}0.948 & \cellcolor{second}0.078
& \cellcolor{second}23.7 & \cellcolor{second}0.919 & \cellcolor{second}0.093 \\

& GPS-Gaussian~\cite{zheng2024gpsgaussian} (2)  
& 24.8 & \cellcolor{second}0.944 & 0.074
& 25.7 & 0.921 & 0.083
& 21.4 & 0.897 & 0.114 \\

& GPS-Gaussian+~\cite{zhou2024gpsplus} (2)
& 23.8 & 0.920 & 0.082
& 23.9 & 0.908 & 0.096
& 20.9 & 0.901 & 0.12 \\

\cmidrule(lr){2-11}

& Ours Best (3)
& \cellcolor{best}25.4 & 0.938 & \cellcolor{best}0.074
& \cellcolor{best}26.7 & \cellcolor{second}0.947 & \cellcolor{best}0.072
& \cellcolor{best}24.1 & \cellcolor{best}0.930 & \cellcolor{best}0.086 \\
\bottomrule
\end{tabular}
}
\end{table*}

%% file: sec/5_conclusion.tex
\section{Conclusion}
\label{sec:conclusion}

We introduced 3DTV, a feedforward framework for real-time sparse-view novel view synthesis that combines principled geometric view selection with efficient depth-guided fusion. By leveraging Delaunay triangulation to select geometrically consistent camera triplets and estimating target-view depth through a coarse-to-fine pyramid, our method enables stable wide-baseline interpolation from only three input cameras without per-scene optimization.

Extensive experiments across multiple multi-view benchmarks demonstrate that 3DTV achieves a strong balance between synthesis quality and computational efficiency, outperforming recent real-time sparse-view baselines while maintaining low latency. Despite being trained solely on synthetic data, the model generalizes robustly to real-world captures and diverse scene configurations. These results suggest that combining lightweight geometric reasoning with feedforward neural synthesis is a promising direction for scalable real-time free-viewpoint video rendering.

%% file: sec/6_acknowledgements.tex
\section*{Acknowledgements}

This work has been funded by the Ministry of Culture and Science North Rhine-Westphalia under grant number PB22-063A (InVirtuo 4.0: Experimental Research in Virtual Environments), and by the state of North RhineWestphalia as part of the Excellency Start-up Center.NRW (U-BO-GROW) under grant number 03ESCNW18B. 
Additionally, the authors gratefully acknowledge the access to the Marvin cluster of the University of Bonn and the support provided by the High Performance Computing \& Analytics Lab of the University of Bonn.

%% file: suppl/0_intro.tex
\section*{Supplementary}
\noindent This appendix provides additional technical details, expanded experimental results, and comprehensive visual comparisons to support the claims presented in the main paper.
Specifically, this supplemental material includes:
\begin{itemize}
    \item \textbf{Technical Implementation:} Detailed derivations regarding the Delaunay Triangulation used for mesh generation, along with a deep dive into the network architecture.
    \item \textbf{Training Protocols:} Comprehensive details on the training regime, including our choice of optimizer, specific loss function weightings, training duration, and the data augmentation strategies employed.
    \item \textbf{Dataset Characteristics:} A breakdown of the datasets used, including source resolutions and the specific preprocessing pipeline applied.
    \item \textbf{Extended Baseline Comparisons:} Evaluations against two additional baseline models. We provide their results here for completeness and to justify their exclusion from the main manuscript.
    \item \textbf{Qualitative \& Quantitative Analysis:} Expanded visual results comparing our method against the primary baselines and across various ablation studies.
\end{itemize}

%% file: suppl/1_triangulation.tex
\section{Delaunay Triangulation Ablation}
\label{sec:triangulation}

\begin{figure*} [t]
    \centering
    \includegraphics[width=0.9\textwidth]{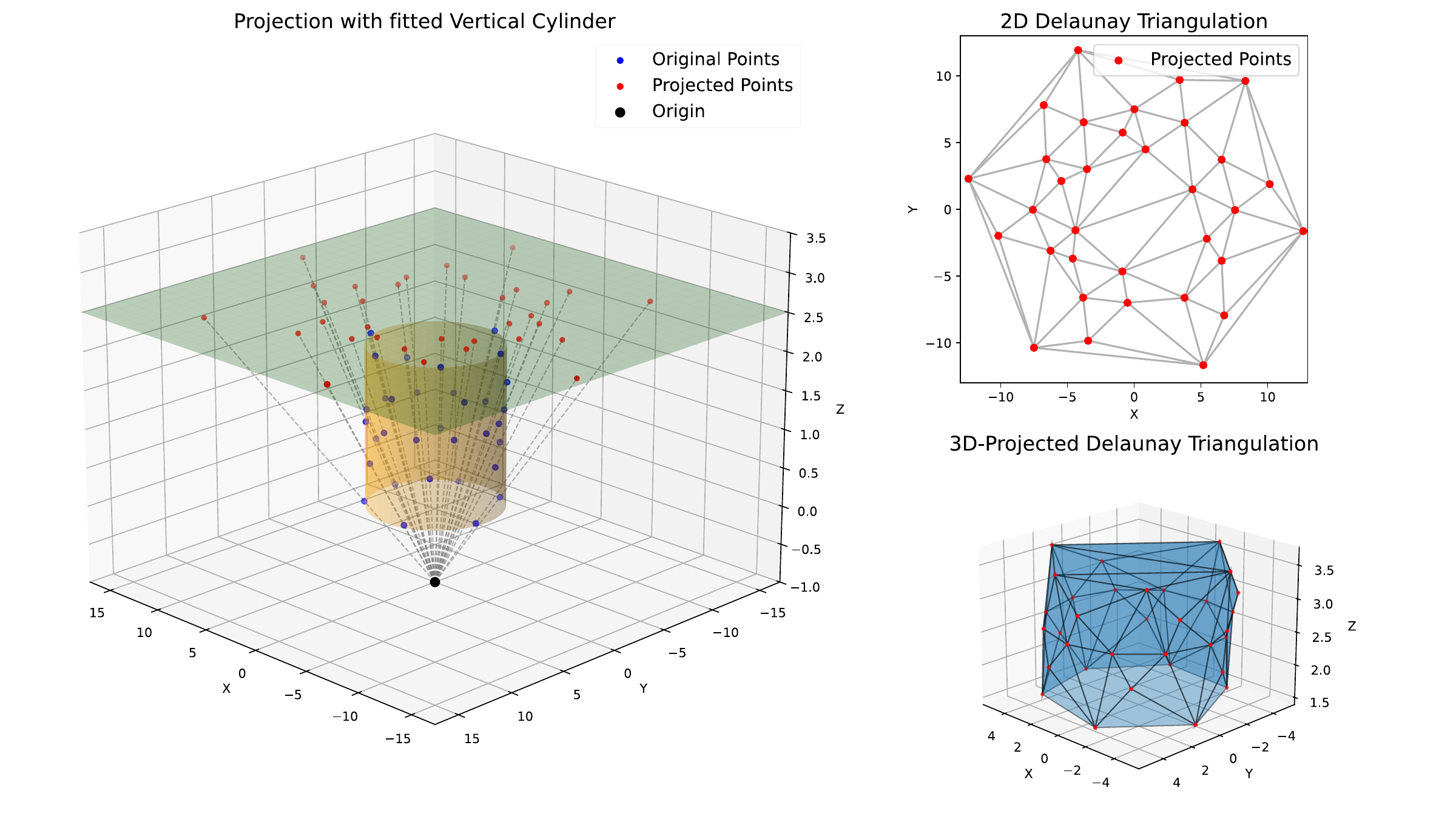}
    \caption{Exemplary Delaunay triangulation for the RIFTCast~\cite{zingsheim2025riftcast} capture stage setup. The scene has been scaled for visualization purposes.}
    \label{fig:triangulation}
\end{figure*}

As described in Sec. 3.1 of the main paper, the triangulation process for camera setups in a capture stage is visualized in Fig.~\ref{fig:triangulation}. First, a cylinder is fitted to the camera positions, and all positions are projected onto its surface. A ray is then cast from an \textit{Origin} point, defined as the center of the elongated base (visualized here as a shifted origin for better clarity), through each camera location and onto a 2D plane positioned parallel to the cylinder's top. 

Standard Delaunay triangulation is computed on this 2D plane, and the resulting faces are back-projected into 3D. These faces are subsequently used to identify the corresponding triangle of source cameras for a novel view. This triangulation approach offers several advantages:

\begin{itemize}
    \item It avoids rendering images from the floor while naturally supporting top-down views, which remain consistently covered.
    \item The method generalizes easily to other geometries, such as cuboid setups, by simply fitting a cuboid instead of a cylinder while keeping the remaining pipeline intact.
\end{itemize}

We evaluate the hyperparameters of this algorithm, specifically the elongation (offset) of the \textit{Origin} point and the placement of the 2D projection plane. Based on the results in Fig.~\ref{fig:triangulation-ablation-1} and Fig.~\ref{fig:triangulation-ablation-2}, we find that positioning the \textit{Origin} 1m below the cylinder and the projection plane 1m above the cylinder yields the optimal triangulation for capture stages. 

Increasing the distance of the \textit{Origin} or the projection plane excessively leads to artifacts: triangles at the bottom are frequently clipped, and the resulting mesh tends to follow vertical lines. This creates thin, elongated triangles rather than the near-equilateral triangles that are preferable for stable interpolation.

\begin{figure*} [t]
    \centering
    \includegraphics[height=0.9\textheight]{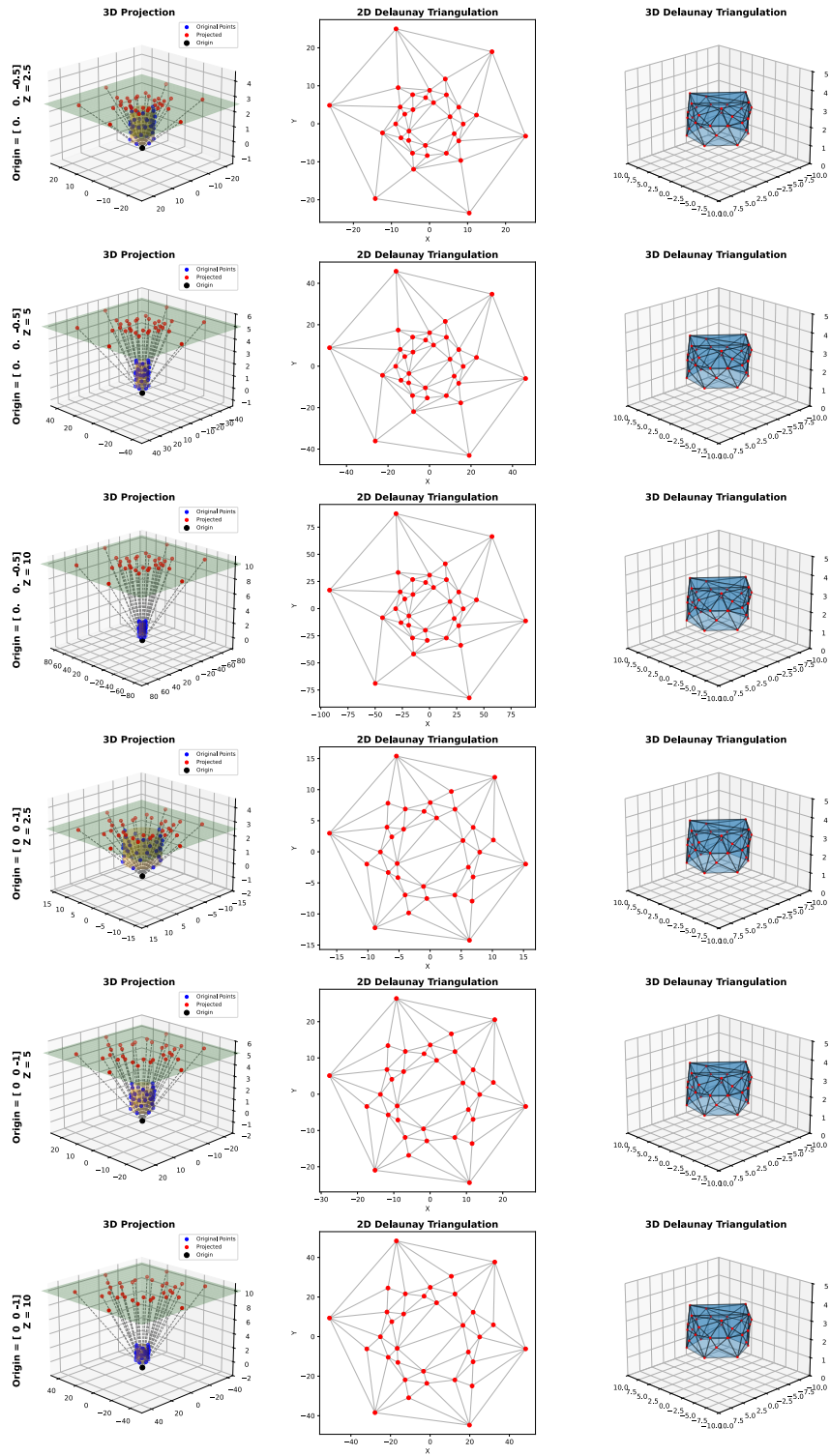}
    \caption{Evaluation of hyperparameter choices and their influence on the resulting triangulation.}
    \label{fig:triangulation-ablation-1}
\end{figure*}

\begin{figure*} [t]
    \centering
    \includegraphics[height=0.9\textheight]{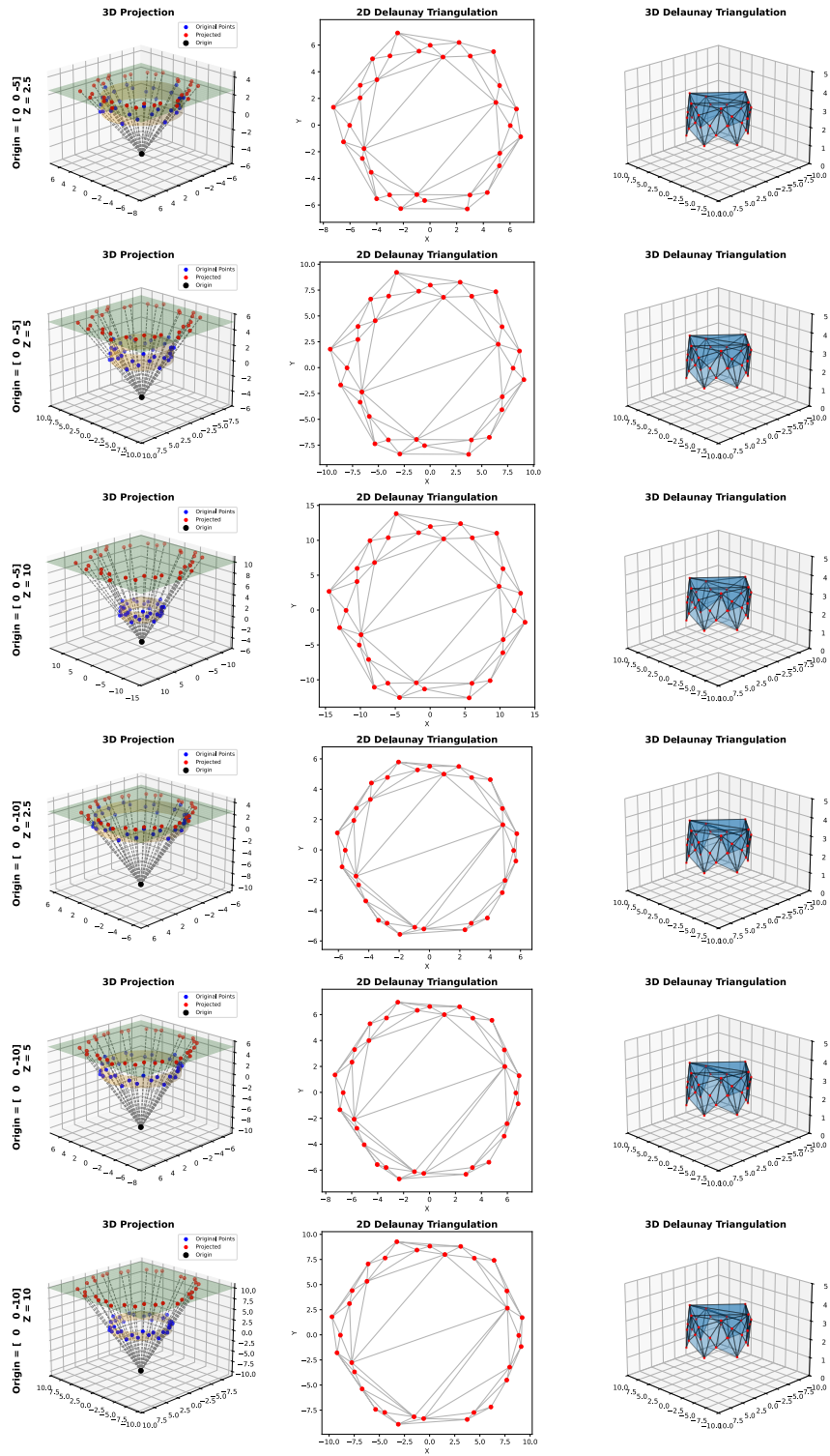}
    \caption{(Cont.) Evaluation of hyperparameter choices and their influence on the resulting triangulation.}
    \label{fig:triangulation-ablation-2}
\end{figure*}

%% file: suppl/2_training_details.tex
\section{Training Details}
\label{sec:training-details}

\subsection{Dataset Curation}

As our network requires ground truth depth maps for stable training and faster convergence, we could not use any real-world dataset for this task. Thus, we train on a synthetic dataset comprising 24,753 samples generated from 357 3D assets ~\cite{polyhaven, sketchfab, humgen3d} and additional curated Blender scenes. We downloaded various blender scenes and material maps from polyhaven ~\cite{polyhaven} and sketchfab ~\cite{sketchfab} ranging from coaches to chemistry workbenches and scaled the scenes to fit in a volume of $[-2m, 2m] \times [-2m, 2m] \times [0m, 2m]$ which approximately matches standard capture stage volumes. Additionally, we used the HumanGenerator3D add-on ~\cite{humgen3d} for Blender to render humans in different poses and clothing styles. Most of the scenes contain only one (complex) object so decided to add custom scenes. These scenes are mad up of randomly placed, scaled, and deformed cubes with different images covering the surfaces of each cube. This was used to help the model resolve ambiguous depth estimations for neighboring pixels. To cover a wide range of illumination patterns and real world lighting we used environment maps from polyhaven ~\cite{polyhaven} in Blender. As a final step, random camera positions and random camera amounts with different depth offsets along the approximative cylinder covering the volume are created. To render novel views, we apply the triangulation discussed in Sec. 3.1 in the main paper and place cameras based on randomly chosen barycentric weights. The orientation is adjusted accordingly to maintain a proper rotation matrix. With all cameras set up, 24,753 samples consisting of 3 source views and a target view have been rendered for training.

\subsection{Optimization and Training Schedule}

The 3DTV network is optimized using the Adam optimizer with an initial learning rate of $10^{-5}$. We employ a \texttt{OneCycleLR} scheduler with a maximum learning rate of $10^{-3}$, using a linear annealing strategy and a final division factor of 50.

Training is performed in a two-stage resolution-progressive manner to balance high-frequency detail with global structure:
\begin{itemize}
    \item \textbf{Phase 1 (Base):} 100 epochs at $512 \times 512$ resolution. This phase required approximately 4 days of compute on a single NVIDIA A40 GPU with a batch size of 2.
    \item \textbf{Phase 2 (Fine-tuning):} 25 epochs at $1024 \times 1024$ resolution to refine high-fidelity textures and interpolation boundaries. This phase required about 36 more hours under the same conditions.
\end{itemize}

The total training objective is a multi-task loss formulated as:
\begin{equation}
\begin{split}
    \mathcal{L}_{\text{tot}} = \lambda_{1} \mathcal{L}_1 &+ \lambda_{2} \mathcal{L}_{\text{RGB}} + \lambda_{3} \mathcal{L}_{\text{depth}} + \lambda_{4} \mathcal{L}_{\text{offset}} + \lambda_{5} \mathcal{L}_{\alpha} \\
    &+ \lambda_{6} \mathcal{L}_{\text{vgg}} + \lambda_{7} \mathcal{L}_{\text{style}}
\end{split}
\end{equation}

The weighting coefficients $\lambda_i$ are adjusted throughout the training phases to prioritize structural convergence initially, followed by perceptual refinement. The specific schedule is detailed in Table~\ref{tab:loss_weights}.

\begin{table*}[h]
    \centering
    \caption{\textbf{Loss Weighting Schedule.} Weights for each loss component across the three training phases.}
    \begin{tabular}{lccc}
        \hline
        \textbf{Loss Term} \hspace{0.5cm} & \textbf{Phase 1.1} ($\leq 25$) \hspace{0.1cm} & \textbf{Phase 1.2} ($26-100$) \hspace{0.1cm} & \textbf{Phase 2} ($> 100$) \\ \hline
        $\lambda_{1}$ & 2.0 & 2.0 & 2.0 \\
        $\lambda_{2}$ & 2.0 & 2.0 & 2.0 \\
        $\lambda_{3}$ & 1.0 & 1.0 & 0.5 \\
        $\lambda_{4}$ & 1.0 & 1.0 & 0.5 \\
        $\lambda_{5}$ & 1.0 & 1.0 & 1.0 \\
        $\lambda_{6}$ & 0.0 & 0.5 & 0.5 \\
        $\lambda_{7}$ & 0.0 & 0.0 & 0.5 \\ \hline
    \end{tabular}
    \label{tab:loss_weights}
\end{table*}

\subsection{Data Augmentation Pipeline}
To ensure robustness against varied capture conditions and sensor noise, we implement a custom augmentation suite that helps us bridge perfect synthetic training and noisy real-world data conditions. Let $I$ be the input image and $D$ the corresponding depth map:

\begin{enumerate}
    \item \textbf{Photometric Jitter:} Standard color jittering is applied to brightness ($0.3$), contrast ($0.3$), saturation ($0.2$), and hue ($0.2$).
    \item \textbf{Exposure and Gamma:} With a 30\% probability, we apply $I_{out} = I^{\gamma} \cdot e$, where $\gamma \sim \mathcal{U}(0.7, 1.3)$ and $e \sim \mathcal{U}(0.8, 1.0)$.
    \item \textbf{Depth-Conditioned Background Noise:} Since the renderings all contain black backgrounds and we don't want the model to overfit to that we simulate uncertainty in distant regions or "missing" geometry where $D(p) = 0$ (black), we inject Gaussian noise $\epsilon \sim \mathcal{N}(0, \sigma^2)$ exclusively to the background:
    \[
    I_{noisy} = 
    \begin{cases} 
    I + \epsilon & \text{if } D(p) = 0 \\
    I & \text{otherwise}
    \end{cases}
    \]
    where $\sigma \sim \mathcal{U}(0.00, 0.05)$.
    \item \textbf{Post-Processing Simulation:} To mimic real-world capture artifacts, we apply a 3x3 Gaussian blur ($\sigma \in [0.01, 0.5]$) with 50\% probability. Additionally, we simulate quantization/banding with 30\% probability by discretizing the color space into $N \in \{128, 192, 255\}$ levels.
    \item \textbf{Stochastic Scaling:} Random scaling and cropping are applied (20\% chance). For $1024^2$ inputs, the scale factor $s \sim \mathcal{U}(0.5, 0.7)$, which is halved for $512^2$ inputs.
\end{enumerate}

\subsection{Deployment and Inference Optimization}
For real-time performance, the model is deployed using NVIDIA TensorRT. The PyTorch weights are first exported to an ONNX intermediate representation. We then utilize the \texttt{trtexec} tool to compile a Bfloat16 (BF16) optimized engine. The compilation uses the \texttt{--best} flag to enable exhaustive search of the CUDA kernel space for the target hardware. Throughput benchmarks are reported as the average over 100 inference cycles with a warm up phase beforehand of 10 cycles. To prevent a decrease in model performance due to the reduction in precision, the model is trained already in Bfloat16.

%% file: suppl/3_datasets.tex
\section{Dataset and Evaluation Details}
\label{sec:datasets}

To ensure a fair comparison across all methods, all evaluations are performed at a standardized resolution of $1024 \times 1024$ pixels. For datasets with higher native resolutions, we apply bilinear downsampling. To maintain geometric consistency, the camera intrinsic matrices $K$ are scaled by the corresponding ratios in both the $x$ and $y$ dimensions.

\subsection{RIFTCast ~\cite{zingsheim2025riftcast}}
The RIFTCast dataset contains 32 dynamic scenes captured via a 35-camera multi-view rig. The sequences feature complex multi-actor interactions (e.g., playing drums) and vary from 2 to 20 seconds at 25 FPS.
\begin{itemize}
    \item \textbf{Data Usage:} We utilize the provided RGB images, binary masks, and background matting. For real-time methods, we evaluate across all 34 available sequences.
    \item \textbf{Evaluation Split:} One camera is reserved strictly for evaluation. For offline baseline comparison, we restrict training to the first set of images across the first 10 scenes.
\end{itemize}

\subsection{DNA Rendering ~\cite{cheng2023dnarenderingdiverseneuralactor}}
DNA Rendering is a large-scale, high-fidelity dataset for human performance rendering. We follow the selection protocol of RIFTCast~\cite{zingsheim2025riftcast}, specifically utilizing the first 67 scenes from Part 2.
\begin{itemize}
    \item \textbf{Preprocessing:} Original $2448 \times 2048$ images are downscaled to $1024 \times 1024$.
    \item \textbf{Split:} We use the 5-megapixel camera subset at frame 80, excluding cameras 3, 8, 19, and 40. Offline methods are trained on the first 10 scenes using all available views except the test cameras.
\end{itemize}

\subsection{LLFF (Local Light Field Fusion) ~\cite{mildenhall2020nerfrepresentingscenesneural}}
The LLFF dataset consists of forward-facing scenes with varying scales. We utilize the standard scenes: \textit{fern, flower, horns, leaves, orchids,} and \textit{room}.
\begin{itemize}
    \item \textbf{Configuration:} To optimize performance for 2-view and 3-view real-time setups, we selected camera poses on a single horizontal level rather than the original 3-view split.
    \item \textbf{Cleaning:} We explicitly removed blurry frames that negatively impact offline optimization, specifically images 4030 and 4037 from \textit{fern} and 2971 from \textit{flower}.
\end{itemize}

\begin{table*}[h]
    \centering
    \caption{\textbf{LLFF View Selection.} Training views for 2-view (*) and 3-view setups. Test views are held out for all methods; offline baselines utilize all remaining views.}
    \label{tab:supp-llff-setup}
    \begin{tabular}{@{}lll@{}}
        \toprule
        \textbf{Scene} & \textbf{Train Views} & \textbf{Test View} \\ \midrule
        Fern & IMG\_4035*, IMG\_4041*, IMG\_4029 & IMG\_4043 \\
        Flower & IMG\_2962*, IMG\_2970*, IMG\_2983 & IMG\_2984 \\
        Horns & DJI\_2020...842*, DJI\_2020...122*, DJI\_2020...633 & DJI\_2020...048 \\
        Leaves & IMG\_2997*, IMG\_3004*, IMG\_3011 & IMG\_3010 \\
        Orchids & IMG\_4468*, IMG\_4472*, IMG\_4476 & IMG\_4470 \\
        Room & DJI\_2020...006*, DJI\_2020...455*, DJI\_2020...082 & DJI\_2020...910 \\ \bottomrule
    \end{tabular}
\end{table*}

\subsection{MVHuman ~\cite{xiong2024mvhumannet}}
MVHuman provides high-resolution captures of human subjects in various attire. We utilize Part 1 of the dataset, downscaling the $1500 \times 2048$ images to $1024 \times 1024$.
\begin{itemize}
    \item \textbf{Data Usage:} We evaluate on sequences 100001 through 100010.
    \item \textbf{View Setup:} Cameras A055 and A005 are used for 2-view experiments, with A037 added for the 3-view setup. Camera A029 serves as the test view.
\end{itemize}

\subsection{THuman2.1 ~\cite{tao2021function4d}}
THuman2.1 consists of high-quality 3D human scans. We rendered these models using a random stage setup with virtual camera extrinsics mimicking the Bonn Telebench rig.
\begin{itemize}
    \item \textbf{Preprocessing:} Rendered at $2664 \times 2304$ and downscaled to $1024 \times 1024$.
    \item \textbf{Selection:} We take the first 500 objects for general evaluation. For offline methods, we limit the comparison to the first 10 examples to maintain computational feasibility.
\end{itemize}

\subsection{ZJUMoCap ~\cite{peng2021neural}}
We employ the updated ZJUMoCap dataset, using sequences \textit{313, 315, 377, 386, 387, 390, 392,} and \textit{393}. 
\begin{itemize}
    \item \textbf{Setup:} Images are natively $1024 \times 1024$. For 2-view setups, we use Cameras 1 and 5. For the 3-view setup, Camera 4 is included. Camera 3 is the held-out test view.
    \item \textbf{Temporal Constraint:} To ensure consistency across baselines, we limit the evaluation to the first 500 frames of each sequence.
\end{itemize}

%% file: suppl/4_comparisons_failure.tex
\section{Comparison Method Failures}
\label{sec:comparison-failures}

During our evaluation of state-of-the-art baselines, we explored several additional methods that were ultimately excluded from the main paper. In this section, we analyze the technical challenges and failure modes encountered with these methods, justifying their omission from our final quantitative comparison.

\subsection{Snap-Snap~\cite{lu2025snapsnaptakingimagesreconstruct}}

\begin{figure}[t]
    \centering
    \includegraphics[width=0.48\textwidth]{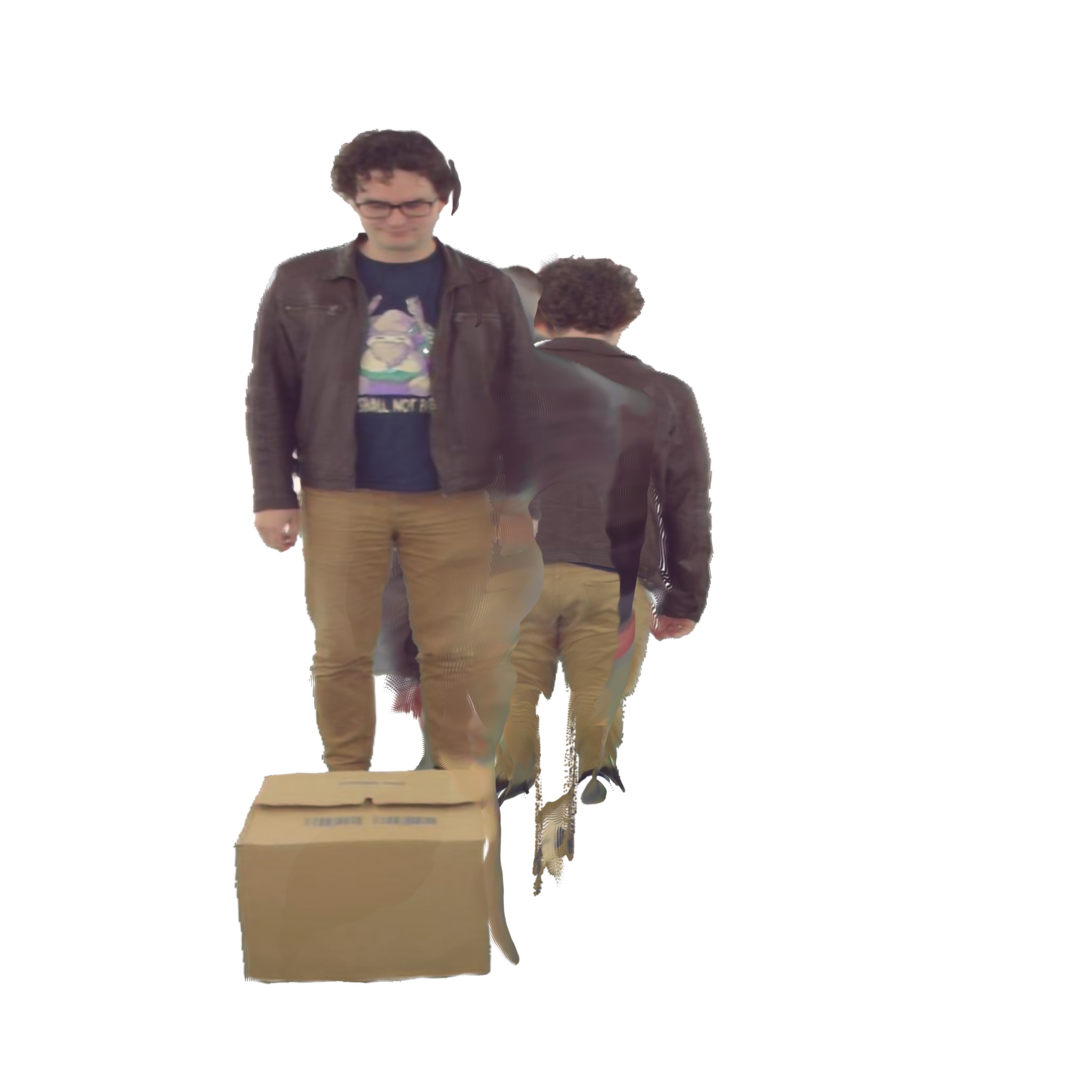}
    \caption{Snap-Snap produces significant ghosting and duplication artifacts when rendering the frontal view from the RIFTCast~\cite{zingsheim2025riftcast} dataset.}
    \label{fig:comparison-failures-snapsnap-1}
\end{figure}

\begin{figure}[t]
    \centering
    \includegraphics[width=0.48\textwidth]{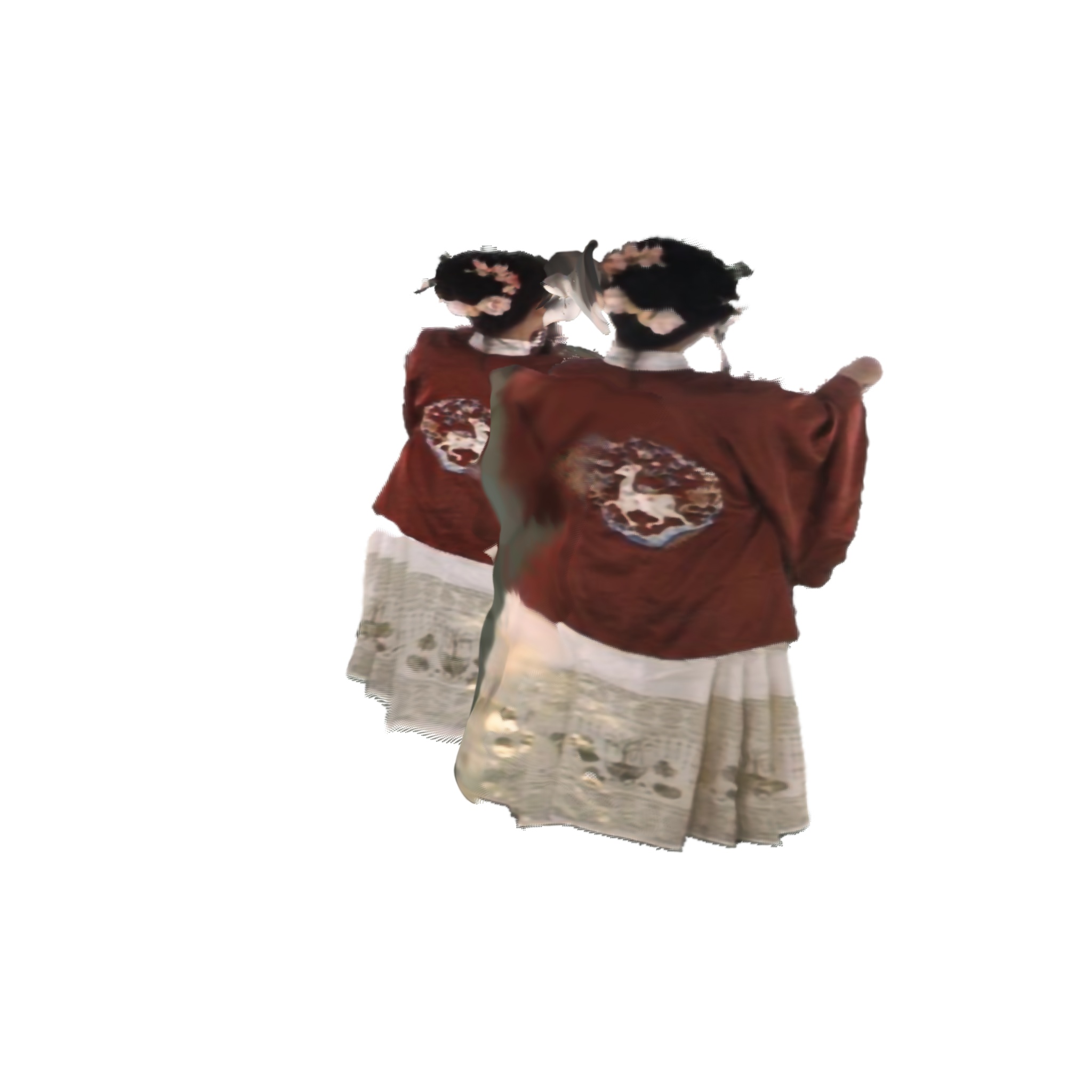}
    \caption{Example of reconstruction failure in DNA Rendering~\cite{cheng2023dnarenderingdiverseneuralactor}, where Snap-Snap fails to properly fuse the front and back views, resulting in duplicated geometry.}
    \label{fig:comparison-failures-snapsnap-2}
\end{figure}

Snap-Snap is a follow-up to GPS-Gaussian~\cite{zheng2024gpsgaussian} designed to reconstruct human 3D Gaussians in milliseconds using only a front and back view as input. We applied this approach to both the RIFTCast and DNA Rendering datasets; however, the results exhibited significant geometric artifacts. Specifically, we observed persistent "ghosting" and duplication of the subject, even when rendering from the original input views (see Fig.~\ref{fig:comparison-failures-snapsnap-1} and Fig.~\ref{fig:comparison-failures-snapsnap-2}). 

To mitigate this, we experimented with various camera configurations, attempting to match the required opposing-view setup as closely as possible within the constraints of real-world capture stages (where perfect centering and alignment are often infeasible). Despite these efforts, the artifacts remained consistent across diverse scenes and subjects. Due to these reproducibility issues and the subsequent limited availability of the official pre-trained weights, we opted to exclude Snap-Snap from the main comparison to ensure a fair and stable baseline evaluation.

\subsection{FWD~\cite{cao2022fwdrealtimenovelview}}

\begin{figure}[h]
    \centering
    \begin{subfigure}{0.45\textwidth}
        \centering
        \includegraphics[width=\linewidth]{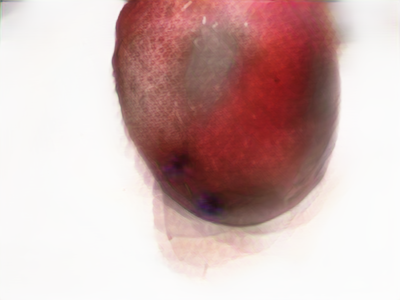}
        \caption{FWD Prediction}
        \label{fig:img1}
    \end{subfigure}
    \hfill
    \begin{subfigure}{0.45\textwidth}
        \centering
        \includegraphics[width=\linewidth]{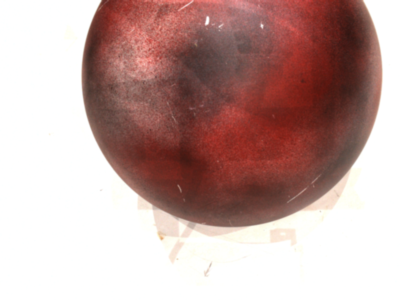}
        \caption{Ground Truth}
        \label{fig:img2}
    \end{subfigure}
    \caption{Failure analysis of FWD: The model fails to maintain proper object proportions and exhibits significant blurring compared to the ground truth.}
    \label{fig:comparison-failures-fwd}
\end{figure}

FWD is a feedforward method that supports an arbitrary number of input views, making it methodologically relevant to our proposed approach. However, in our tests using a 3-view configuration on the provided benchmark data, FWD struggled with geometric fidelity. As shown in Fig.~\ref{fig:comparison-failures-fwd}, the method introduces noticeable aspect-ratio distortion (i.e., "squeezing" the object) and produces overall blurry results. 

Furthermore, FWD is optimized for a native resolution of $640 \times 480$ pixels. Adapting the model to our $1024 \times 1024$ benchmark would have necessitated retraining from scratch—a task that proved infeasible given the time and computational resource constraints of the submission cycle. Consequently, we focus our comparison on baselines that provide more reliable high-resolution performance.

%% file: suppl/5_visual_ours.tex
\section{More qualitative evaluations}
\label{sec:evaluation}

In this section we provide more visual results and comparisons with regard to baseline methods as well as ablation studies we performed.

\subsection{Main Evaluations}

\begin{figure*}[t]
\centering

\newcolumntype{M}[1]{>{\centering\arraybackslash}m{#1}}

\setlength{\tabcolsep}{1pt}
\renewcommand{\arraystretch}{0.5}
\begin{tabular}{M{0.03\linewidth} *{5}{M{0.185\linewidth}}}

& GT & RIFTCast & ENeRF & GPS-Gaussian+ & Ours \\

\rotatebox{90}{\makecell{RIFTCast}} &
\includegraphics[width=\linewidth]{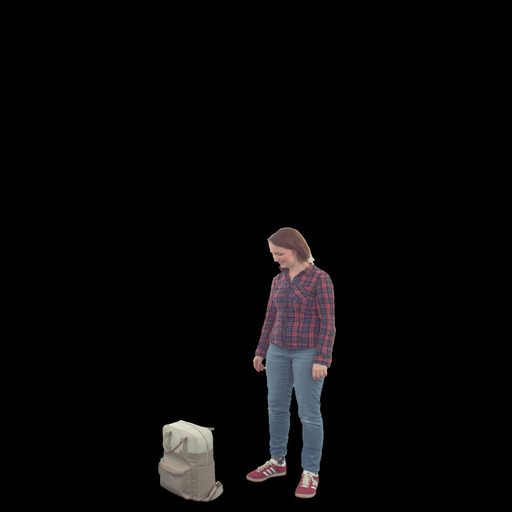} &
\includegraphics[width=\linewidth]{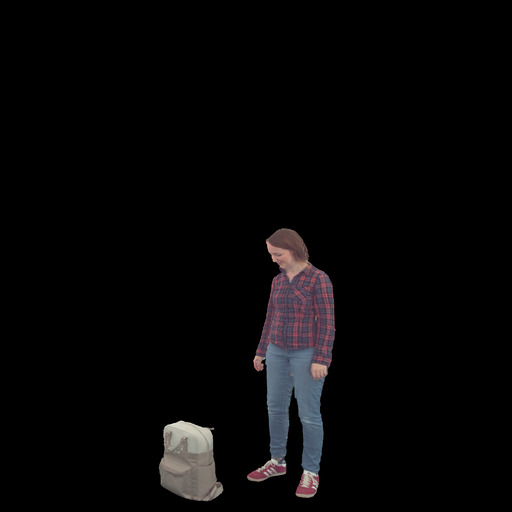} &
\includegraphics[width=\linewidth]{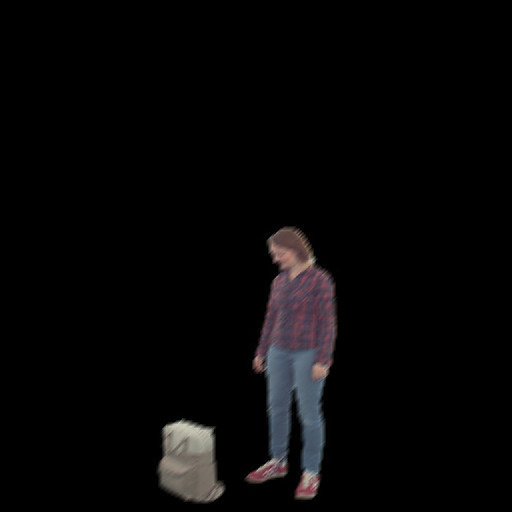} &
\includegraphics[width=\linewidth]{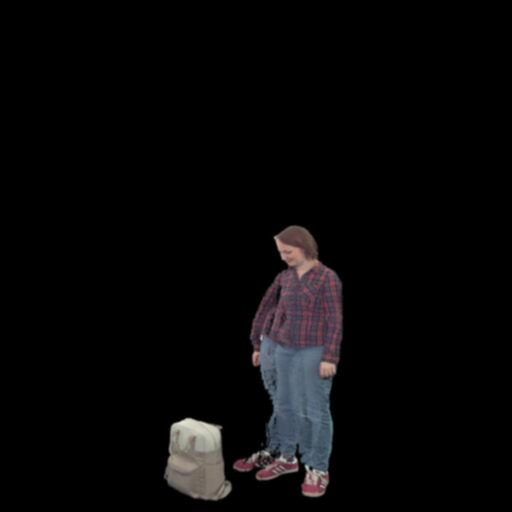} &
\includegraphics[width=\linewidth]{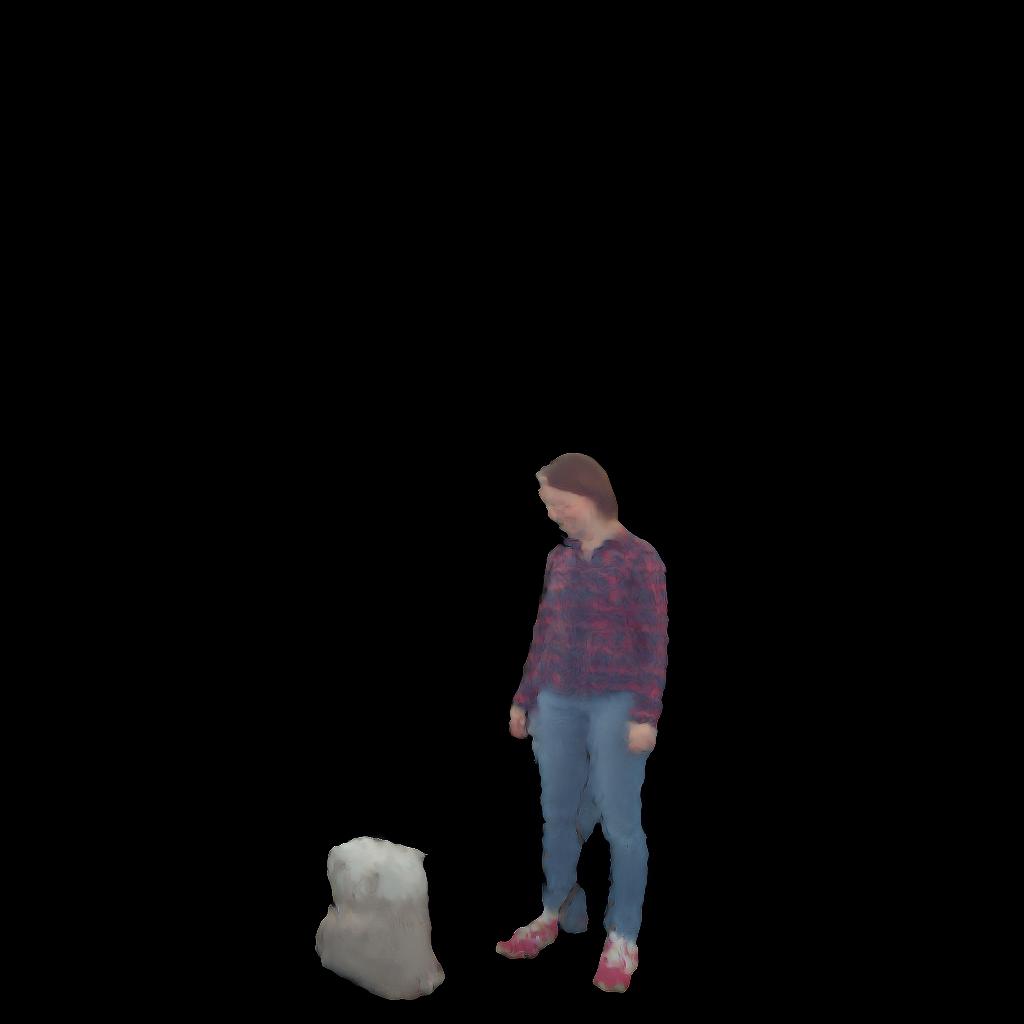} \\

\rotatebox{90}{\makecell{DNA}} &
\includegraphics[width=\linewidth]{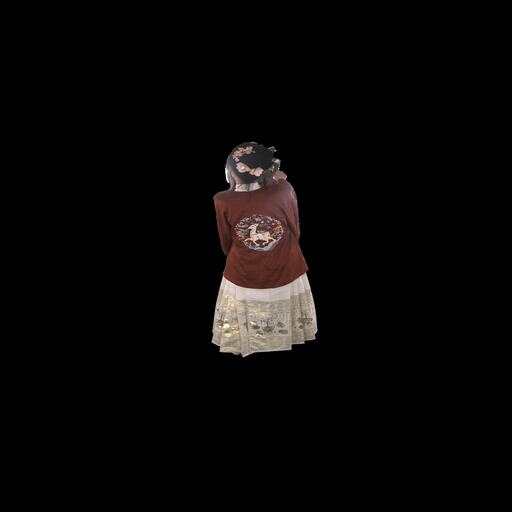} &
\includegraphics[width=\linewidth]{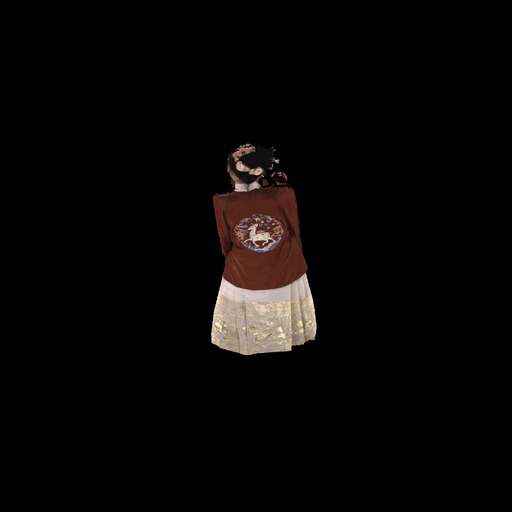} &
\includegraphics[width=\linewidth]{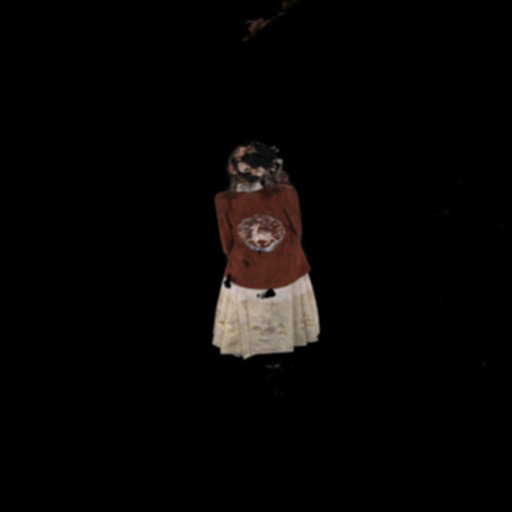} &
\includegraphics[width=\linewidth]{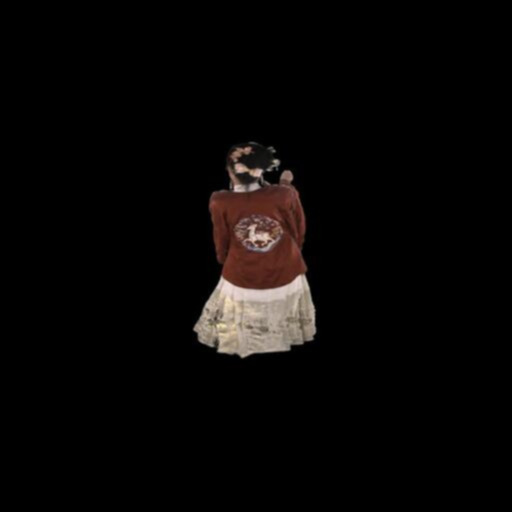} &
\includegraphics[width=\linewidth]{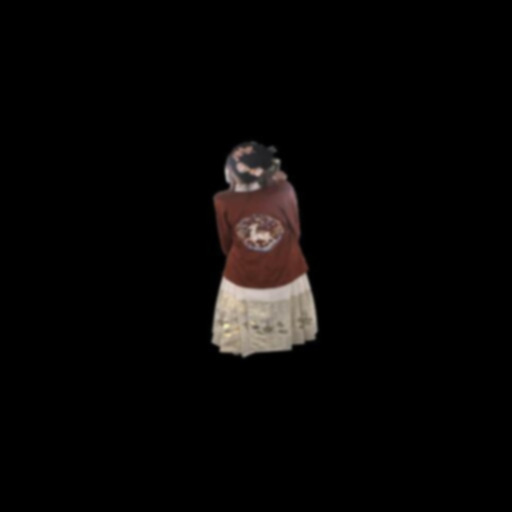} \\

\rotatebox{90}{\makecell{THuman2.1}} &
\includegraphics[width=\linewidth]{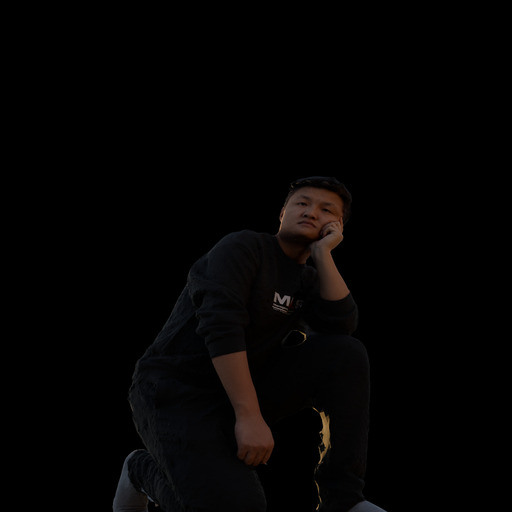} &
\includegraphics[width=\linewidth]{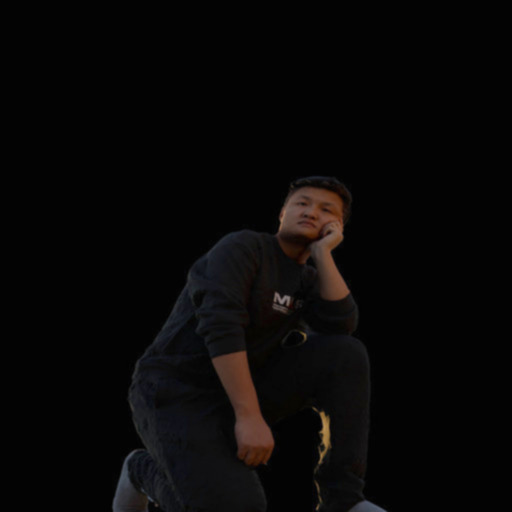} &
\includegraphics[width=\linewidth]{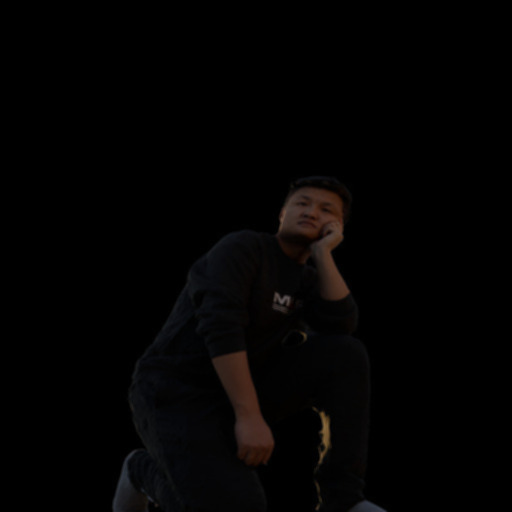} &
\includegraphics[width=\linewidth]{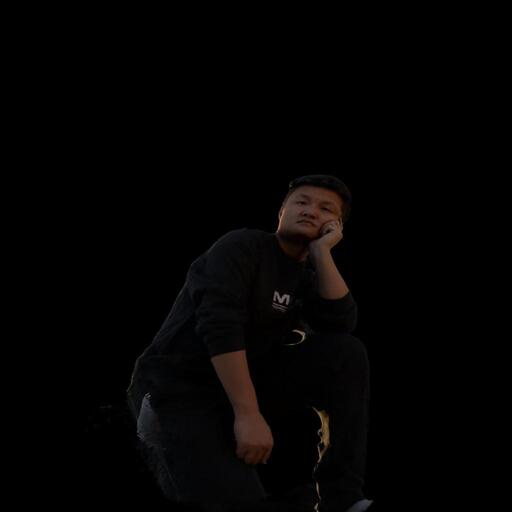} &
\includegraphics[width=\linewidth]{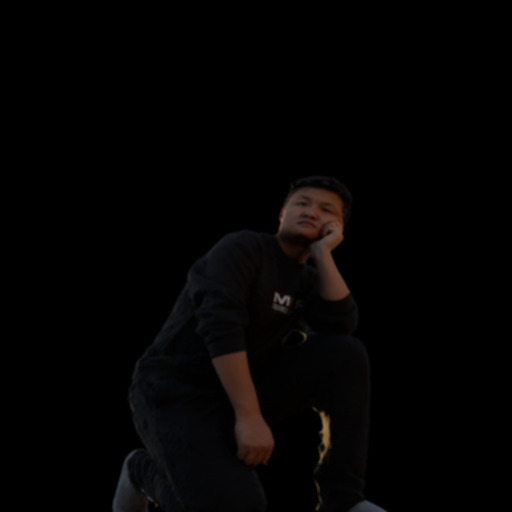} \\

\rotatebox{90}{\makecell{LLFF}} &
\includegraphics[width=\linewidth]{images/6_visual_ours/gt-llff-1.JPG} &
\begin{tikzpicture}
\fill[black] (0,0) rectangle (\linewidth,\linewidth);
\begin{scope}
\clip (0,0) rectangle (\linewidth,\linewidth);
\draw[gray!90, line width=4pt] (0,0) -- (\linewidth,\linewidth);
\draw[gray!90, line width=4pt] (0,\linewidth) -- (\linewidth,0);
\end{scope}
\node[text=white] at (0.5\linewidth,0.5\linewidth) {Not applicable};
\end{tikzpicture} &
\includegraphics[width=\linewidth]{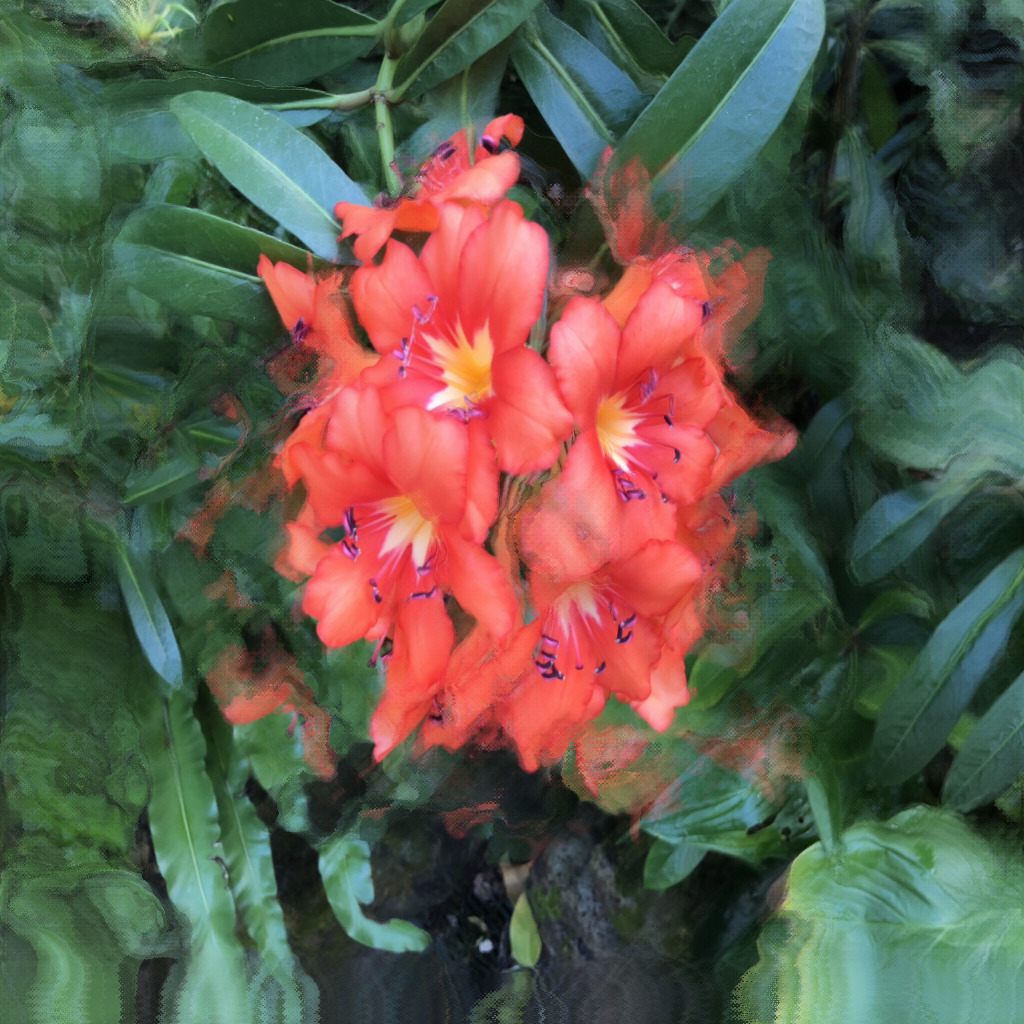} &
\includegraphics[width=\linewidth]{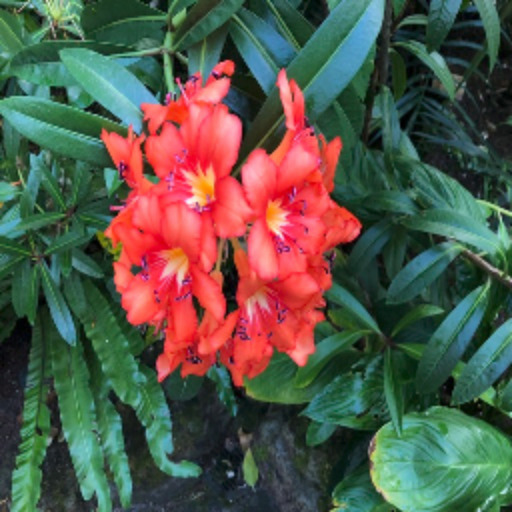} &
\includegraphics[width=\linewidth]{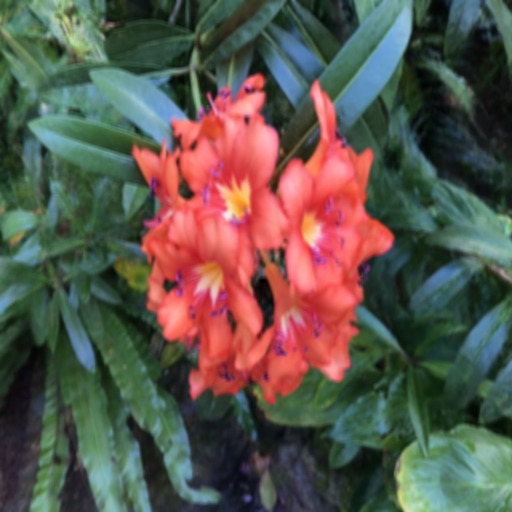} \\

\rotatebox{90}{\makecell{ZJU MoCap}} &
\includegraphics[width=\linewidth]{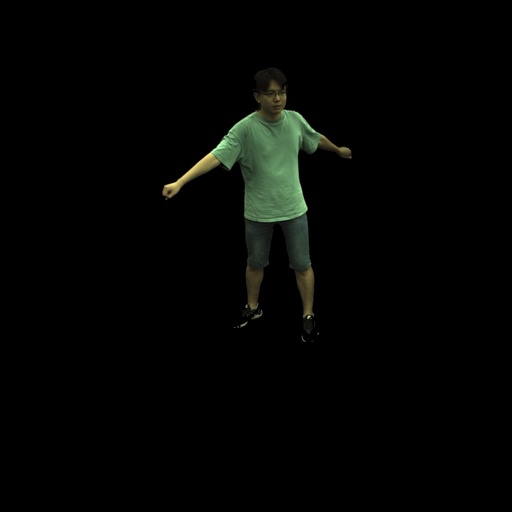} &
\includegraphics[width=\linewidth]{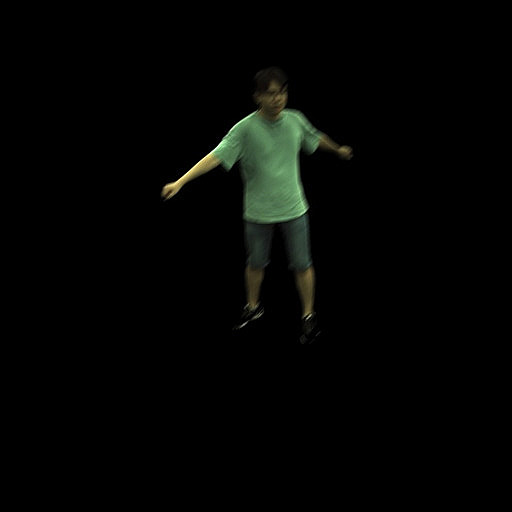} &
\includegraphics[width=\linewidth]{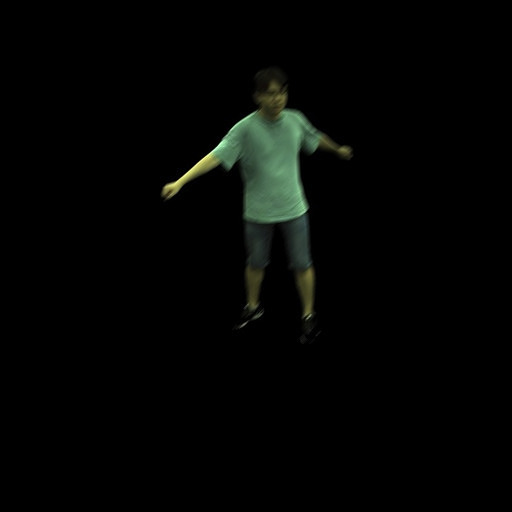} &
\includegraphics[width=\linewidth]{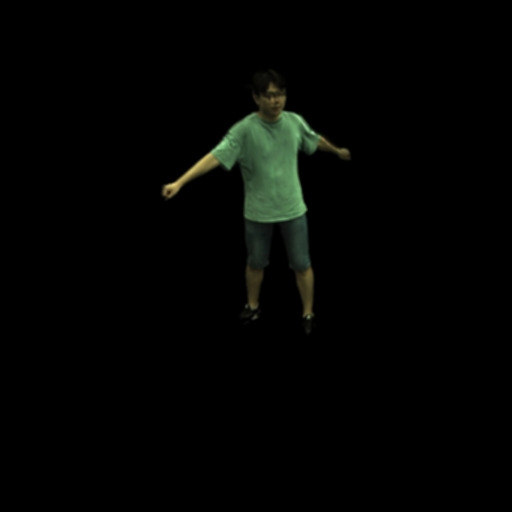} &
\includegraphics[width=\linewidth]{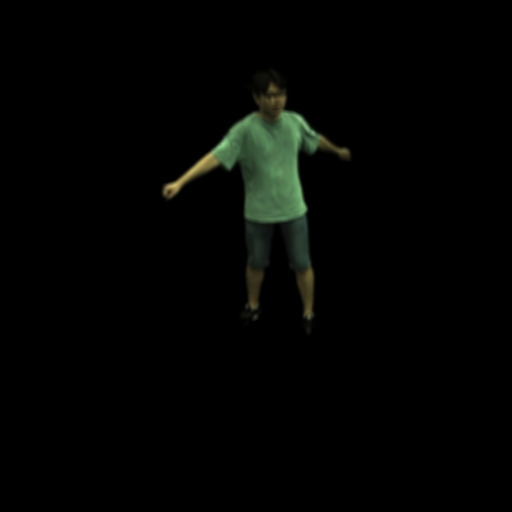} \\

\rotatebox{90}{\makecell{MVHuman}} &
\includegraphics[width=\linewidth]{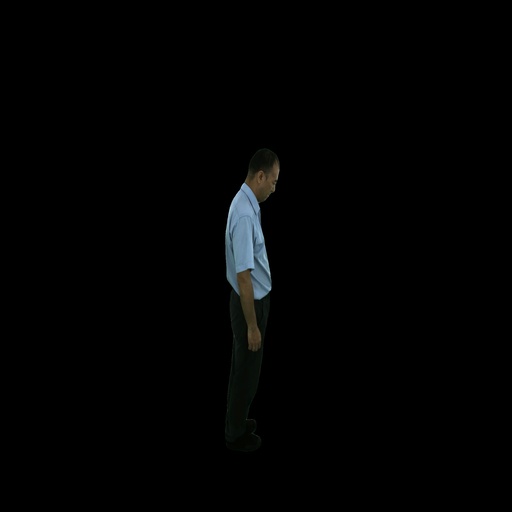} &
\includegraphics[width=\linewidth]{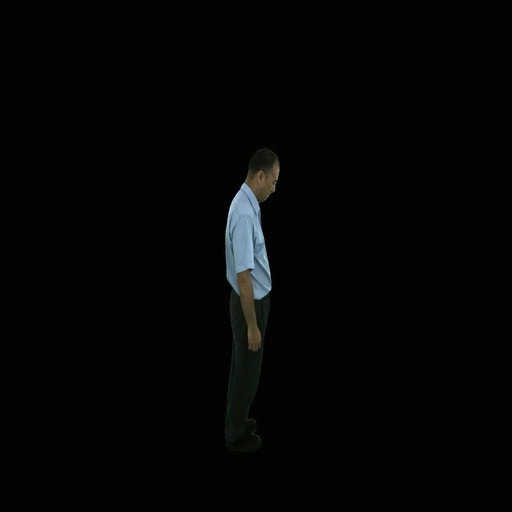} &
\includegraphics[width=\linewidth]{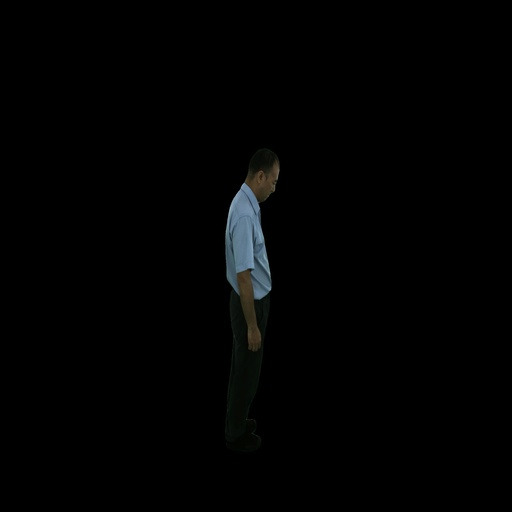} &
\includegraphics[width=\linewidth]{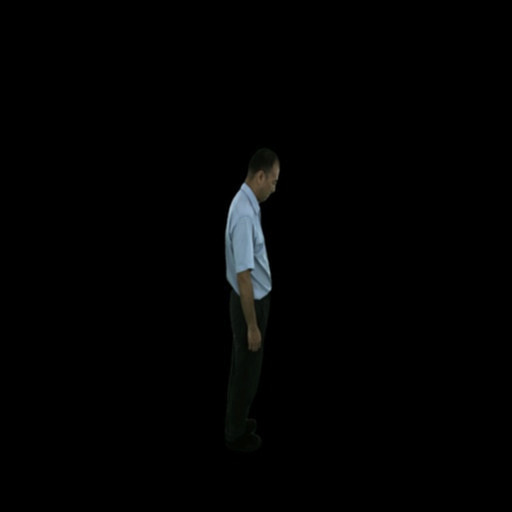} &
\includegraphics[width=\linewidth]{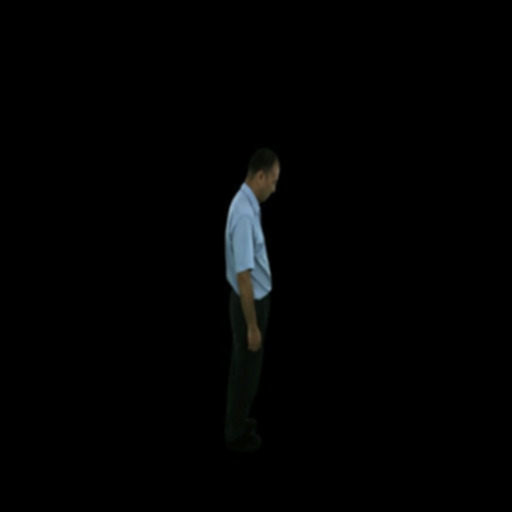} \\

\end{tabular}

\caption{Qualitative comparison of our method to online and offline reconstruction methods. (Part 1.1)}

\label{fig:visual-comparison-1-1}
\end{figure*}

\begin{figure*}[t]
\centering

\newcolumntype{M}[1]{>{\centering\arraybackslash}m{#1}}

\setlength{\tabcolsep}{1pt}
\renewcommand{\arraystretch}{0.5}
\begin{tabular}{M{0.03\linewidth} *{5}{M{0.185\linewidth}}}

& GT & Nerfacto-big & Splatfacto-big & FrugalNeRF & GPS-Gaussian \\

\rotatebox{90}{\makecell{RIFTCast}} &
\includegraphics[width=\linewidth]{images/6_visual_ours/gt-rift-1.jpg} &
\includegraphics[width=\linewidth]{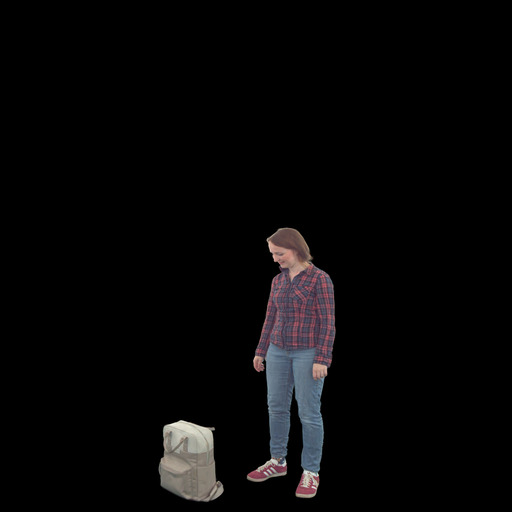} &
\includegraphics[width=\linewidth]{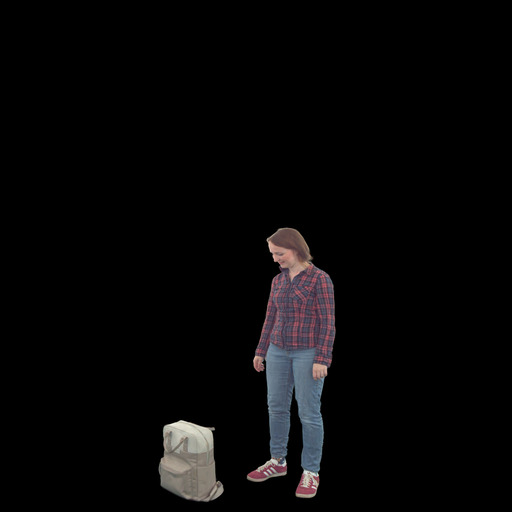} &
\includegraphics[width=\linewidth]{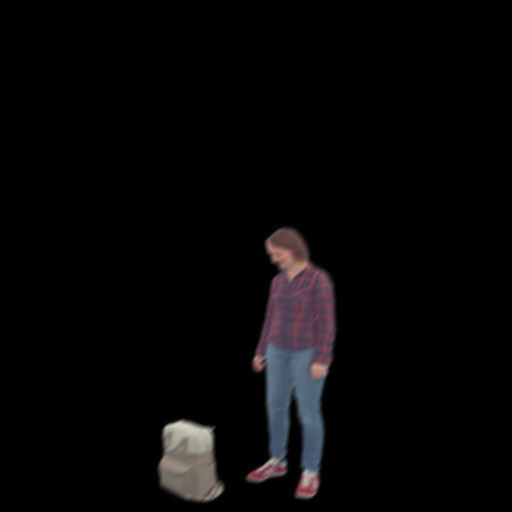} &
\includegraphics[width=\linewidth]{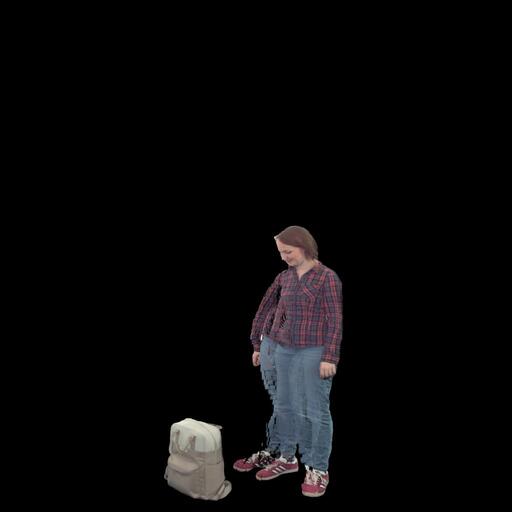} \\

\rotatebox{90}{\makecell{DNA}} &
\includegraphics[width=\linewidth]{images/6_visual_ours/gt-dna-1.jpg} &
\includegraphics[width=\linewidth]{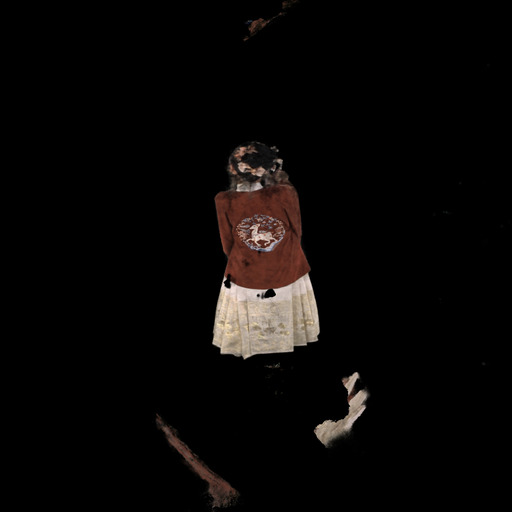} &
\includegraphics[width=\linewidth]{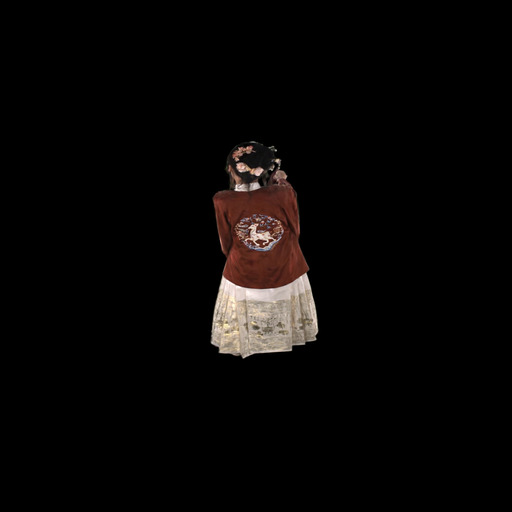} &
\includegraphics[width=\linewidth]{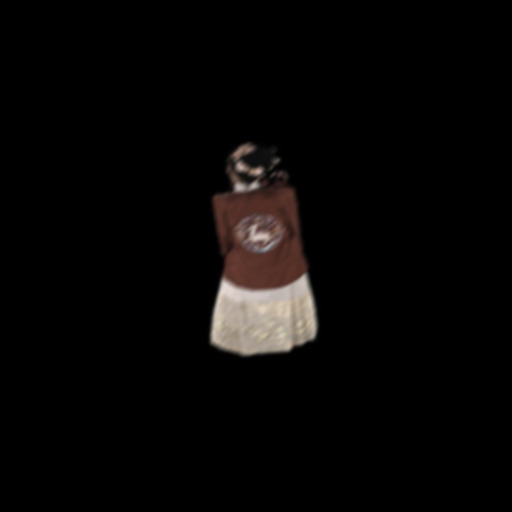} &
\includegraphics[width=\linewidth]{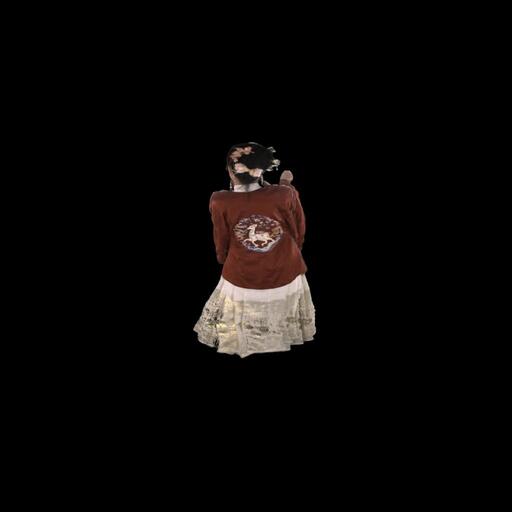} \\

\rotatebox{90}{\makecell{THuman2.1}} &
\includegraphics[width=\linewidth]{images/6_visual_ours/gt-thuman-1.jpg} &
\includegraphics[width=\linewidth]{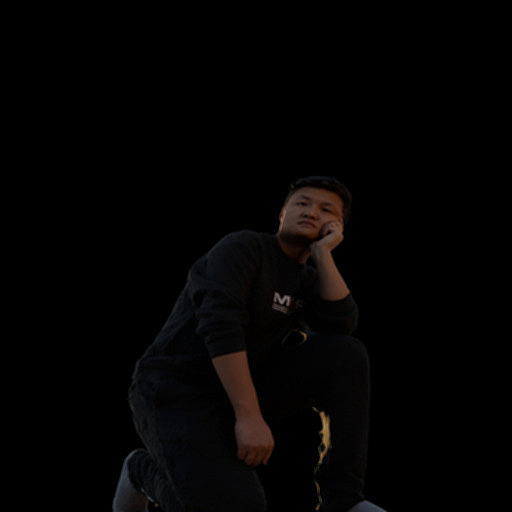} &
\includegraphics[width=\linewidth]{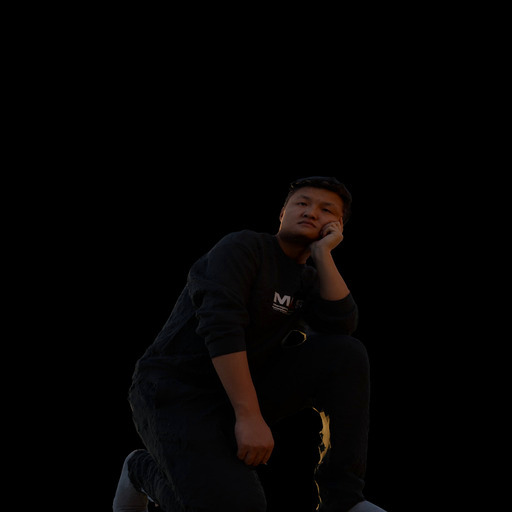} &
\includegraphics[width=\linewidth]{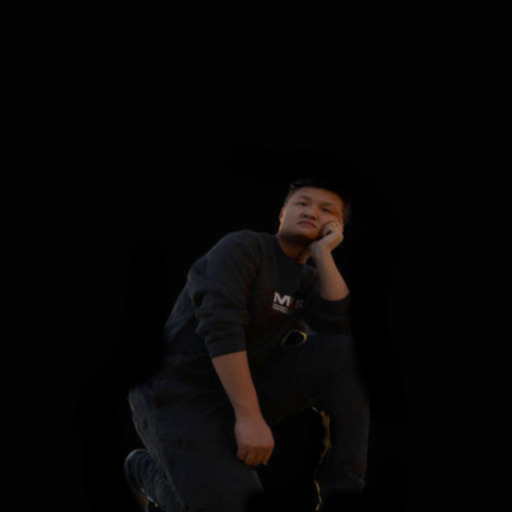} &
\includegraphics[width=\linewidth]{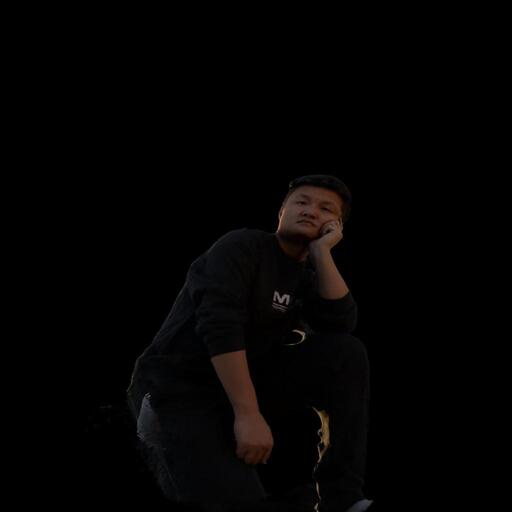} \\

\rotatebox{90}{\makecell{LLFF}} &
\includegraphics[width=\linewidth]{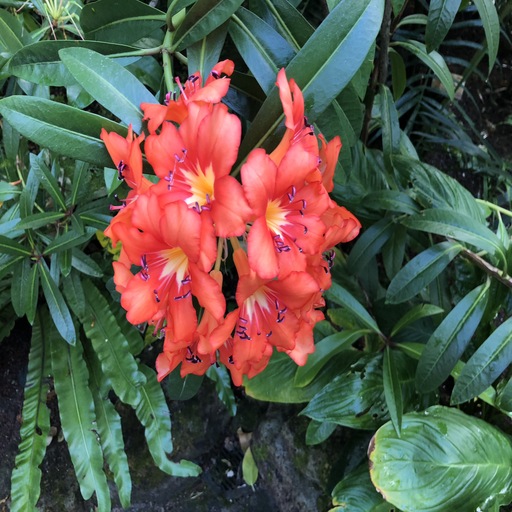} &
\includegraphics[width=\linewidth]{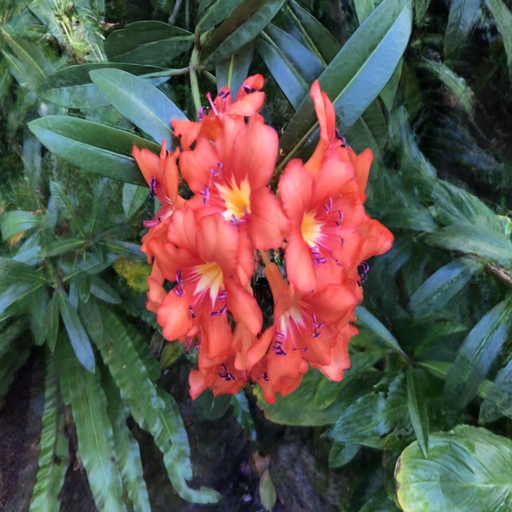} &
\includegraphics[width=\linewidth]{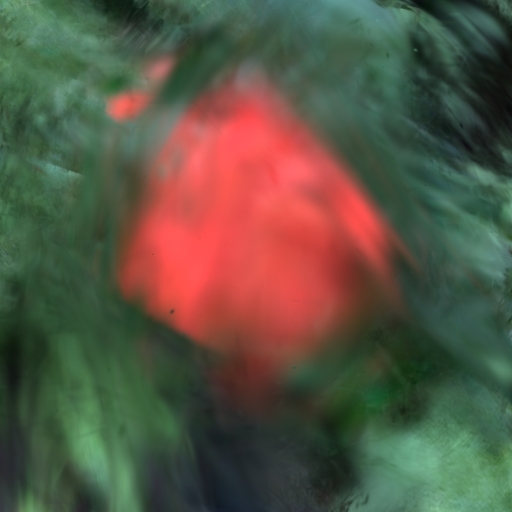} &
\includegraphics[width=\linewidth]{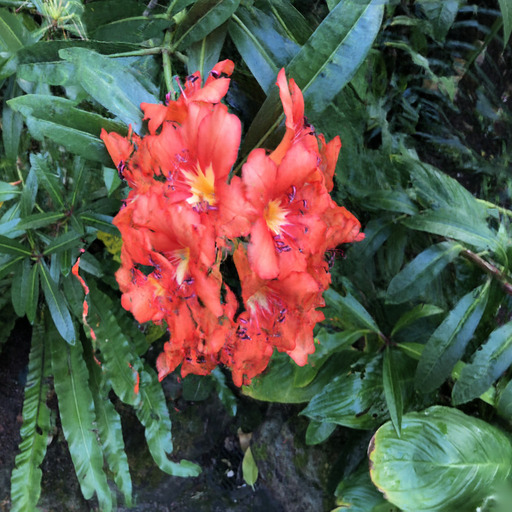} &
\begin{tikzpicture}
\fill[black] (0,0) rectangle (\linewidth,\linewidth);
\begin{scope}
\clip (0,0) rectangle (\linewidth,\linewidth);
\draw[gray!90, line width=4pt] (0,0) -- (\linewidth,\linewidth);
\draw[gray!90, line width=4pt] (0,\linewidth) -- (\linewidth,0);
\end{scope}
\node[text=white] at (0.5\linewidth,0.5\linewidth) {Not applicable};
\end{tikzpicture} \\

\rotatebox{90}{\makecell{ZJU MoCap}} &
\includegraphics[width=\linewidth]{images/6_visual_ours/gt-zju-1.jpg} &
\includegraphics[width=\linewidth]{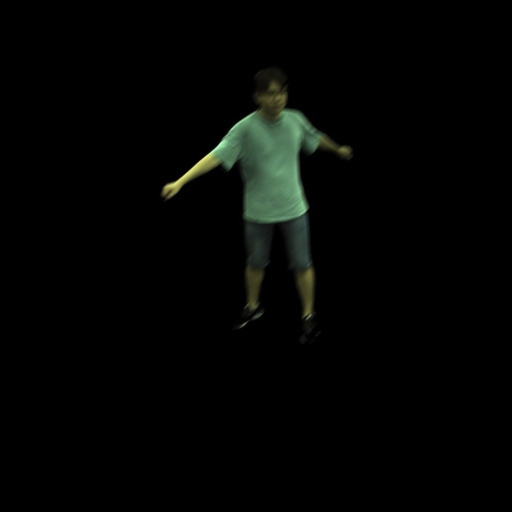} &
\includegraphics[width=\linewidth]{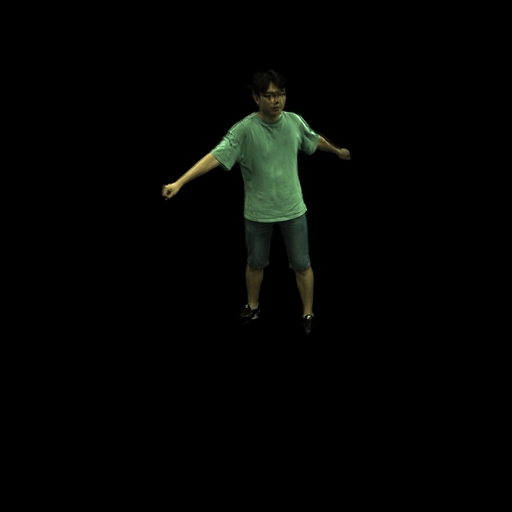} &
\includegraphics[width=\linewidth]{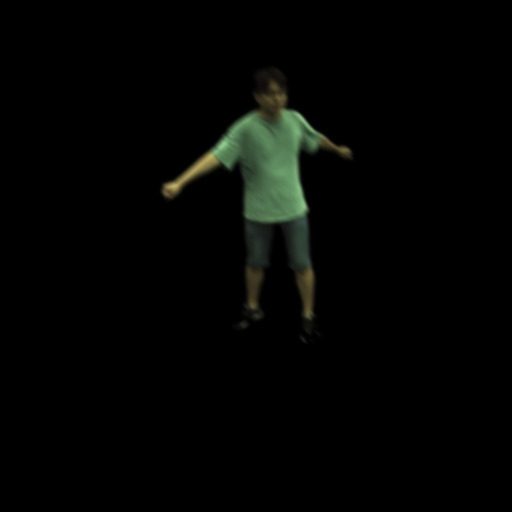} &
\includegraphics[width=\linewidth]{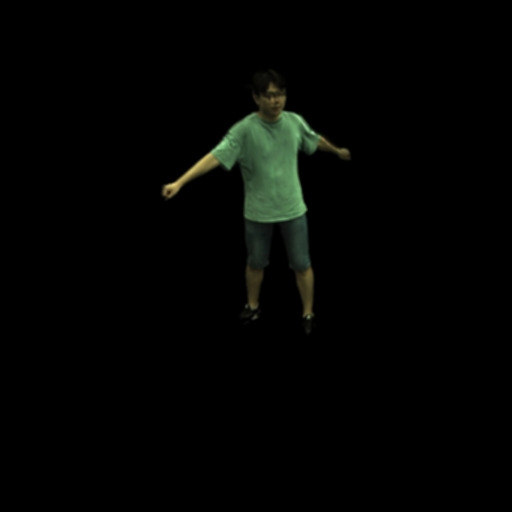} \\

\rotatebox{90}{\makecell{MVHuman}} &
\includegraphics[width=\linewidth]{images/6_visual_ours/gt-mvhuman-1.jpg} &
\includegraphics[width=\linewidth]{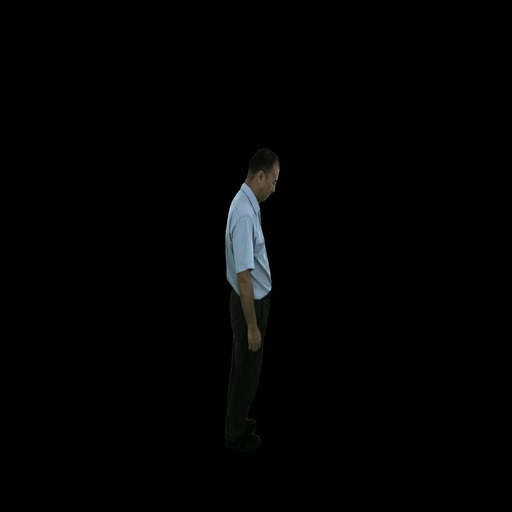} &
\includegraphics[width=\linewidth]{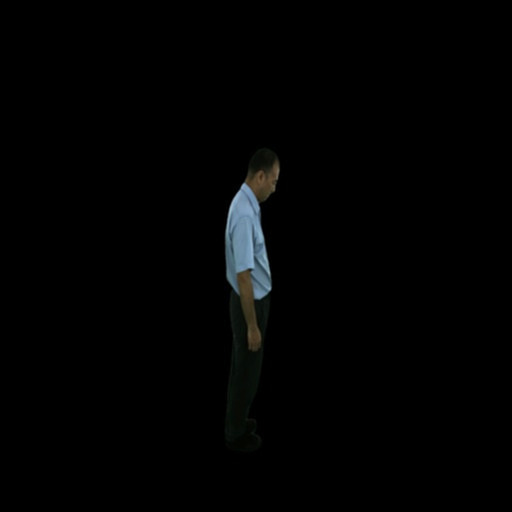} &
\includegraphics[width=\linewidth]{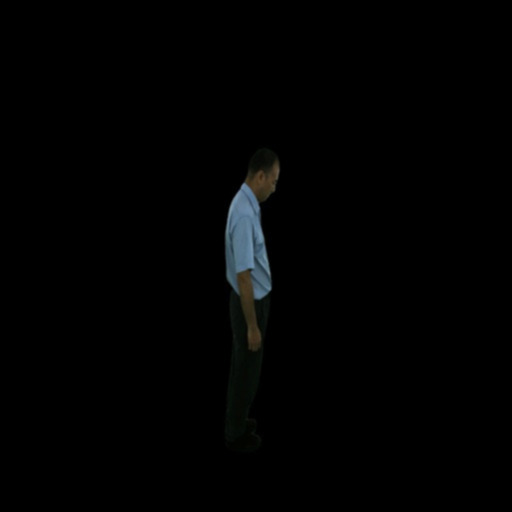} &
\includegraphics[width=\linewidth]{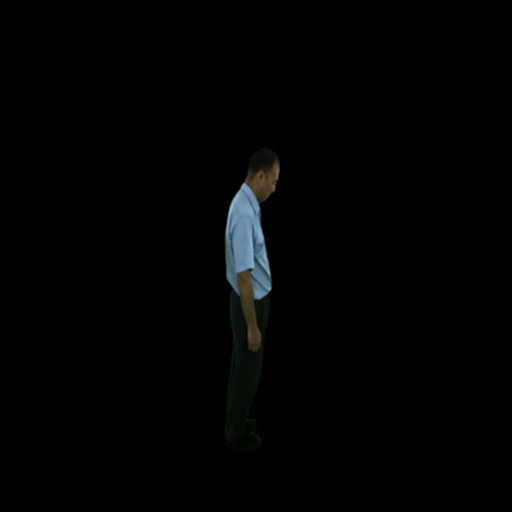} \\

\end{tabular}

\caption{Qualitative comparison of our method to online and offline reconstruction methods. (Part 1.2)}

\label{fig:visual-comparison-1-2}
\end{figure*}

\begin{figure*}[t]
\centering

\newcolumntype{M}[1]{>{\centering\arraybackslash}m{#1}}

\setlength{\tabcolsep}{1pt}
\renewcommand{\arraystretch}{0.5}
\begin{tabular}{M{0.03\linewidth} *{5}{M{0.185\linewidth}}}

& GT & RIFTCast & ENeRF & GPS-Gaussian+ & Ours \\

\rotatebox{90}{\makecell{RIFTCast}} &
\includegraphics[width=\linewidth]{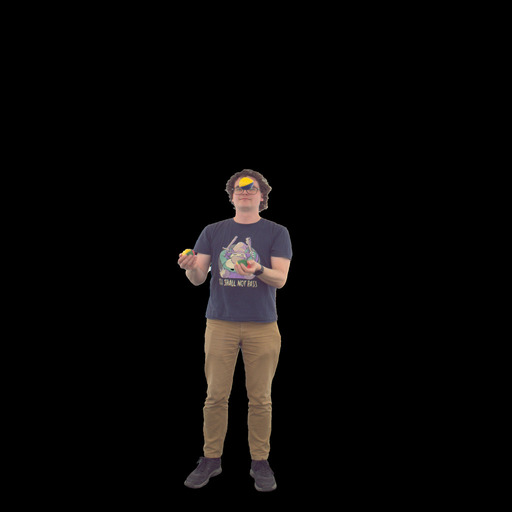} &
\includegraphics[width=\linewidth]{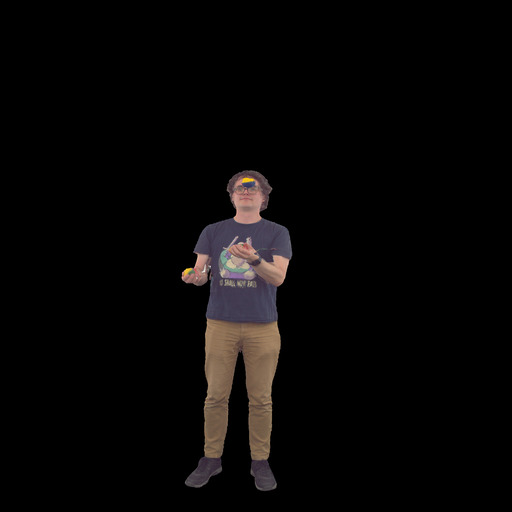} &
\includegraphics[width=\linewidth]{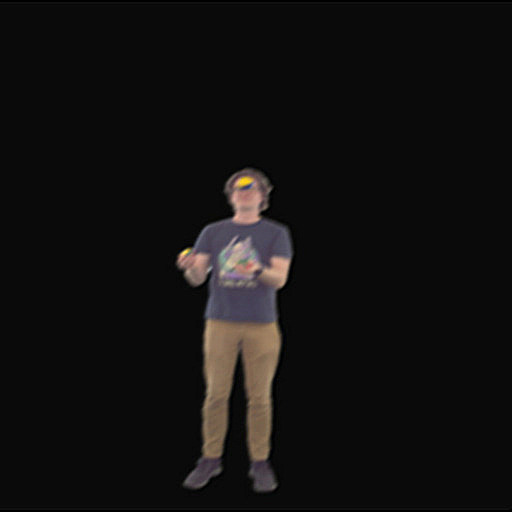} &
\includegraphics[width=\linewidth]{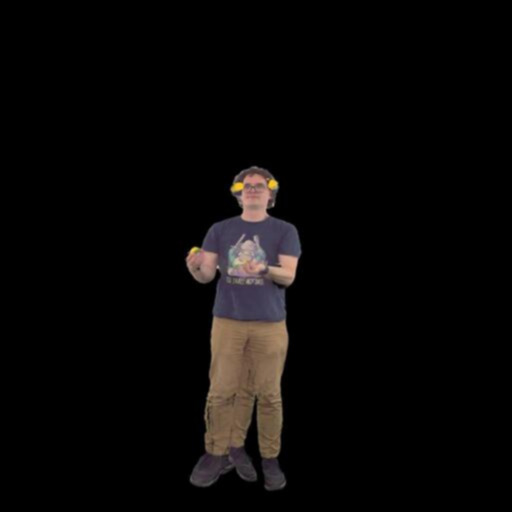} &
\includegraphics[width=\linewidth]{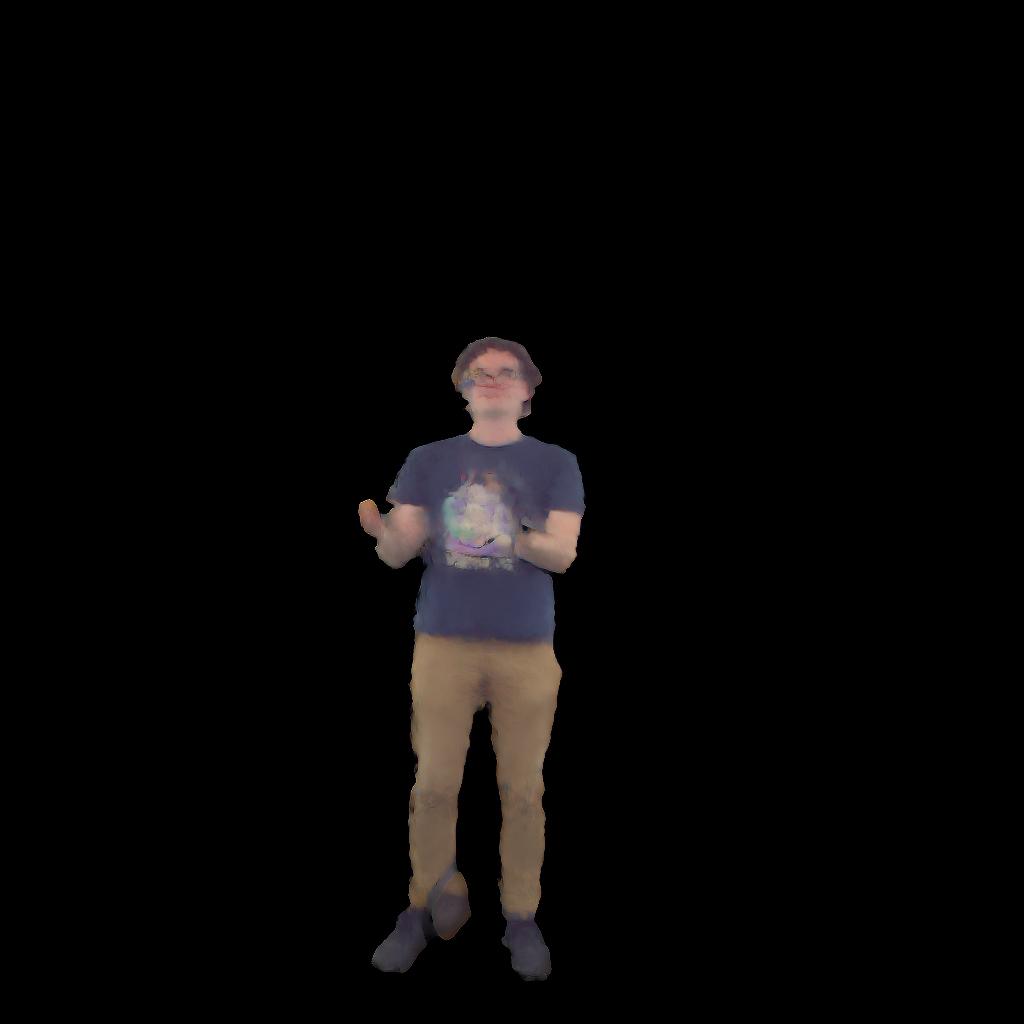} \\

\rotatebox{90}{\makecell{DNA}} &
\includegraphics[width=\linewidth]{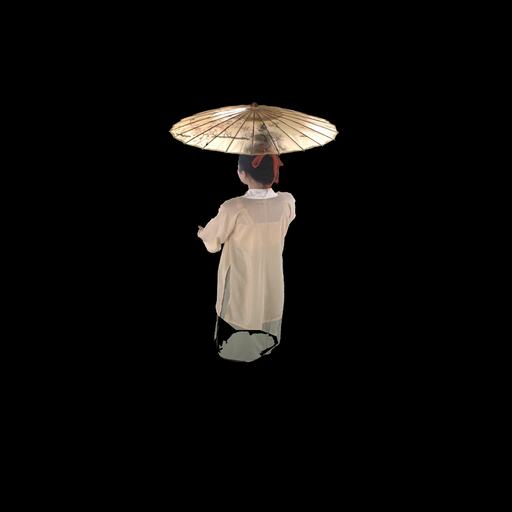} &
\includegraphics[width=\linewidth]{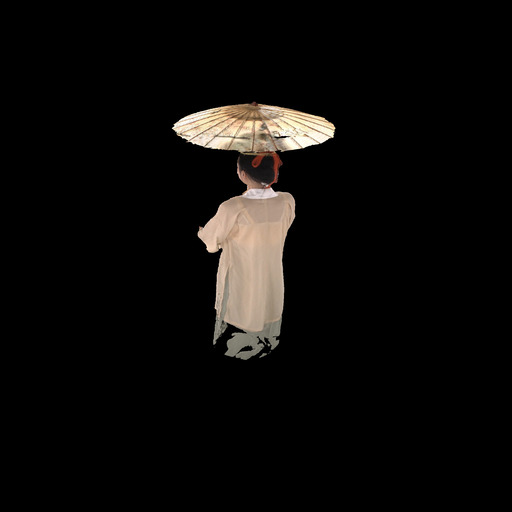} &
\includegraphics[width=\linewidth]{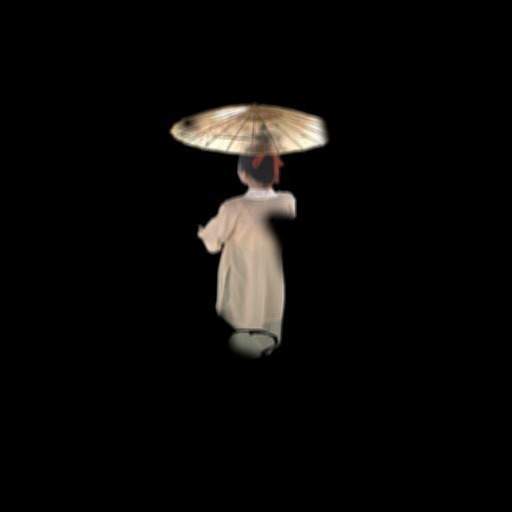} &
\includegraphics[width=\linewidth]{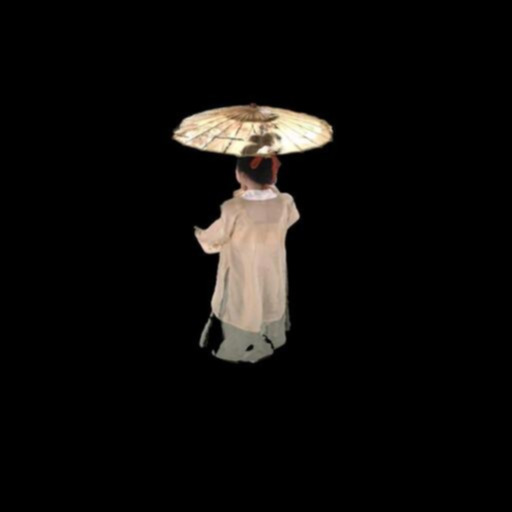} &
\includegraphics[width=\linewidth]{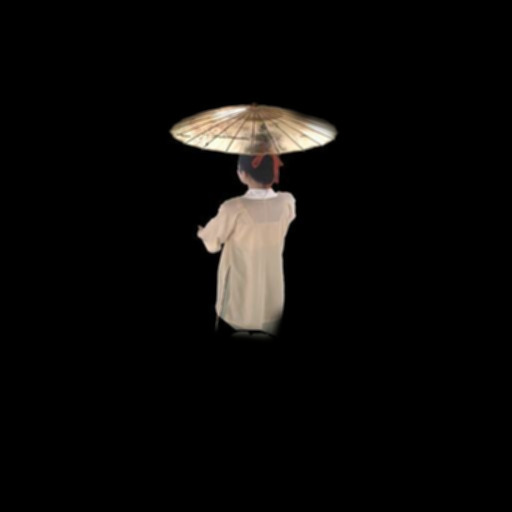} \\

\rotatebox{90}{\makecell{THuman2.1}} &
\includegraphics[width=\linewidth]{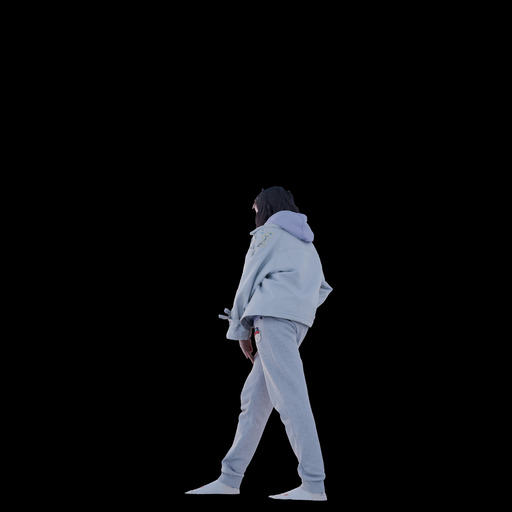} &
\includegraphics[width=\linewidth]{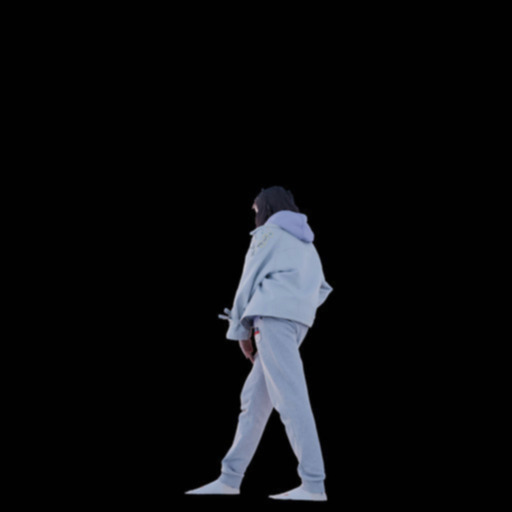} &
\includegraphics[width=\linewidth]{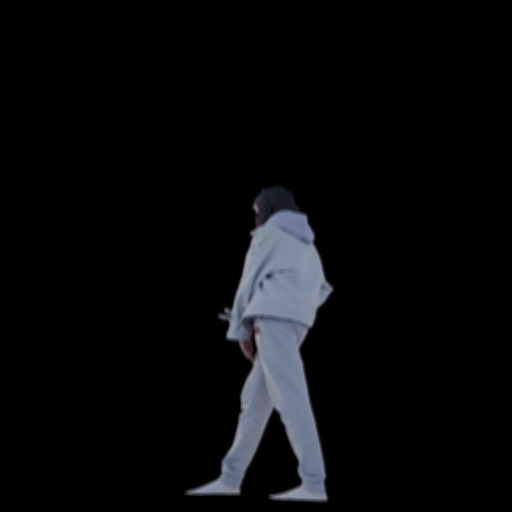} &
\includegraphics[width=\linewidth]{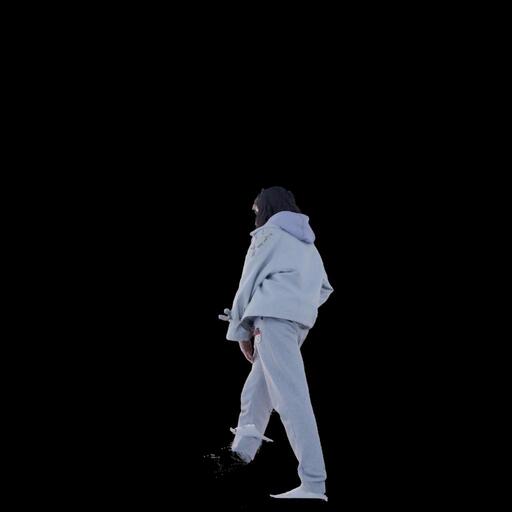} &
\includegraphics[width=\linewidth]{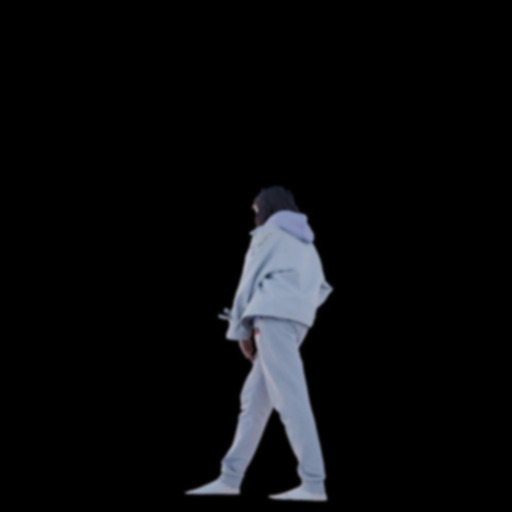} \\

\rotatebox{90}{\makecell{LLFF}} &
\includegraphics[width=\linewidth]{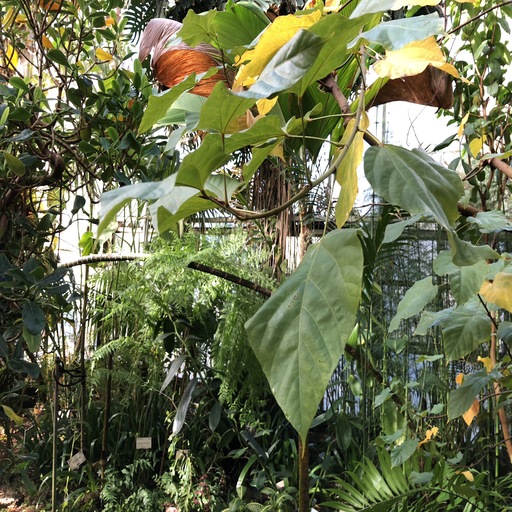} &
\begin{tikzpicture}
\fill[black] (0,0) rectangle (\linewidth,\linewidth);
\begin{scope}
\clip (0,0) rectangle (\linewidth,\linewidth);
\draw[gray!90, line width=4pt] (0,0) -- (\linewidth,\linewidth);
\draw[gray!90, line width=4pt] (0,\linewidth) -- (\linewidth,0);
\end{scope}
\node[text=white] at (0.5\linewidth,0.5\linewidth) {Not applicable};
\end{tikzpicture} &
\includegraphics[width=\linewidth]{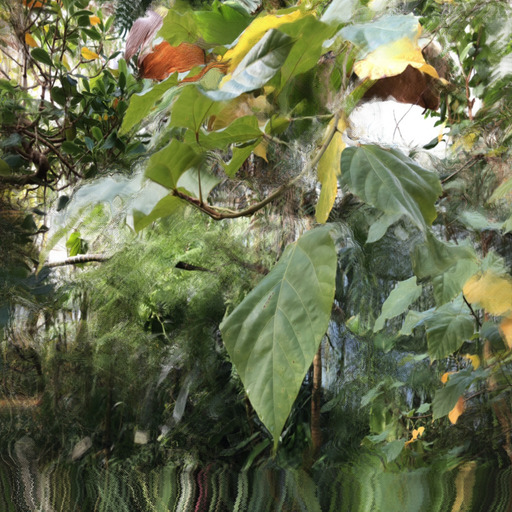} &
\includegraphics[width=\linewidth]{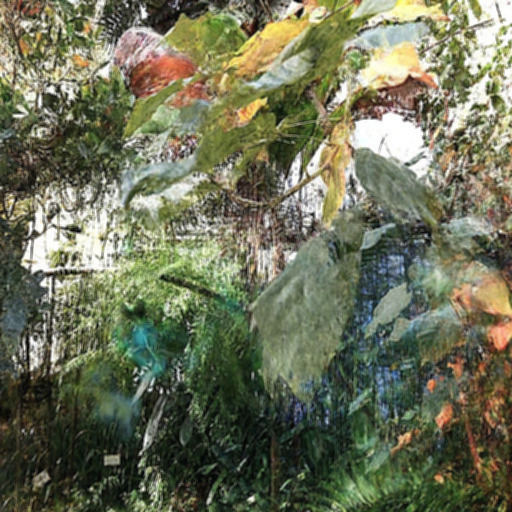} &
\includegraphics[width=\linewidth]{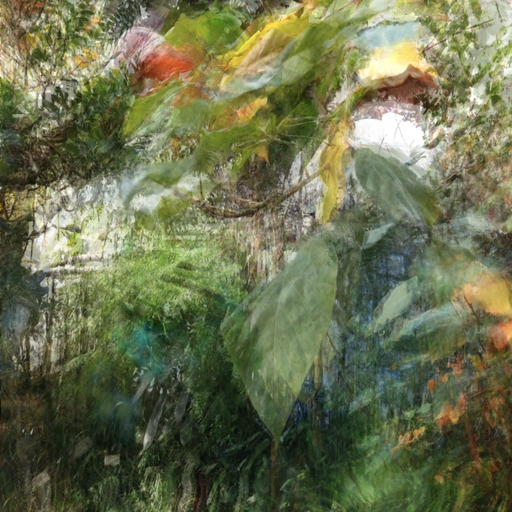} \\

\rotatebox{90}{\makecell{ZJU MoCap}} &
\includegraphics[width=\linewidth]{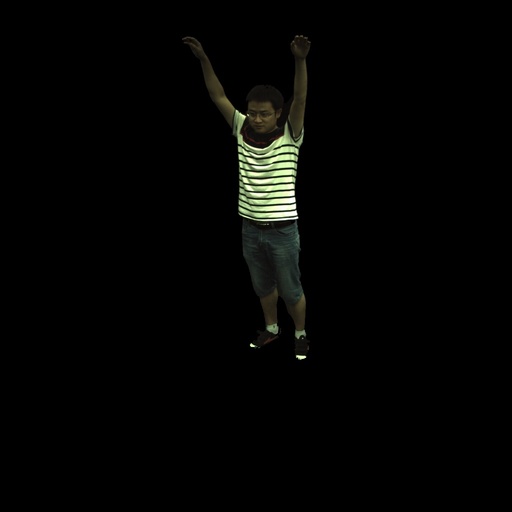} &
\includegraphics[width=\linewidth]{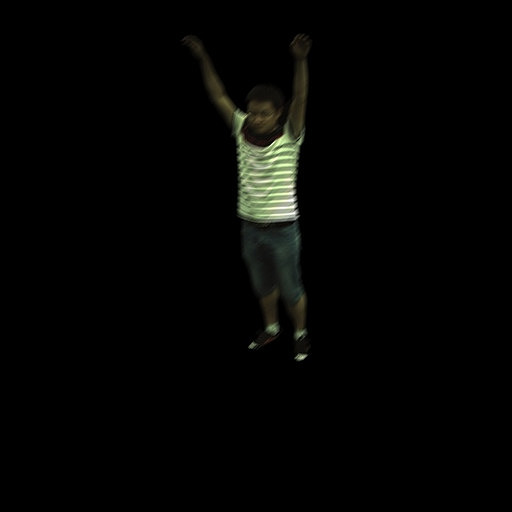} &
\includegraphics[width=\linewidth]{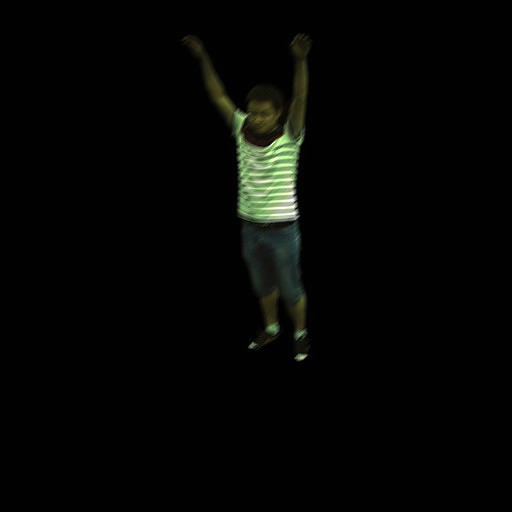} &
\includegraphics[width=\linewidth]{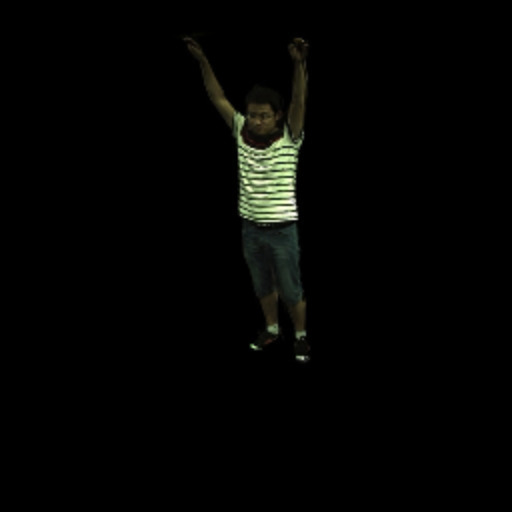} &
\includegraphics[width=\linewidth]{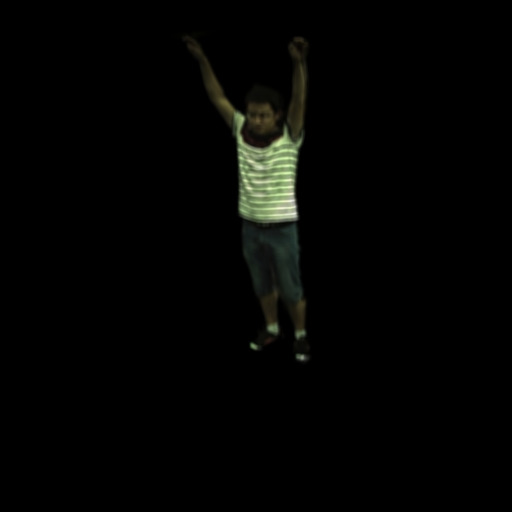} \\

\rotatebox{90}{\makecell{MVHuman}} &
\includegraphics[width=\linewidth]{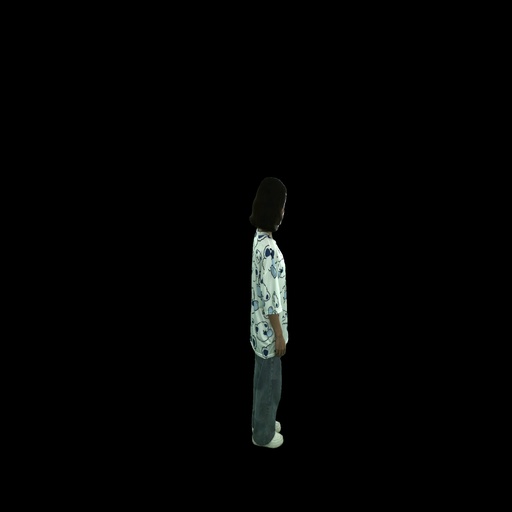} &
\includegraphics[width=\linewidth]{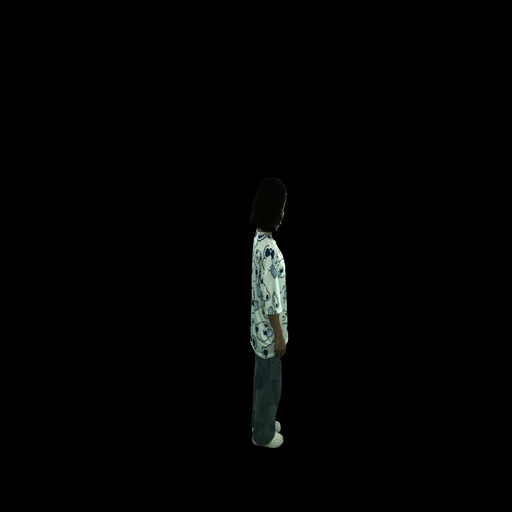} &
\includegraphics[width=\linewidth]{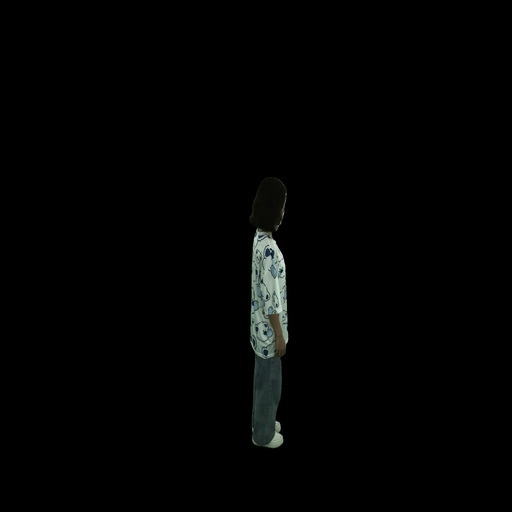} &
\includegraphics[width=\linewidth]{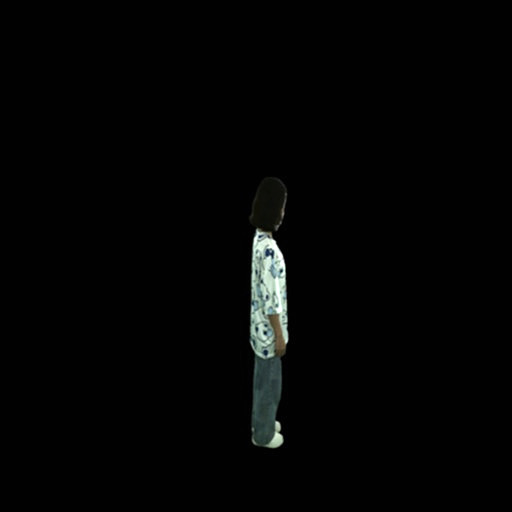} &
\includegraphics[width=\linewidth]{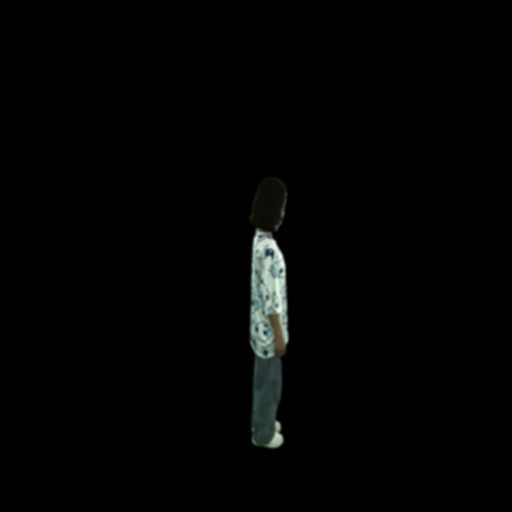} \\

\end{tabular}

\caption{Qualitative comparison of our method to online and offline reconstruction methods. (Part 2.1)}

\label{fig:visual-comparison-2-1}
\end{figure*}

\begin{figure*}[t]
\centering

\newcolumntype{M}[1]{>{\centering\arraybackslash}m{#1}}

\setlength{\tabcolsep}{1pt}
\renewcommand{\arraystretch}{0.5}
\begin{tabular}{M{0.03\linewidth} *{5}{M{0.185\linewidth}}}

& GT & Nerfacto-big & Splatfacto-big & FrugalNeRF & GPS-Gaussian \\

\rotatebox{90}{\makecell{RIFTCast}} &
\includegraphics[width=\linewidth]{images/6_visual_ours/gt-rift-2.jpg} &
\includegraphics[width=\linewidth]{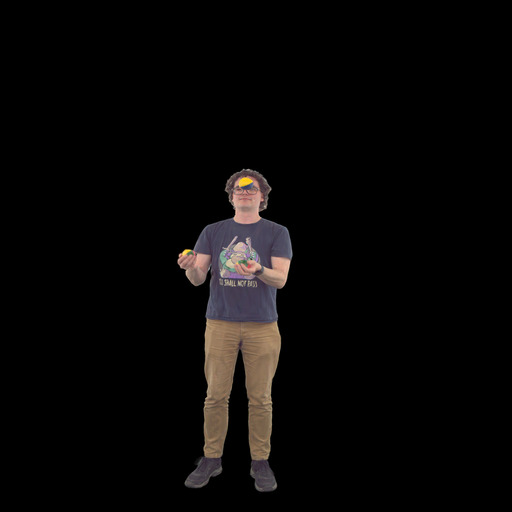} &
\includegraphics[width=\linewidth]{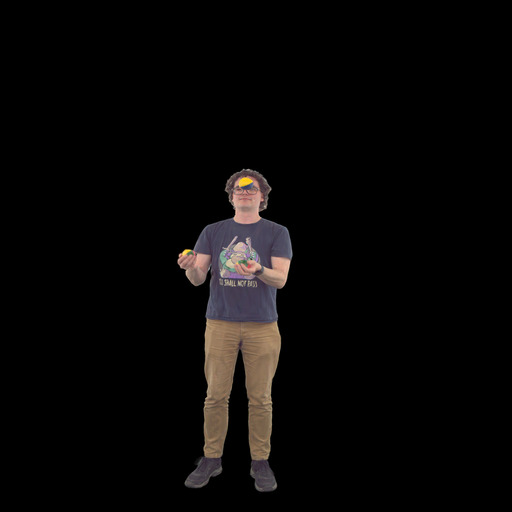} &
\includegraphics[width=\linewidth]{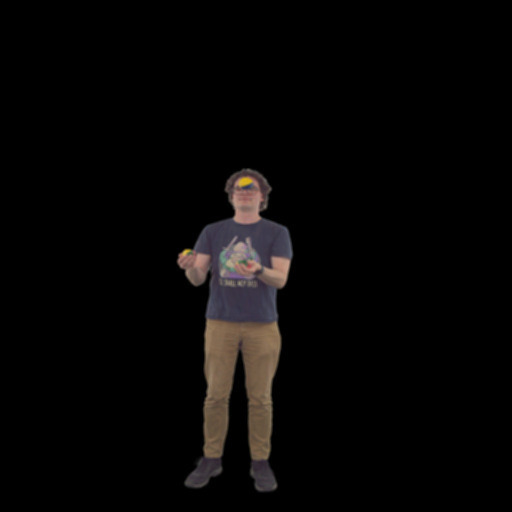} &
\includegraphics[width=\linewidth]{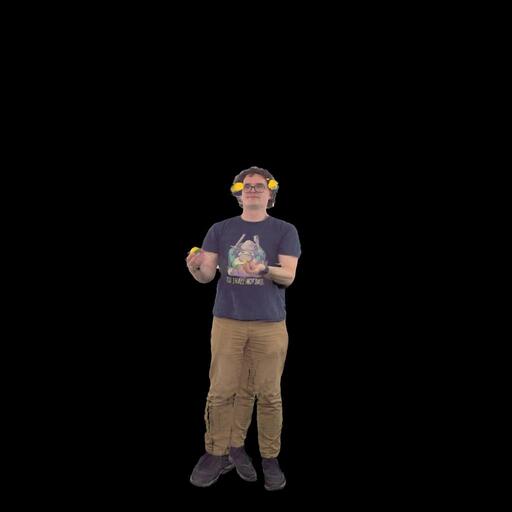} \\

\rotatebox{90}{\makecell{DNA}} &
\includegraphics[width=\linewidth]{images/6_visual_ours/gt-dna-2.jpg} &
\includegraphics[width=\linewidth]{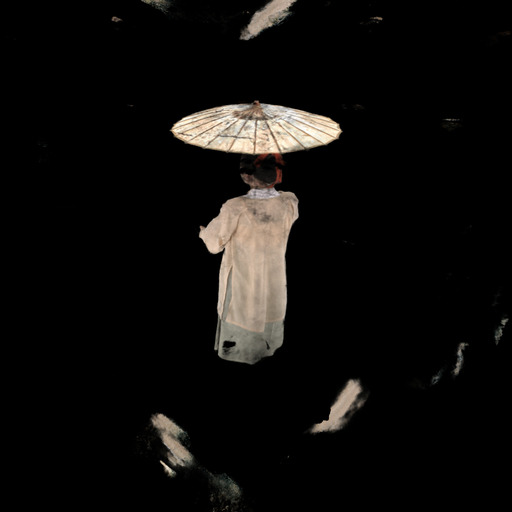} &
\includegraphics[width=\linewidth]{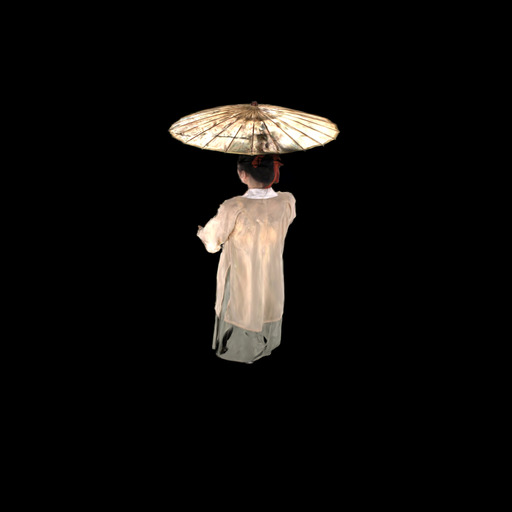} &
\includegraphics[width=\linewidth]{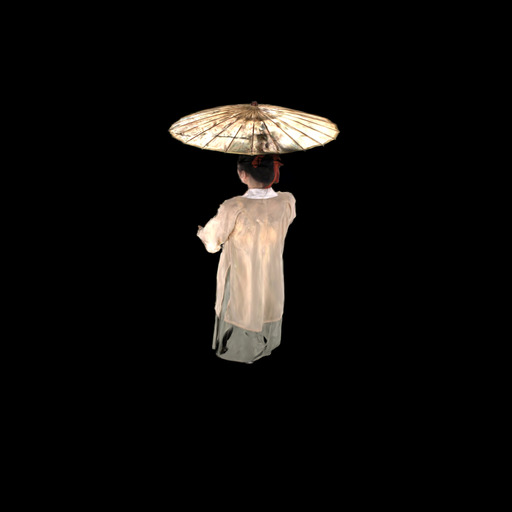} &
\includegraphics[width=\linewidth]{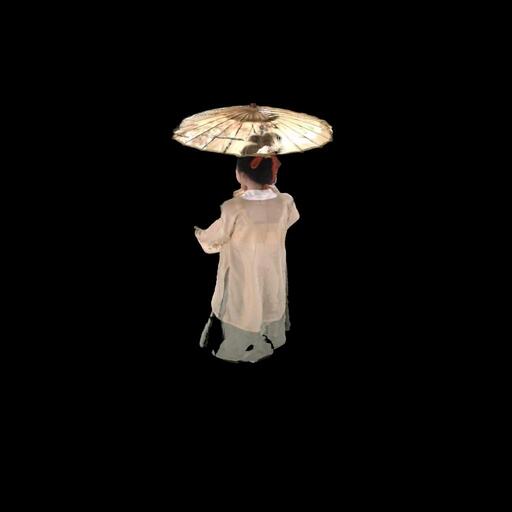} \\

\rotatebox{90}{\makecell{THuman2.1}} &
\includegraphics[width=\linewidth]{images/6_visual_ours/gt-thuman-2.jpg} &
\includegraphics[width=\linewidth]{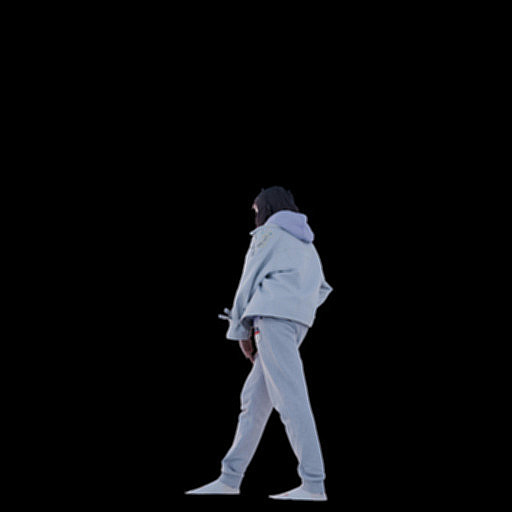} &
\includegraphics[width=\linewidth]{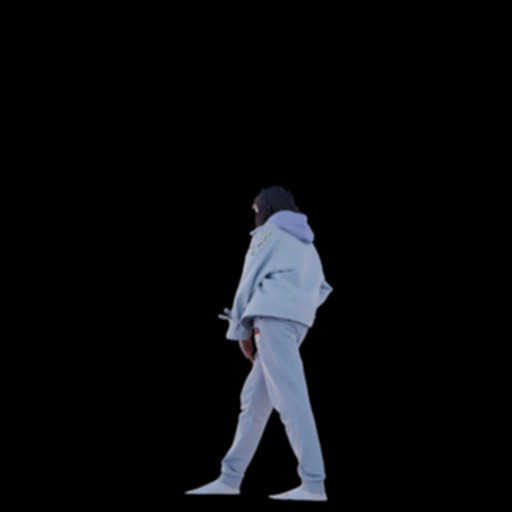} &
\includegraphics[width=\linewidth]{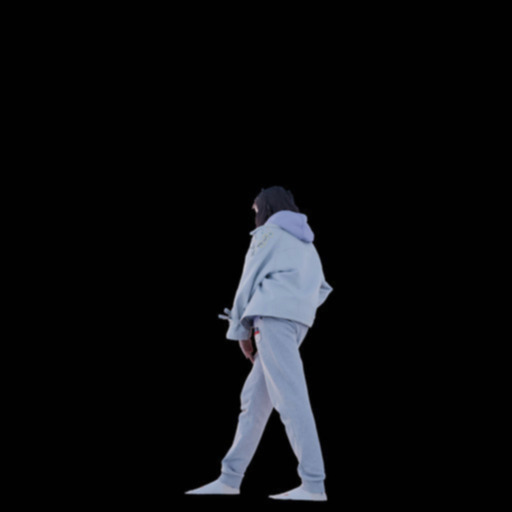} &
\includegraphics[width=\linewidth]{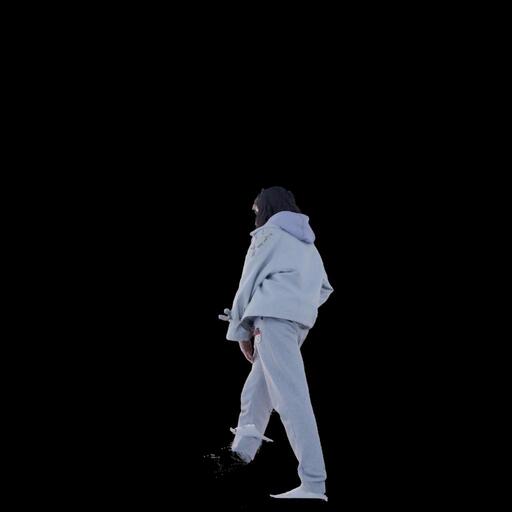} \\

\rotatebox{90}{\makecell{LLFF}} &
\includegraphics[width=\linewidth]{images/6_visual_ours/gt-llff-2.jpg} &
\includegraphics[width=\linewidth]{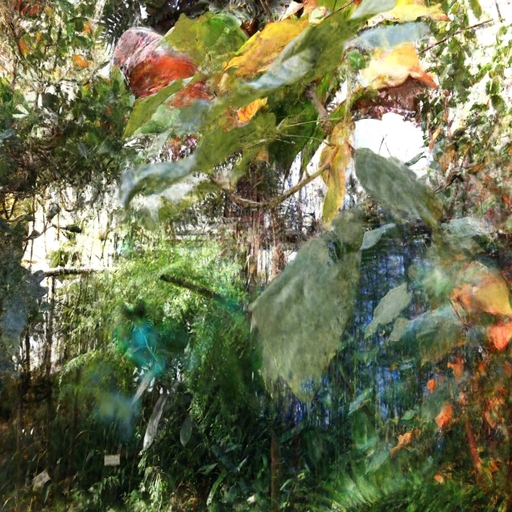} &
\includegraphics[width=\linewidth]{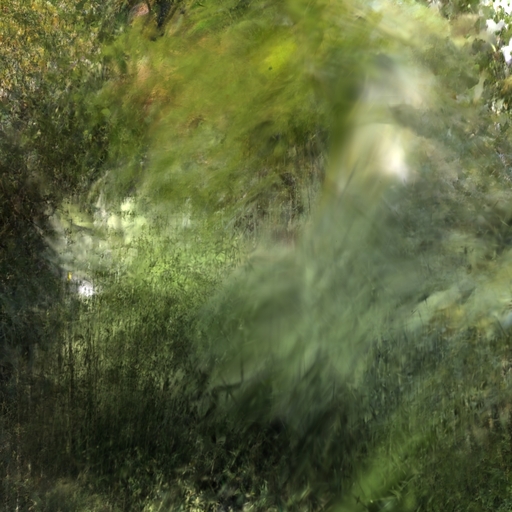} &
\includegraphics[width=\linewidth]{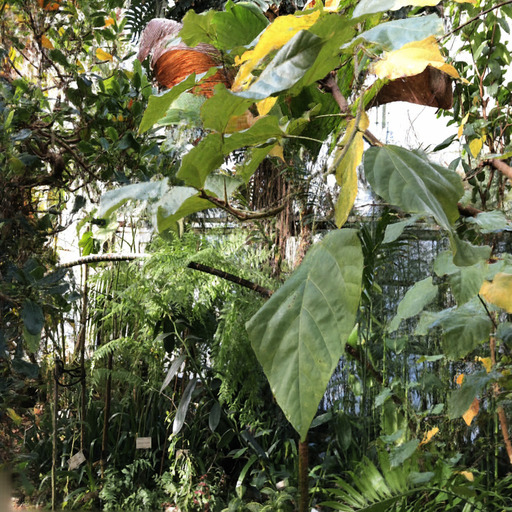} &
\begin{tikzpicture}
\fill[black] (0,0) rectangle (\linewidth,\linewidth);
\begin{scope}
\clip (0,0) rectangle (\linewidth,\linewidth);
\draw[gray!90, line width=4pt] (0,0) -- (\linewidth,\linewidth);
\draw[gray!90, line width=4pt] (0,\linewidth) -- (\linewidth,0);
\end{scope}
\node[text=white] at (0.5\linewidth,0.5\linewidth) {Not applicable};
\end{tikzpicture} \\

\rotatebox{90}{\makecell{ZJU MoCap}} &
\includegraphics[width=\linewidth]{images/6_visual_ours/gt-zju-2.jpg} &
\includegraphics[width=\linewidth]{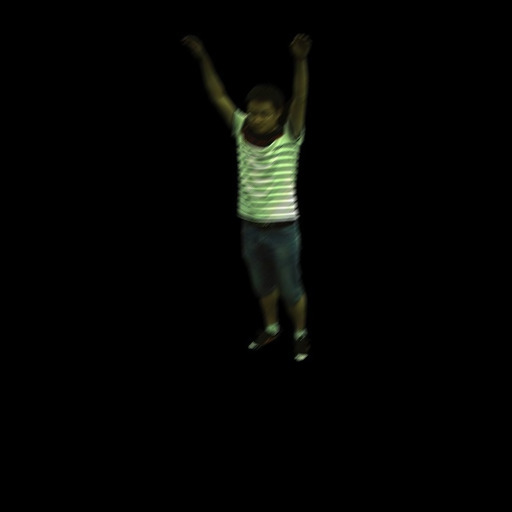} &
\includegraphics[width=\linewidth]{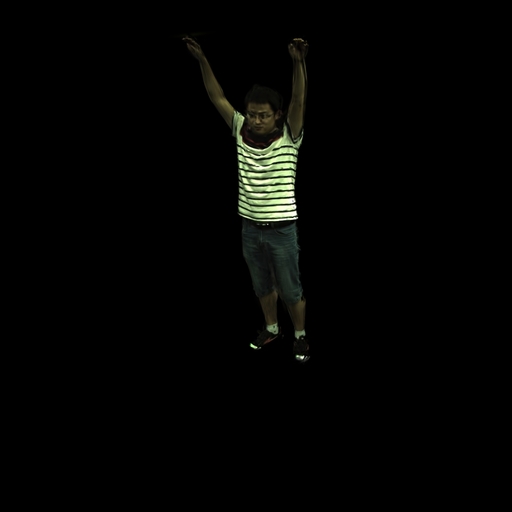} &
\includegraphics[width=\linewidth]{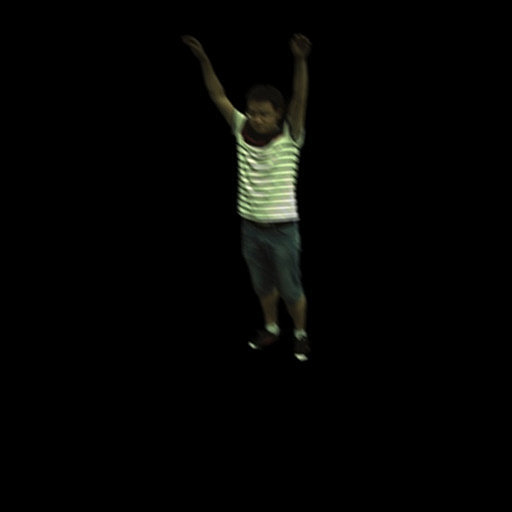} &
\includegraphics[width=\linewidth]{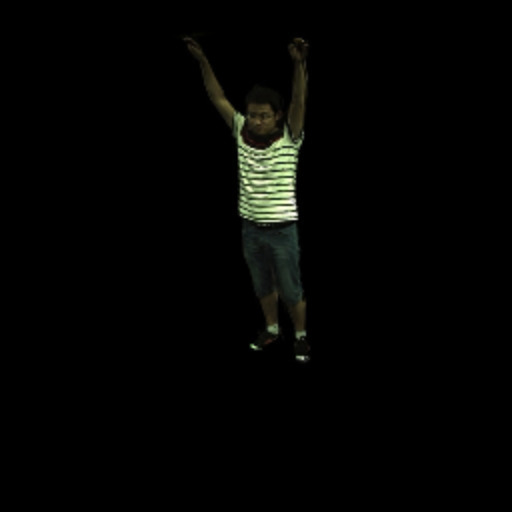} \\

\rotatebox{90}{\makecell{MVHuman}} &
\includegraphics[width=\linewidth]{images/6_visual_ours/gt-mvhuman-2.jpg} &
\includegraphics[width=\linewidth]{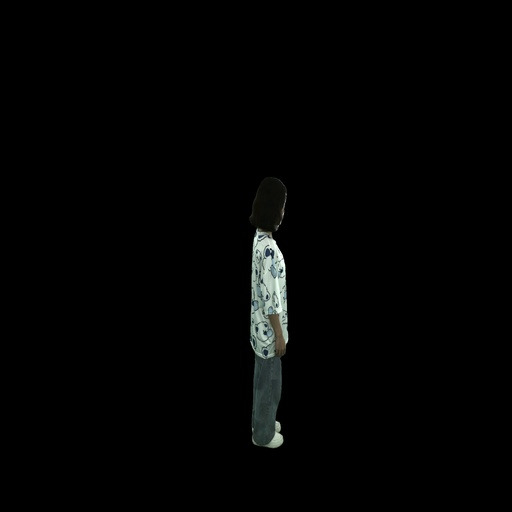} &
\includegraphics[width=\linewidth]{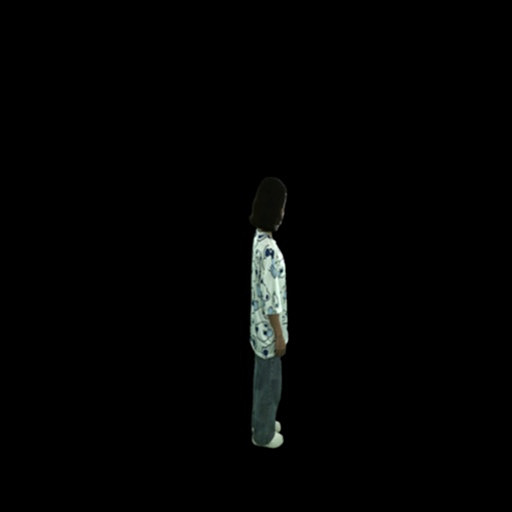} &
\includegraphics[width=\linewidth]{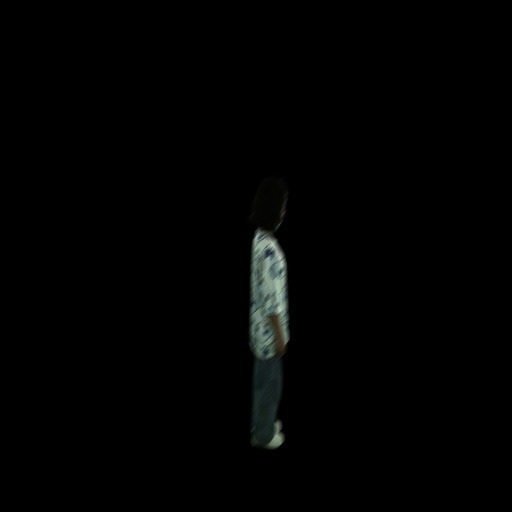} &
\includegraphics[width=\linewidth]{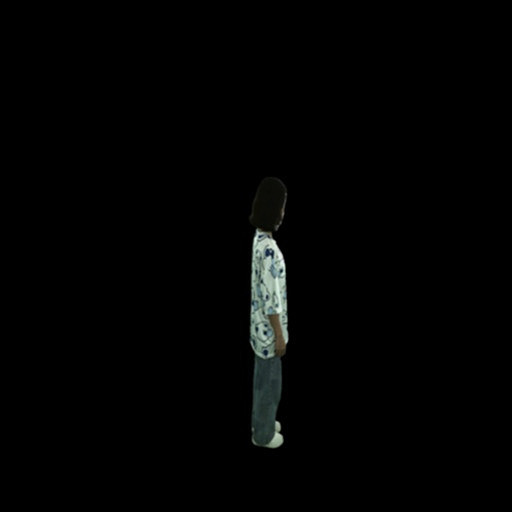} \\

\end{tabular}

\caption{Qualitative comparison of our method to online and offline reconstruction methods. (Part 2.2)}

\label{fig:visual-comparison-2-2}
\end{figure*}

\subsection{Ablation Evaluations}

\begin{figure*}[t]
\centering

\newcolumntype{M}[1]{>{\centering\arraybackslash}m{#1}}

\setlength{\tabcolsep}{1pt}
\renewcommand{\arraystretch}{0.5}
\begin{tabular}{M{0.08\linewidth} *{4}{M{0.185\linewidth}}}

& DNA 1 & DNA 2 & RIFTCast 1 & RIFTCast 2 \\

\rotatebox{90}{\makecell{GT}} &
\includegraphics[width=\linewidth]{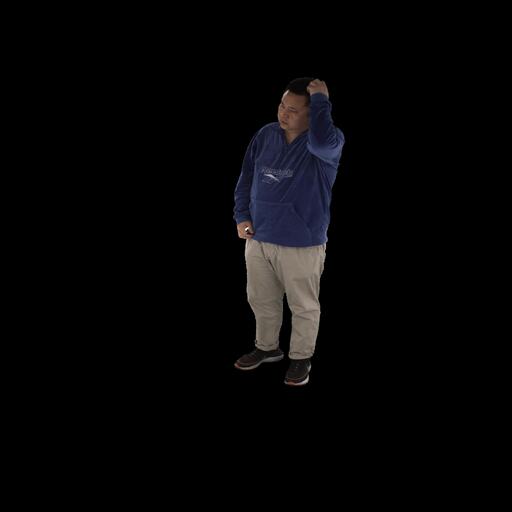} &
\includegraphics[width=\linewidth]{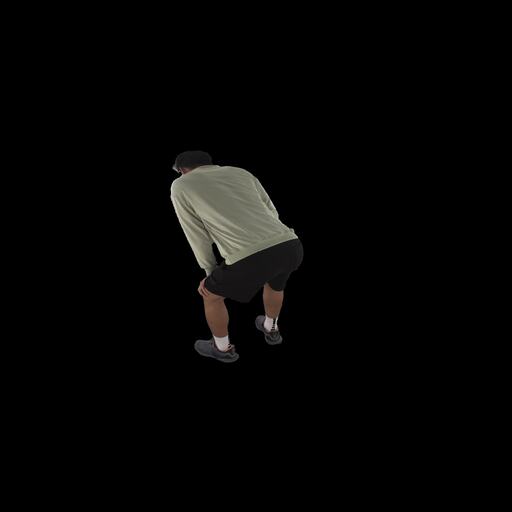} &
\includegraphics[width=\linewidth]{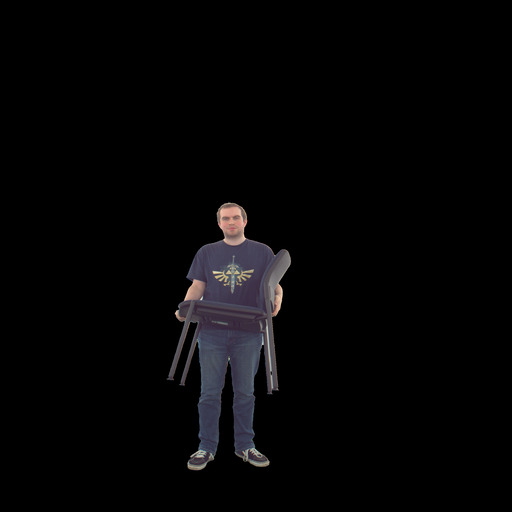} &
\includegraphics[width=\linewidth]{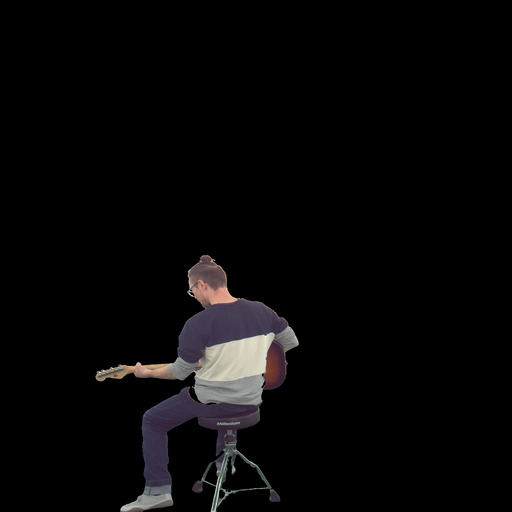} \\

\rotatebox{90}{\makecell{full model}} &
\includegraphics[width=\linewidth]{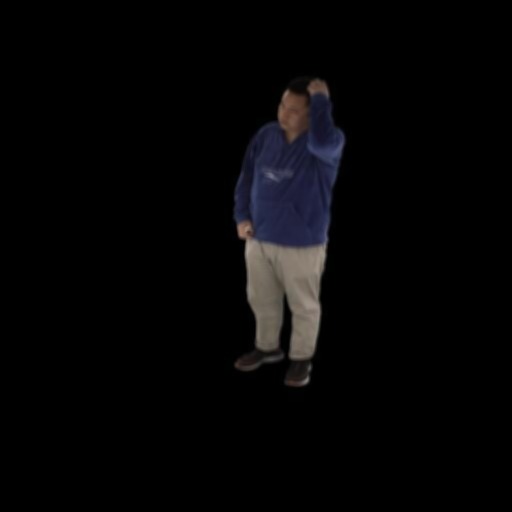} &
\includegraphics[width=\linewidth]{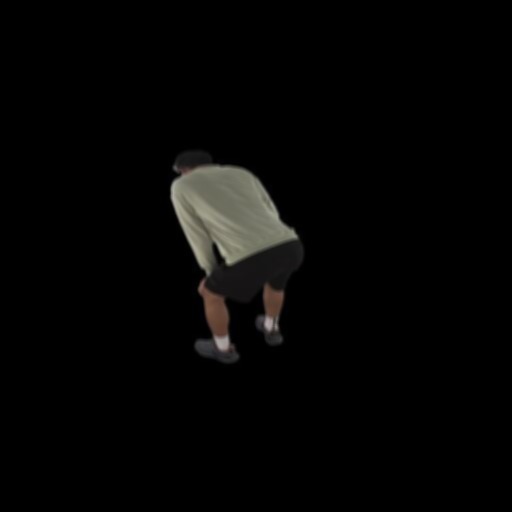} &
\includegraphics[width=\linewidth]{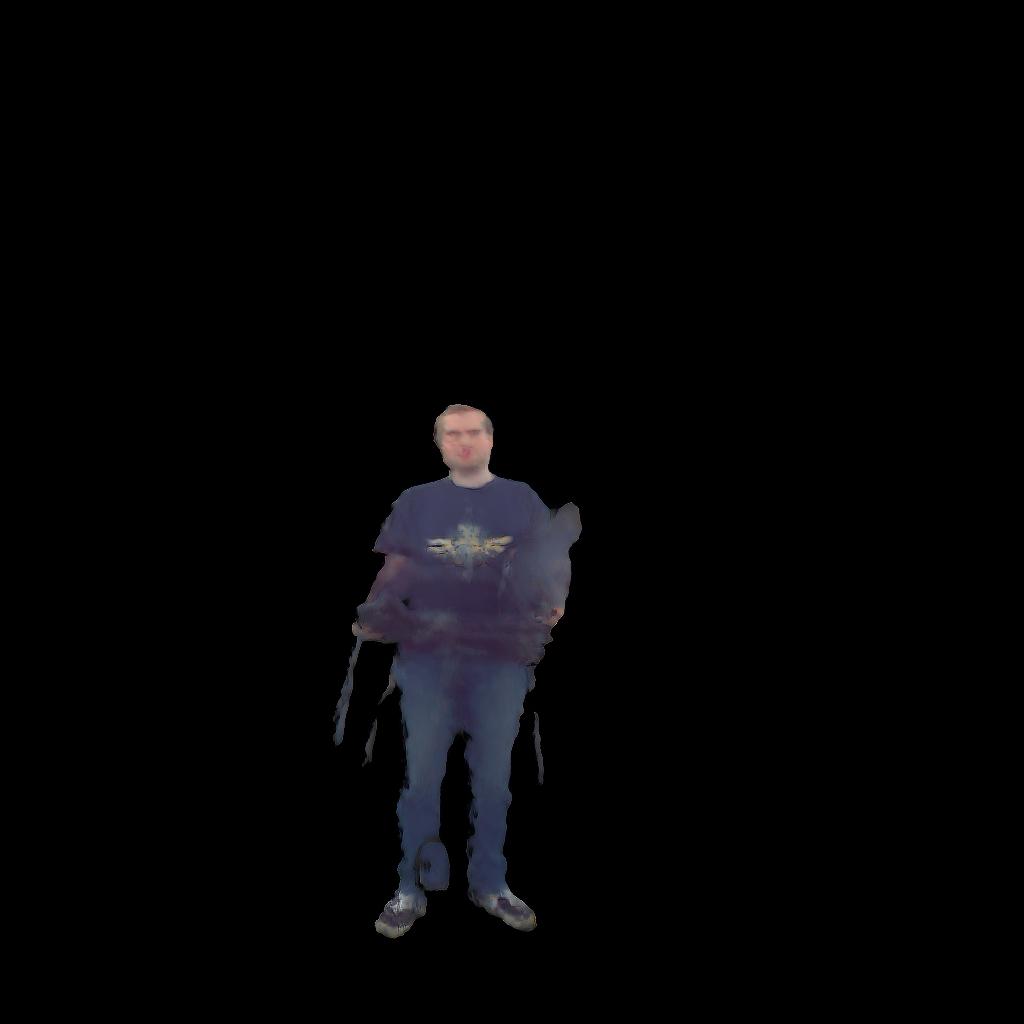} &
\includegraphics[width=\linewidth]{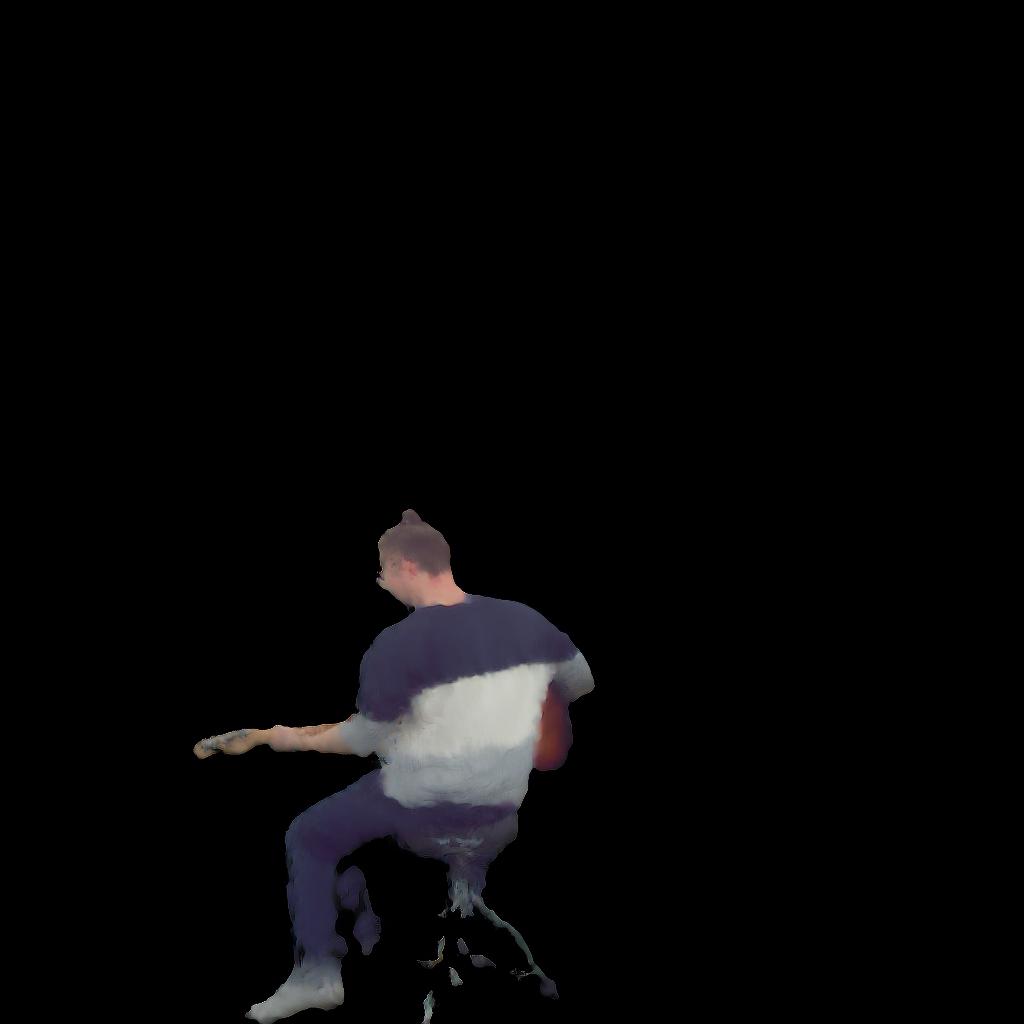} \\

\rotatebox{90}{\makecell{2 views}} &
\includegraphics[width=\linewidth]{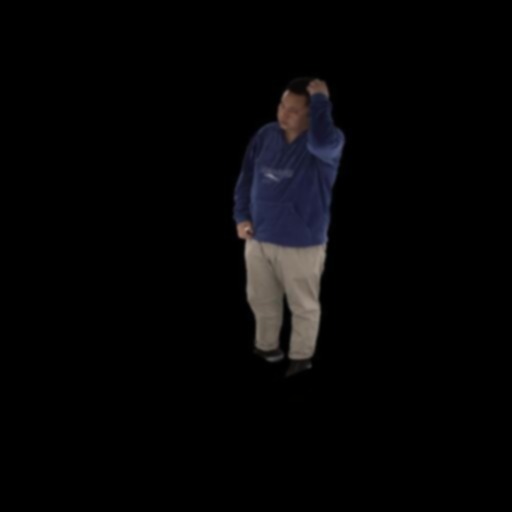} &
\includegraphics[width=\linewidth]{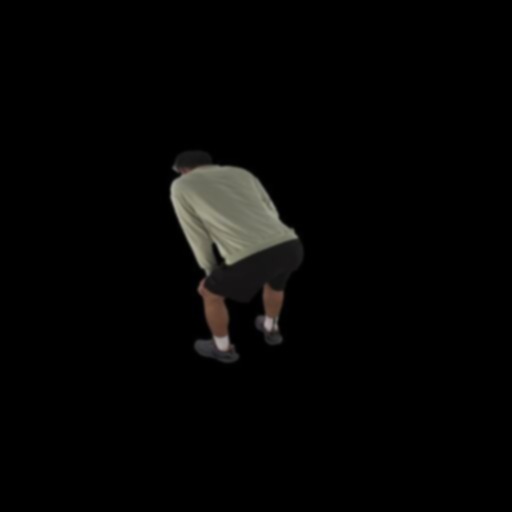} &
\includegraphics[width=\linewidth]{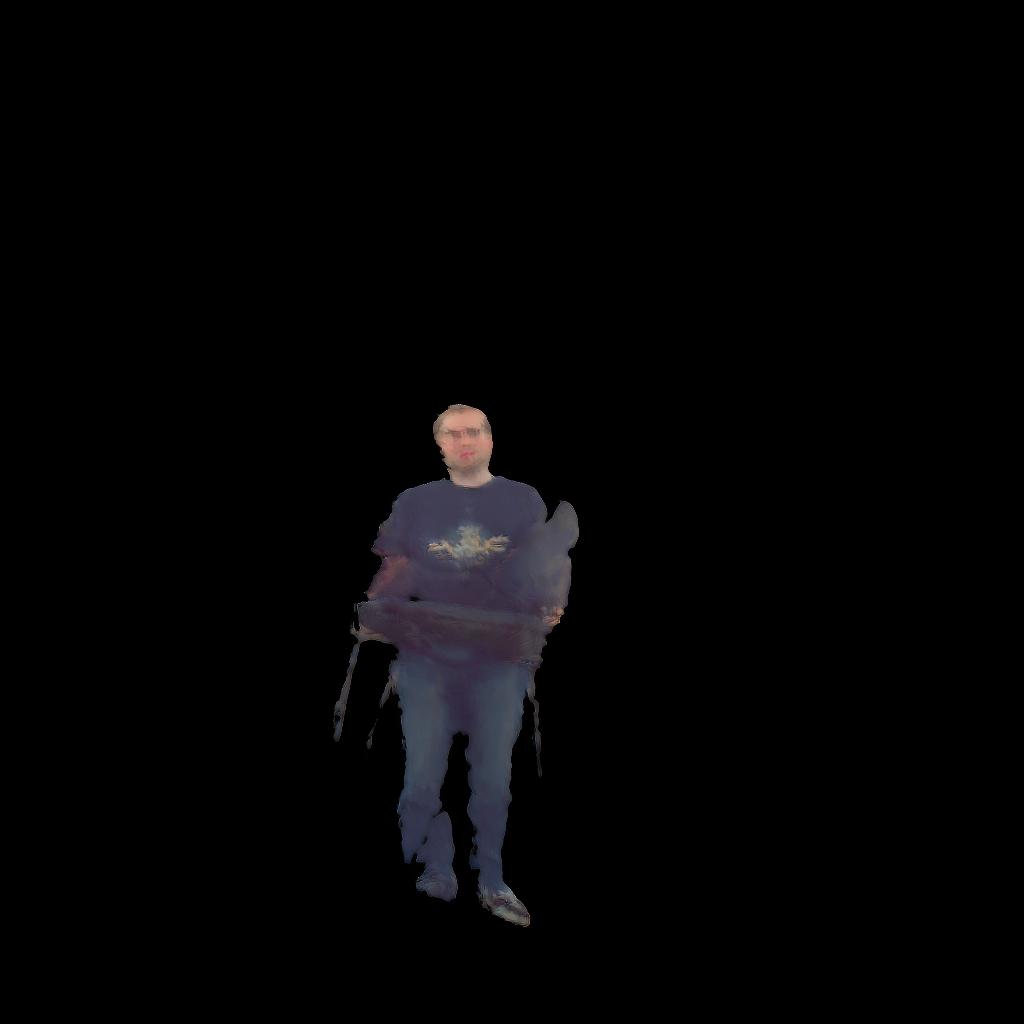} &
\includegraphics[width=\linewidth]{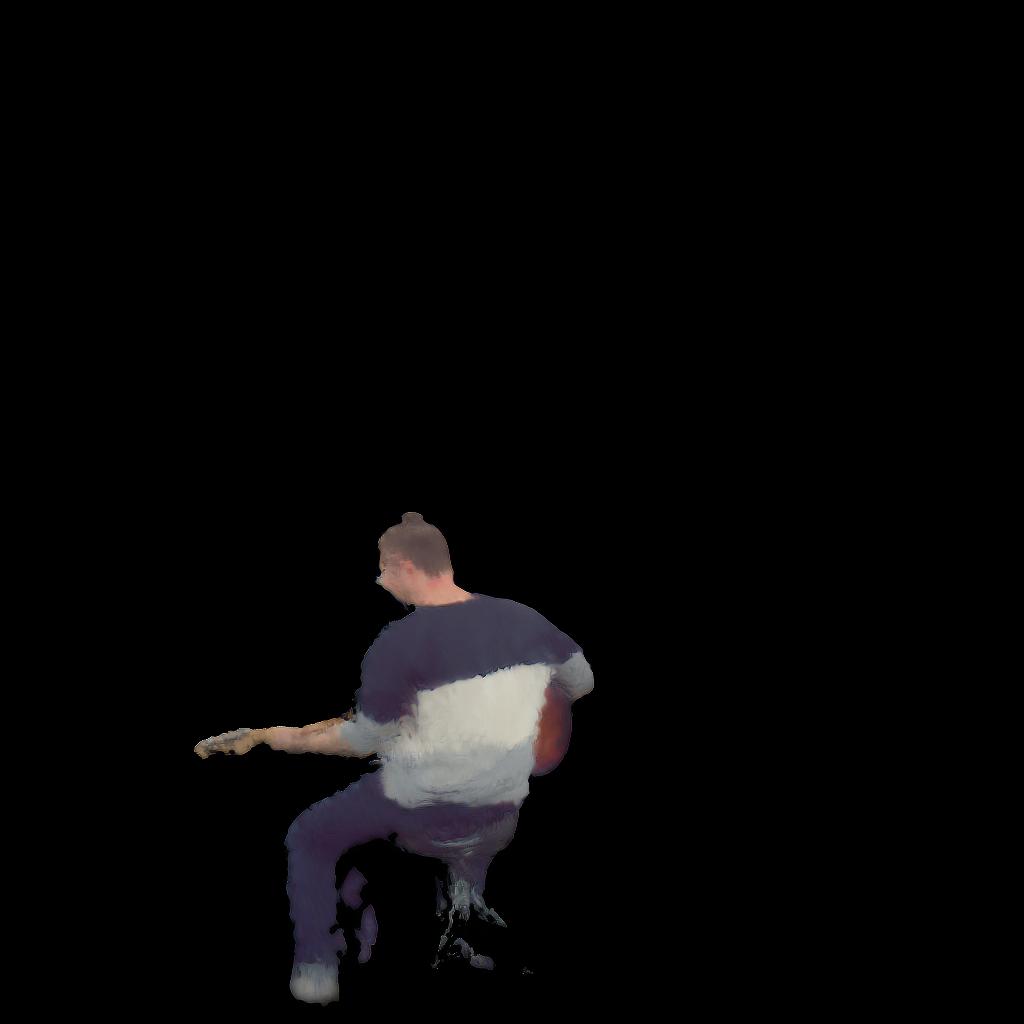} \\

\rotatebox{90}{\makecell{\# channels \\ halved}} &
\includegraphics[width=\linewidth]{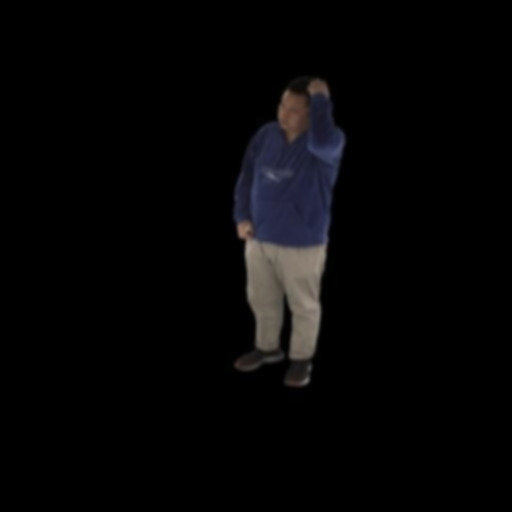} &
\includegraphics[width=\linewidth]{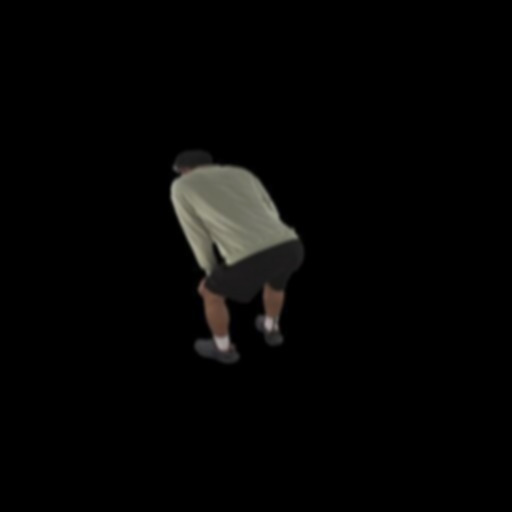} &
\includegraphics[width=\linewidth]{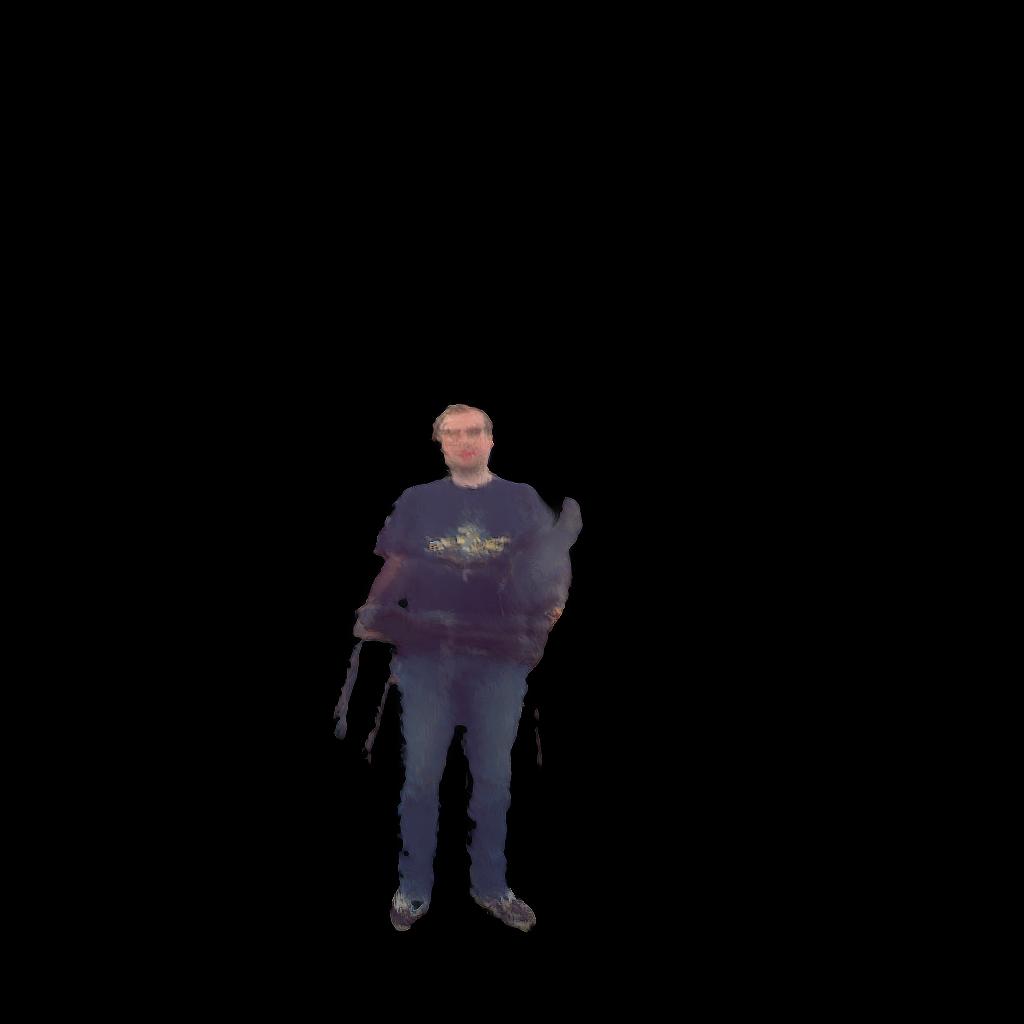} &
\includegraphics[width=\linewidth]{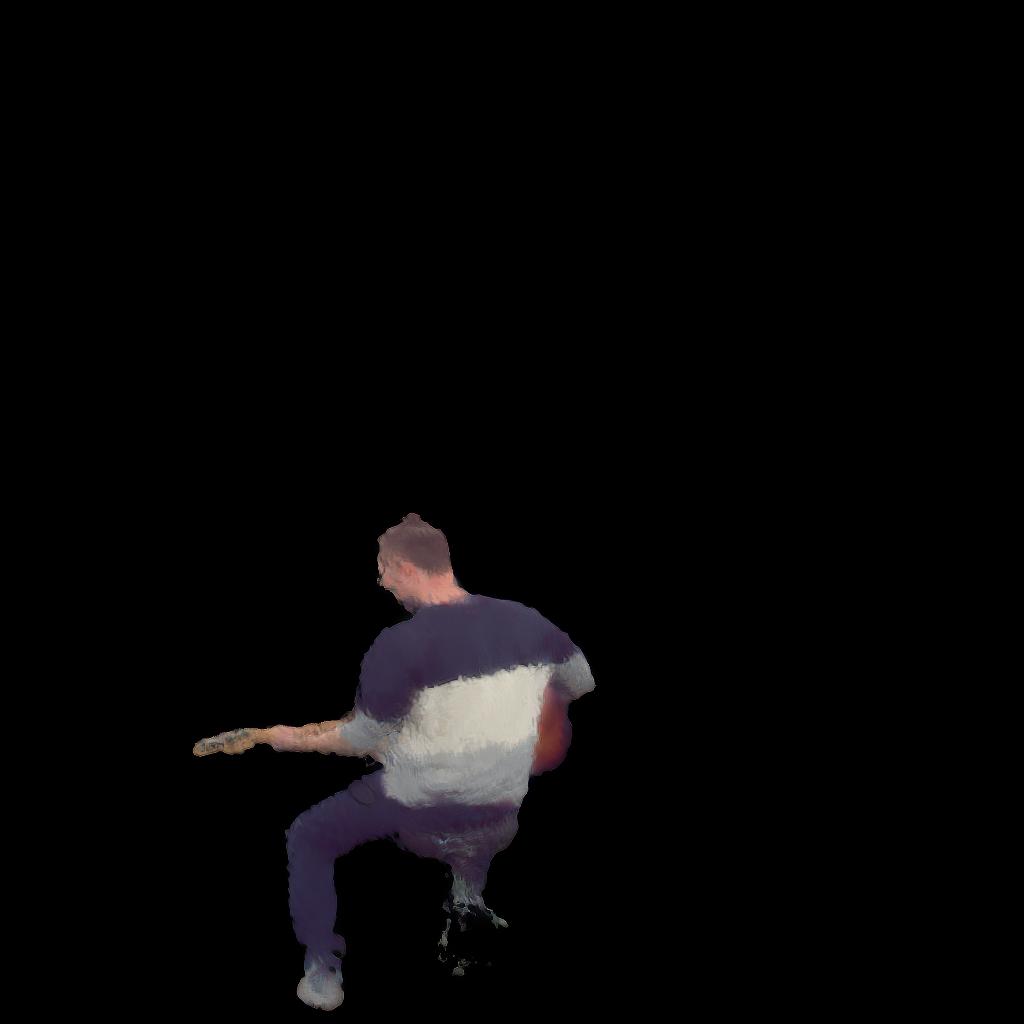} \\

\rotatebox{90}{\makecell{\# channels \\ doubled}} &
\includegraphics[width=\linewidth]{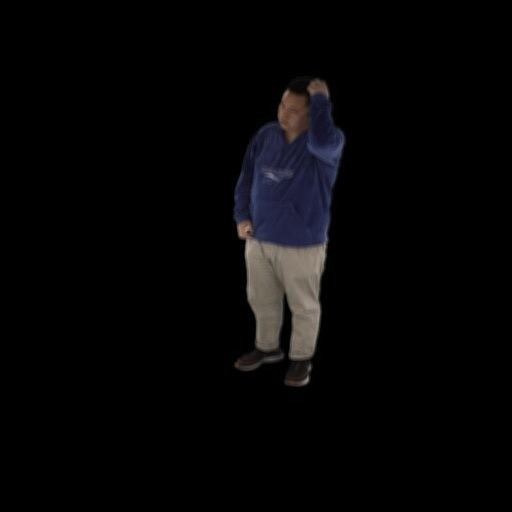} &
\includegraphics[width=\linewidth]{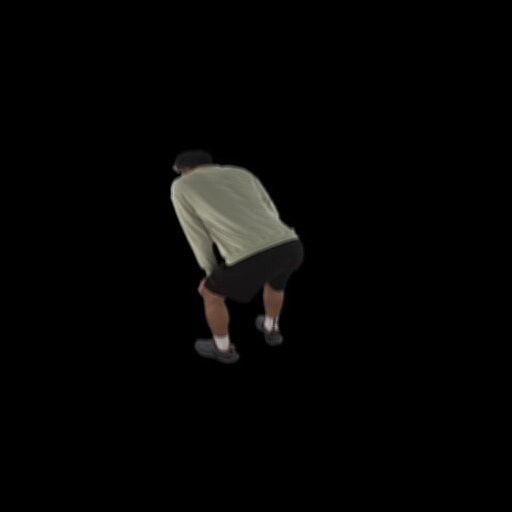} &
\includegraphics[width=\linewidth]{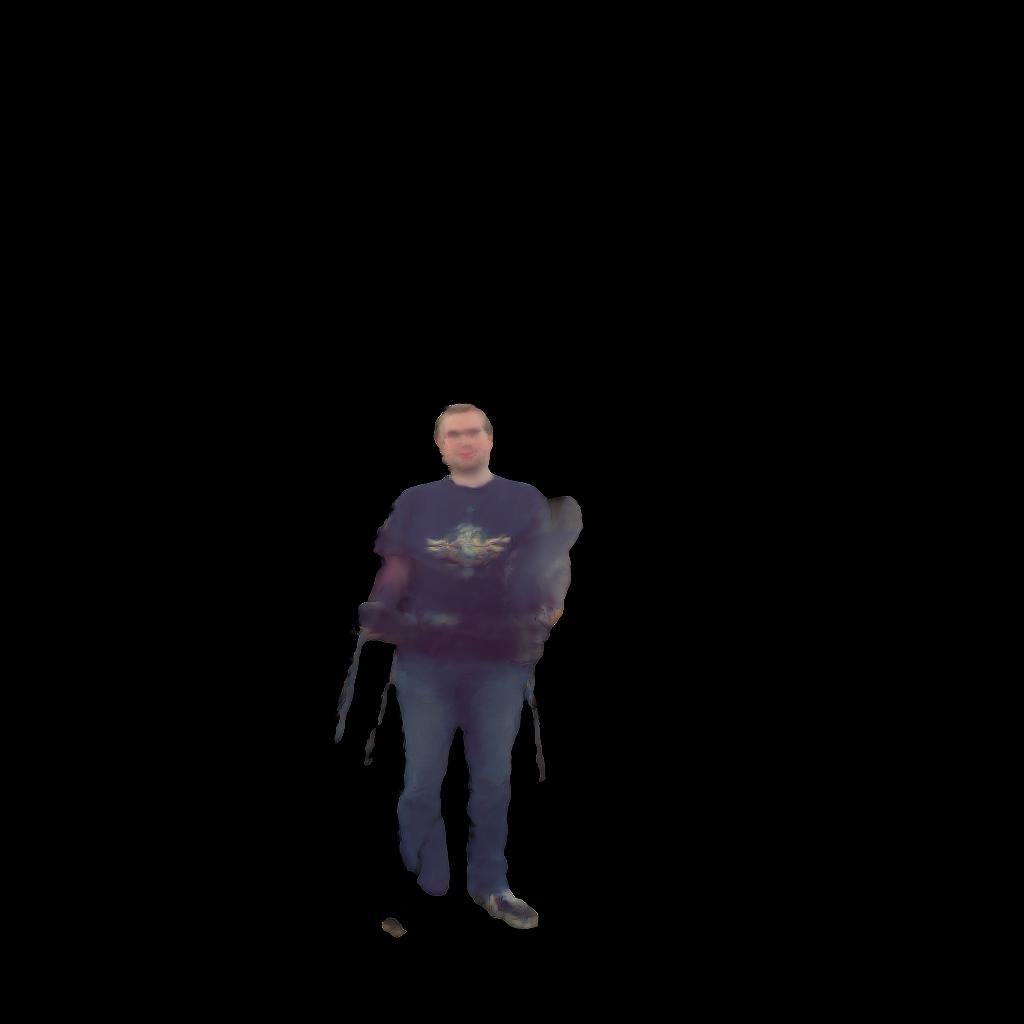} &
\includegraphics[width=\linewidth]{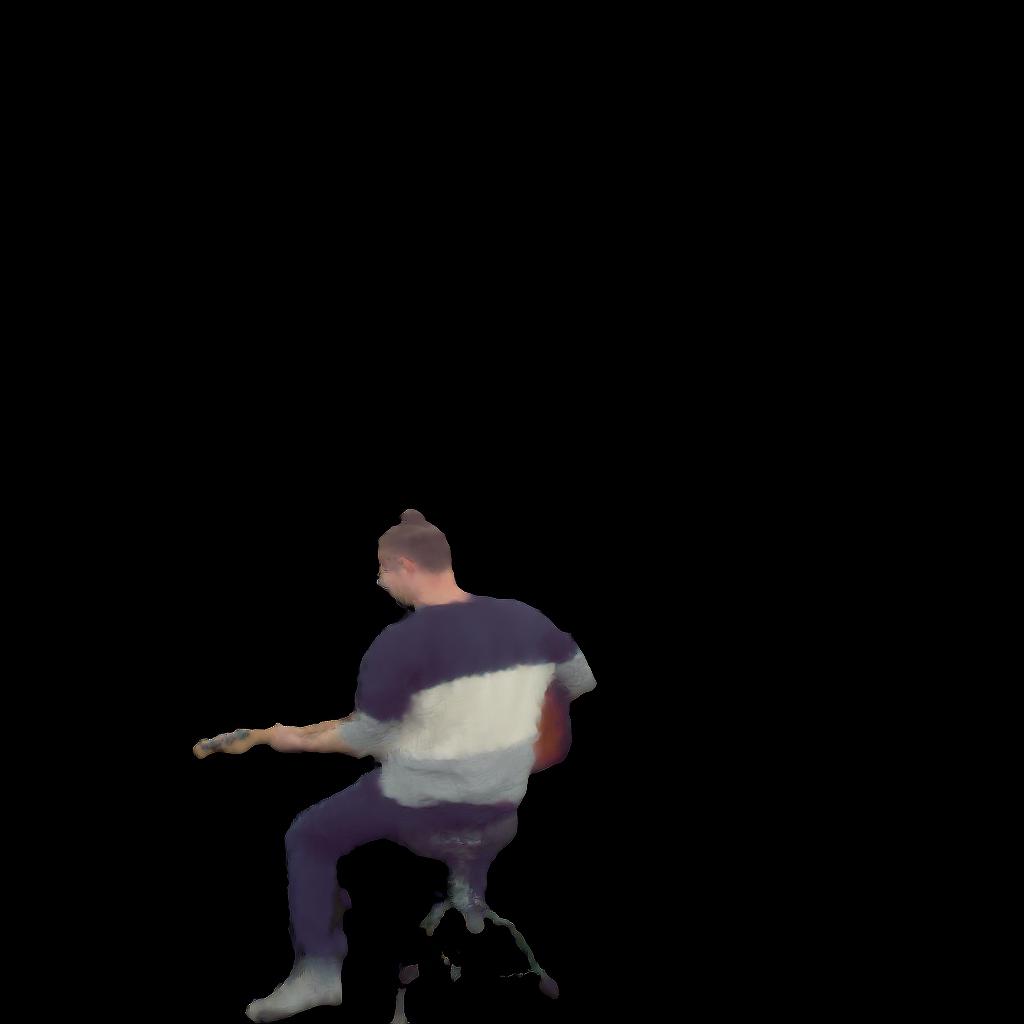} \\

\rotatebox{90}{\makecell{3-level \\ pyramid}} &
\includegraphics[width=\linewidth]{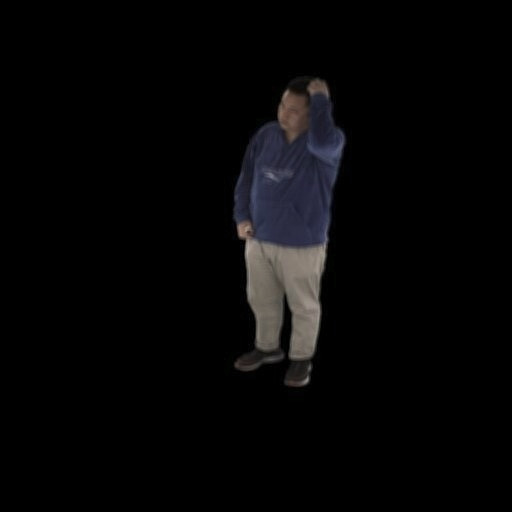} &
\includegraphics[width=\linewidth]{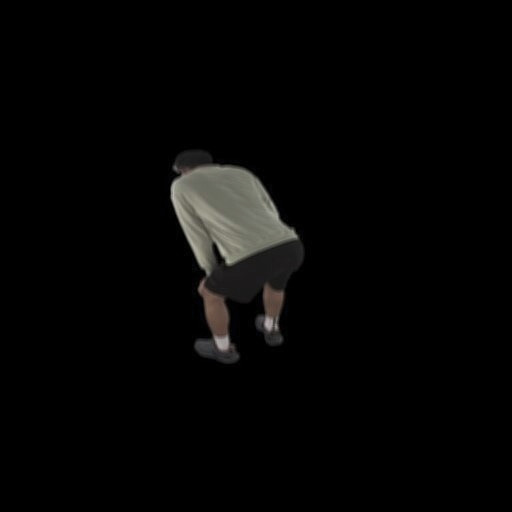} &
\includegraphics[width=\linewidth]{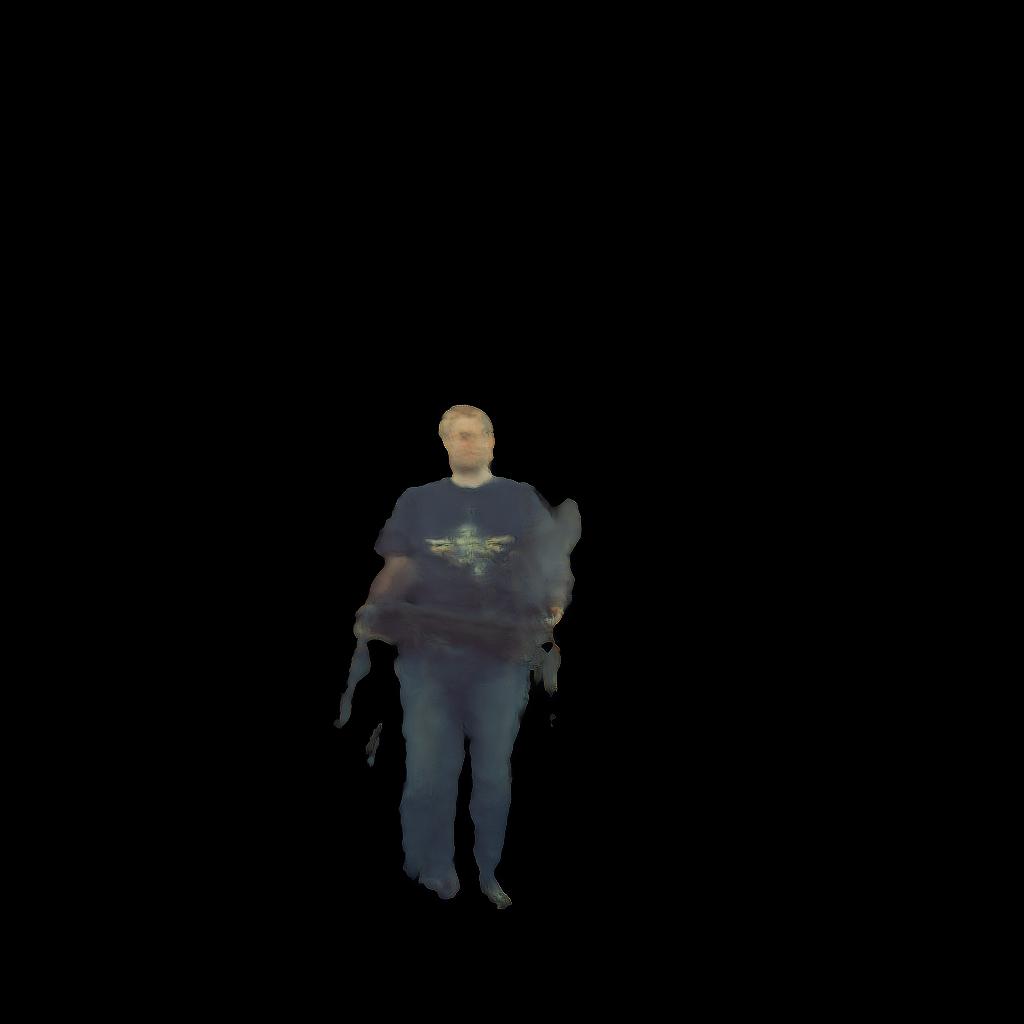} &
\includegraphics[width=\linewidth]{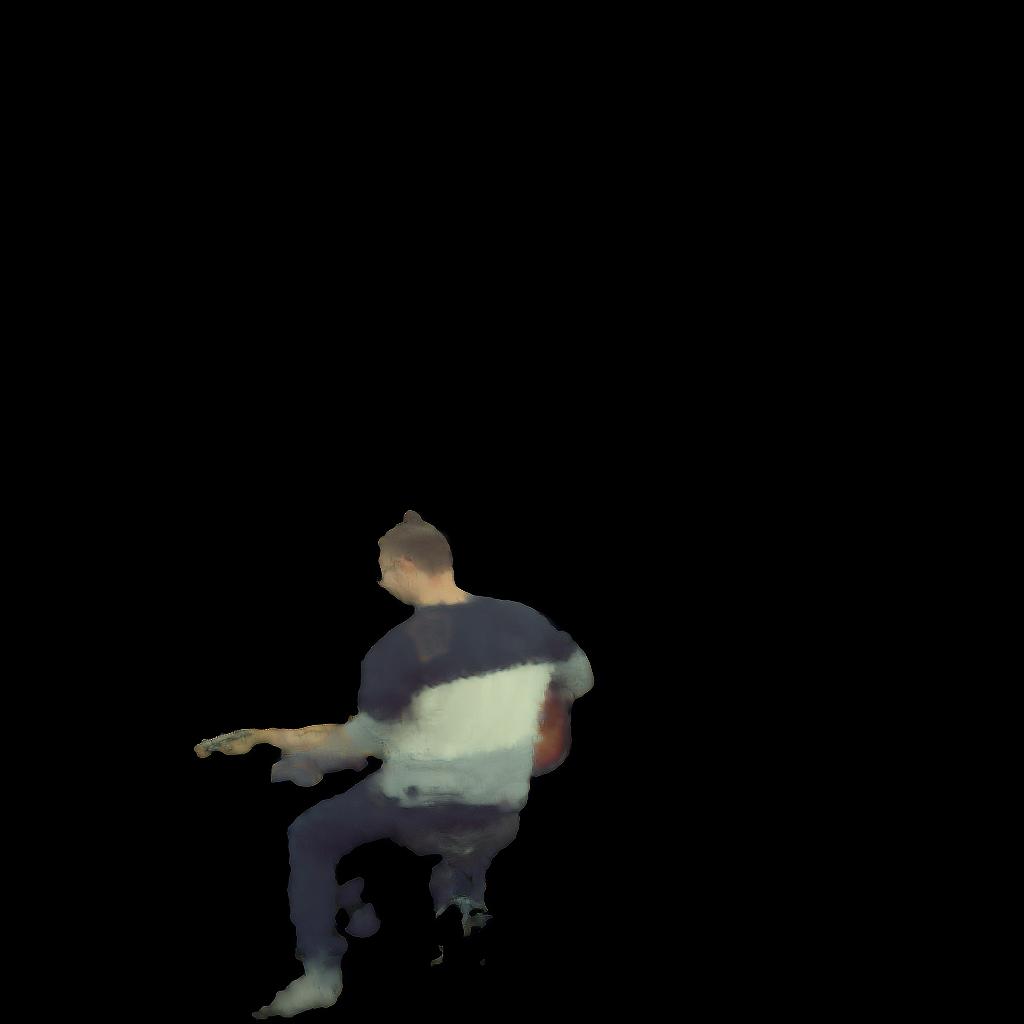} \\

\rotatebox{90}{\makecell{6-level \\ pyramid}} &
\includegraphics[width=\linewidth]{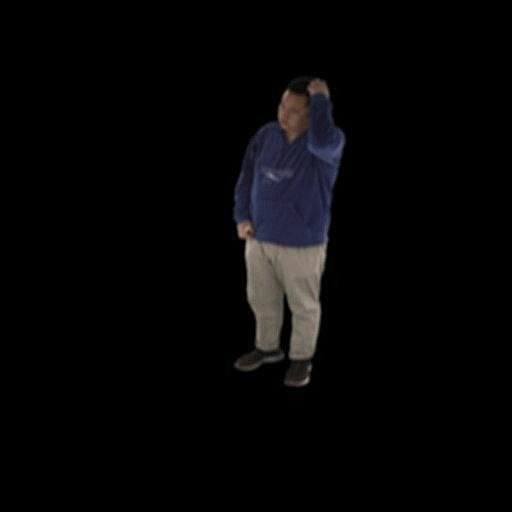} &
\includegraphics[width=\linewidth]{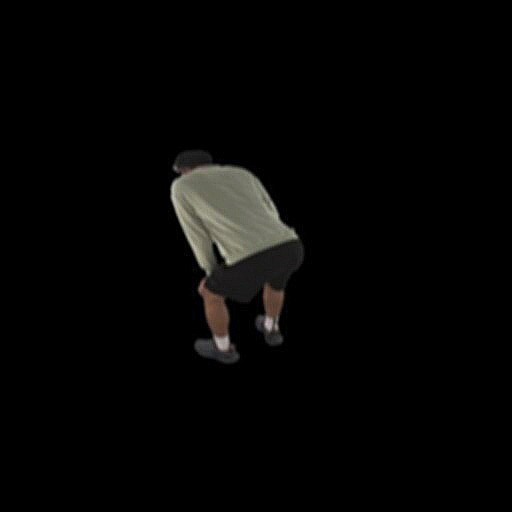} &
\includegraphics[width=\linewidth]{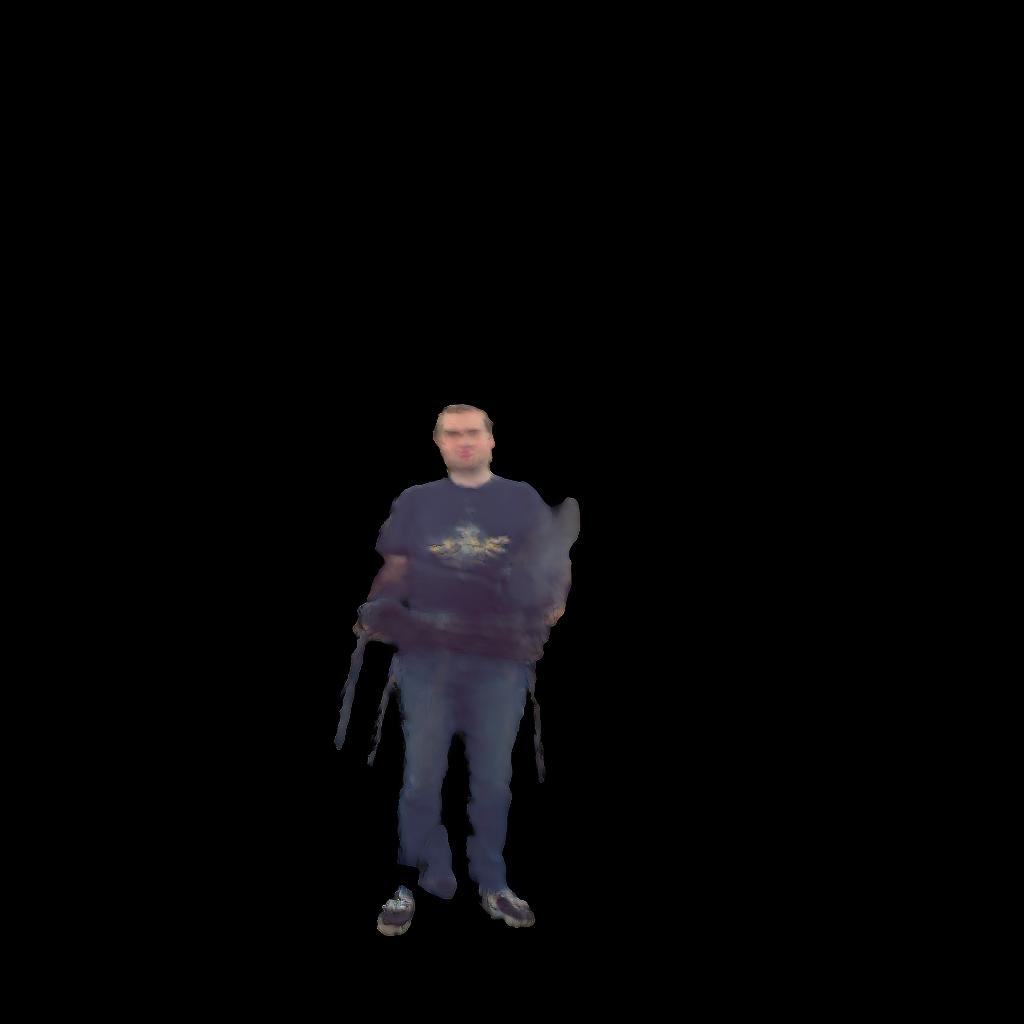} &
\includegraphics[width=\linewidth]{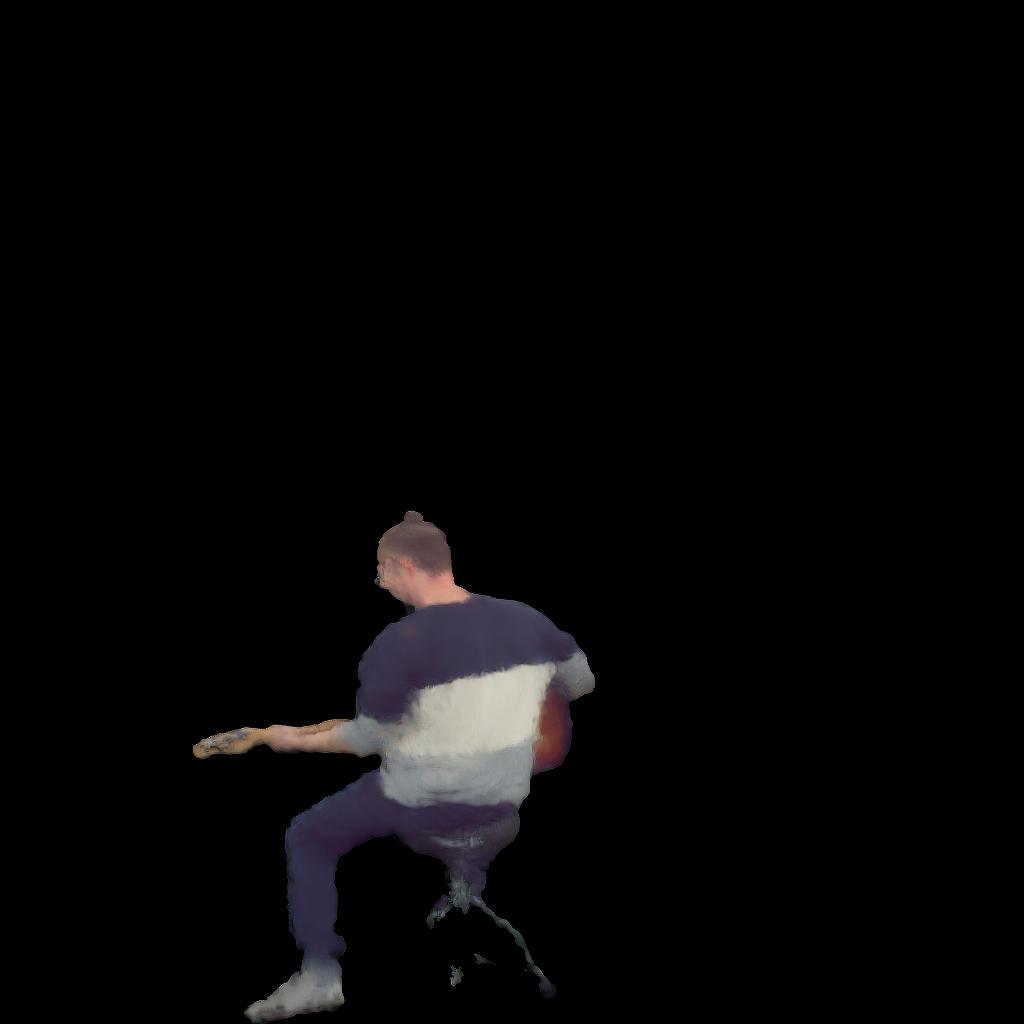} \\

\end{tabular}

\caption{Qualitative evaluation (Part 1) of our model and the different influences of individual parts to our overall network.}

\label{fig:visual-comparison-ablations-1}
\end{figure*}

\begin{figure*}[t]
\centering

\newcolumntype{M}[1]{>{\centering\arraybackslash}m{#1}}

\setlength{\tabcolsep}{1pt}
\renewcommand{\arraystretch}{0.5}
\begin{tabular}{M{0.08\linewidth} *{4}{M{0.185\linewidth}}}

& DNA 1 & DNA 2 & RIFTCast 1 & RIFTCast 2 \\

\rotatebox{90}{\makecell{GT}} &
\includegraphics[width=\linewidth]{images/6_visual_ours/ablation/gt-dna-1.jpg} &
\includegraphics[width=\linewidth]{images/6_visual_ours/ablation/gt-dna-2.jpg} &
\includegraphics[width=\linewidth]{images/6_visual_ours/ablation/gt-rift-1.jpg} &
\includegraphics[width=\linewidth]{images/6_visual_ours/ablation/gt-rift-2.jpg} \\

\rotatebox{90}{\makecell{w/o residual \\ depth}} &
\includegraphics[width=\linewidth]{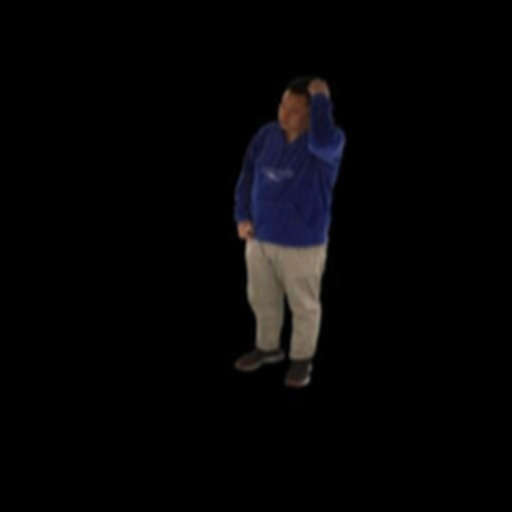} &
\includegraphics[width=\linewidth]{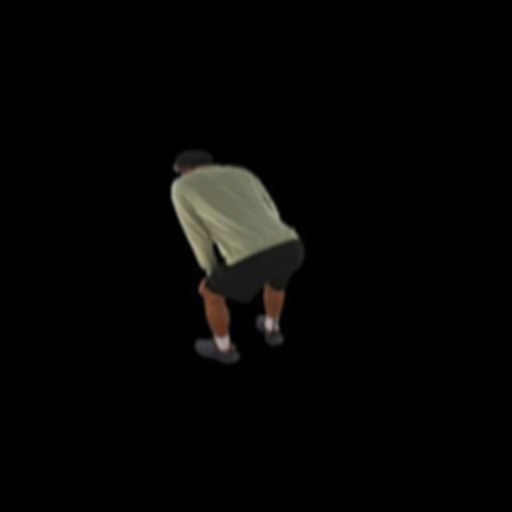} &
\includegraphics[width=\linewidth]{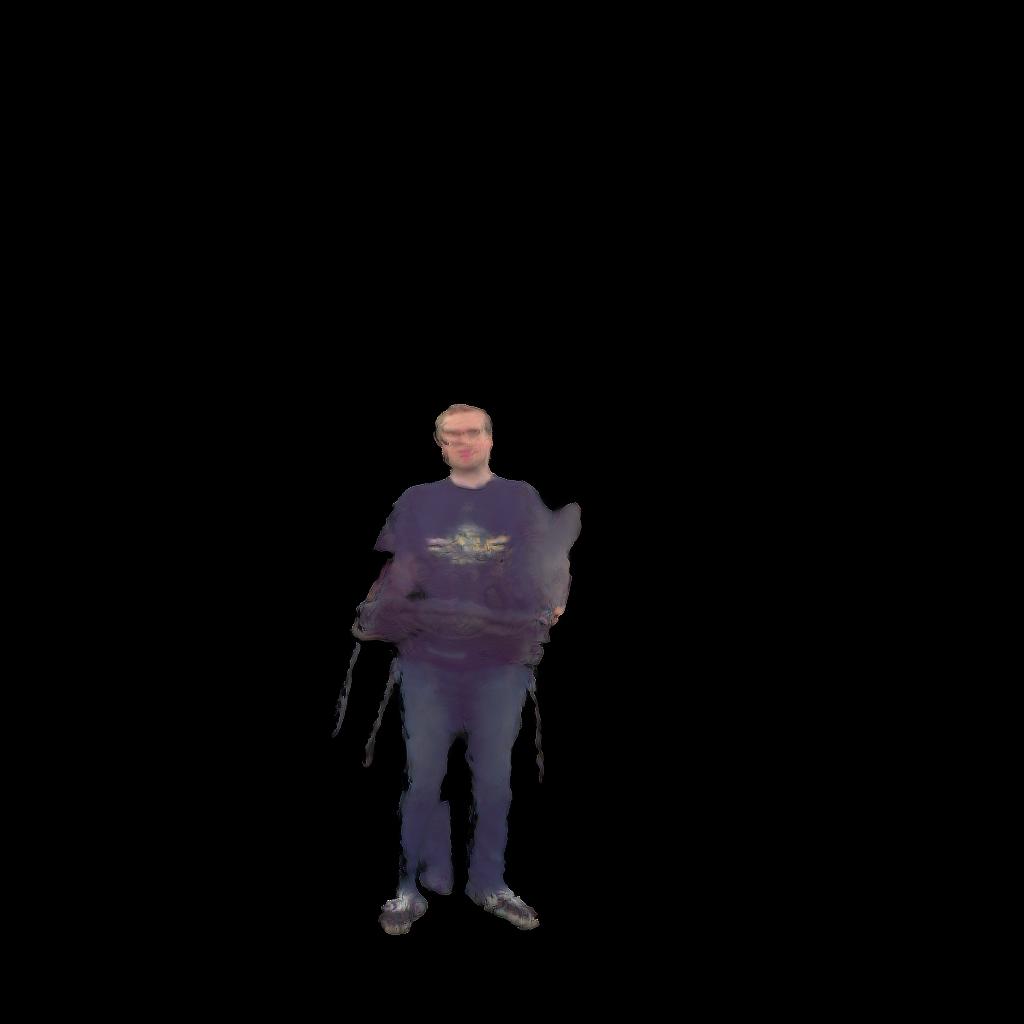} &
\includegraphics[width=\linewidth]{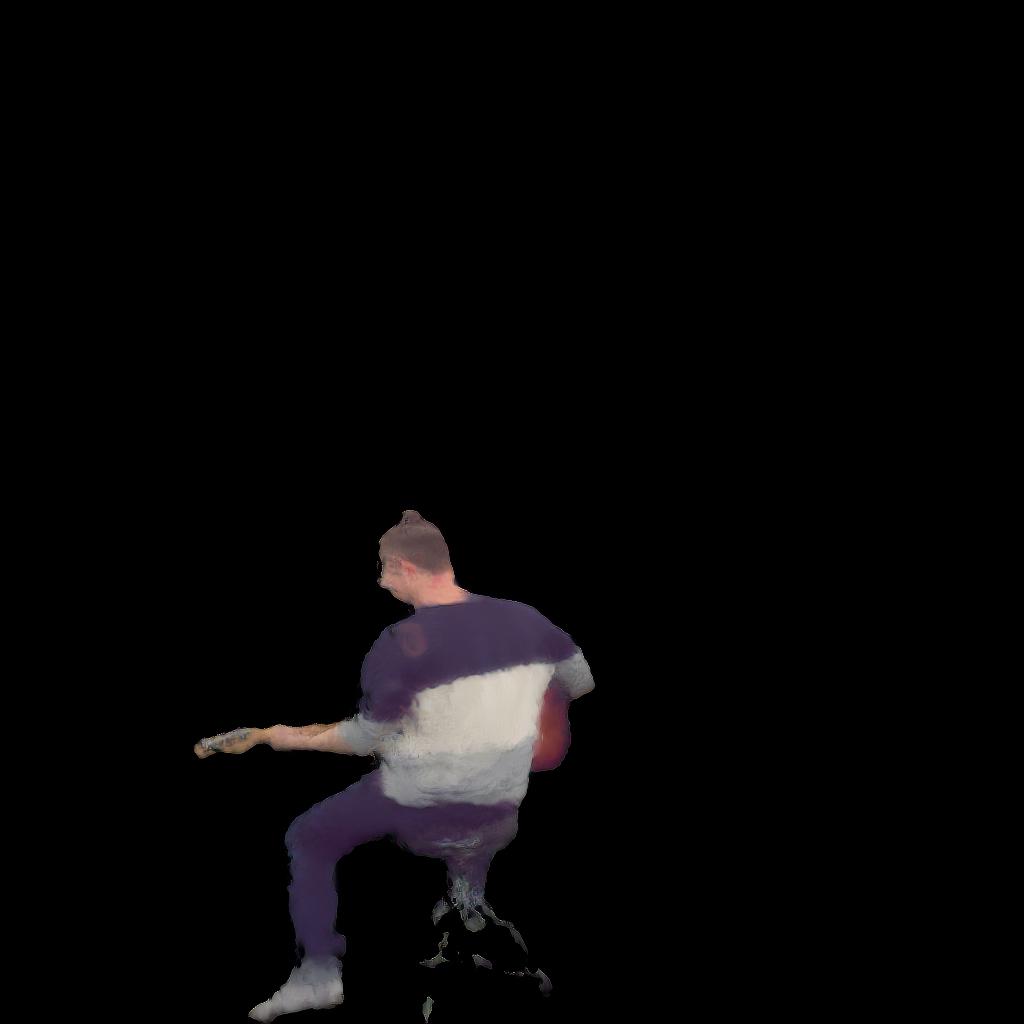} \\

\rotatebox{90}{\makecell{w/o residual \\ projector addition}} &
\includegraphics[width=\linewidth]{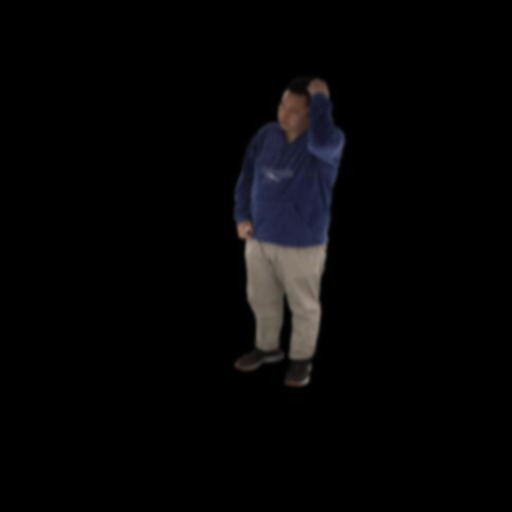} &
\includegraphics[width=\linewidth]{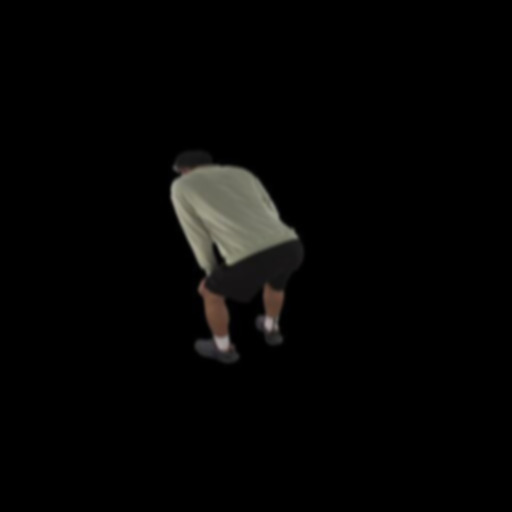} &
\includegraphics[width=\linewidth]{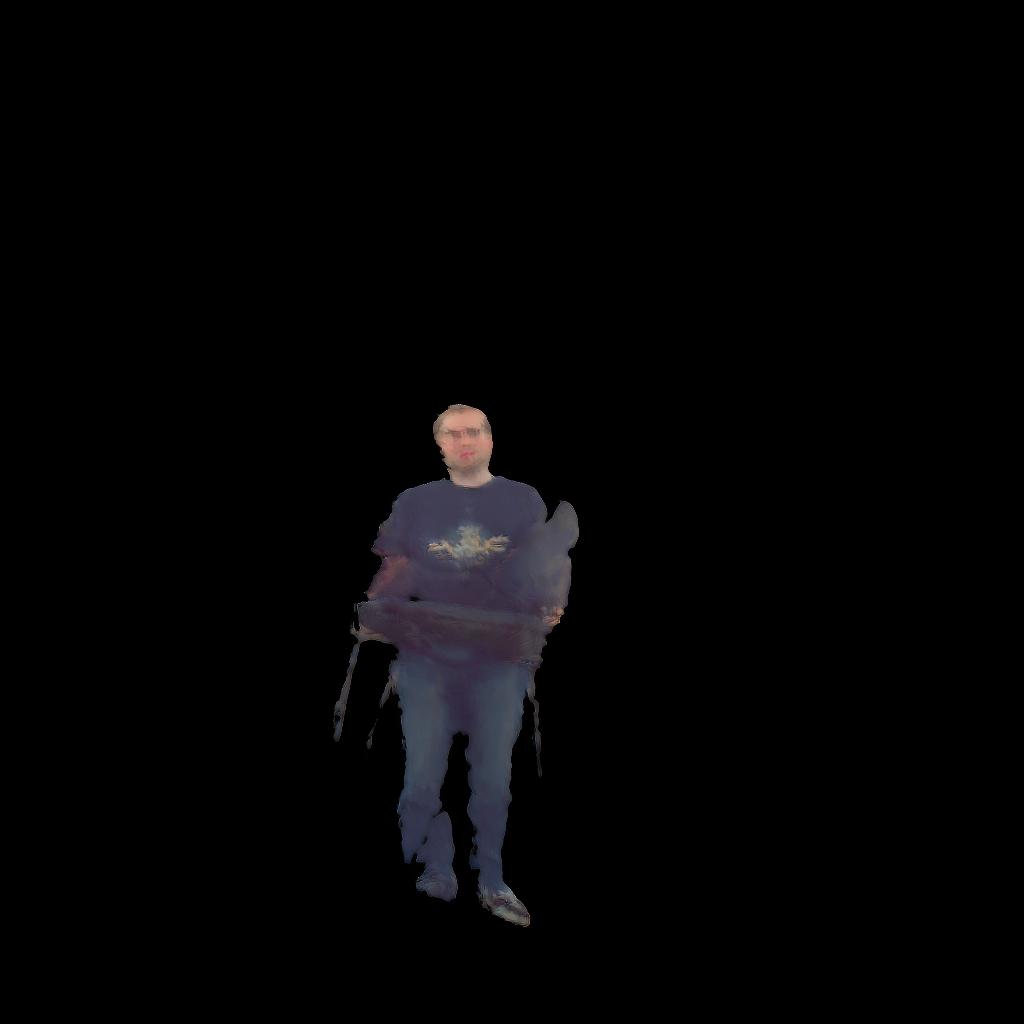} &
\includegraphics[width=\linewidth]{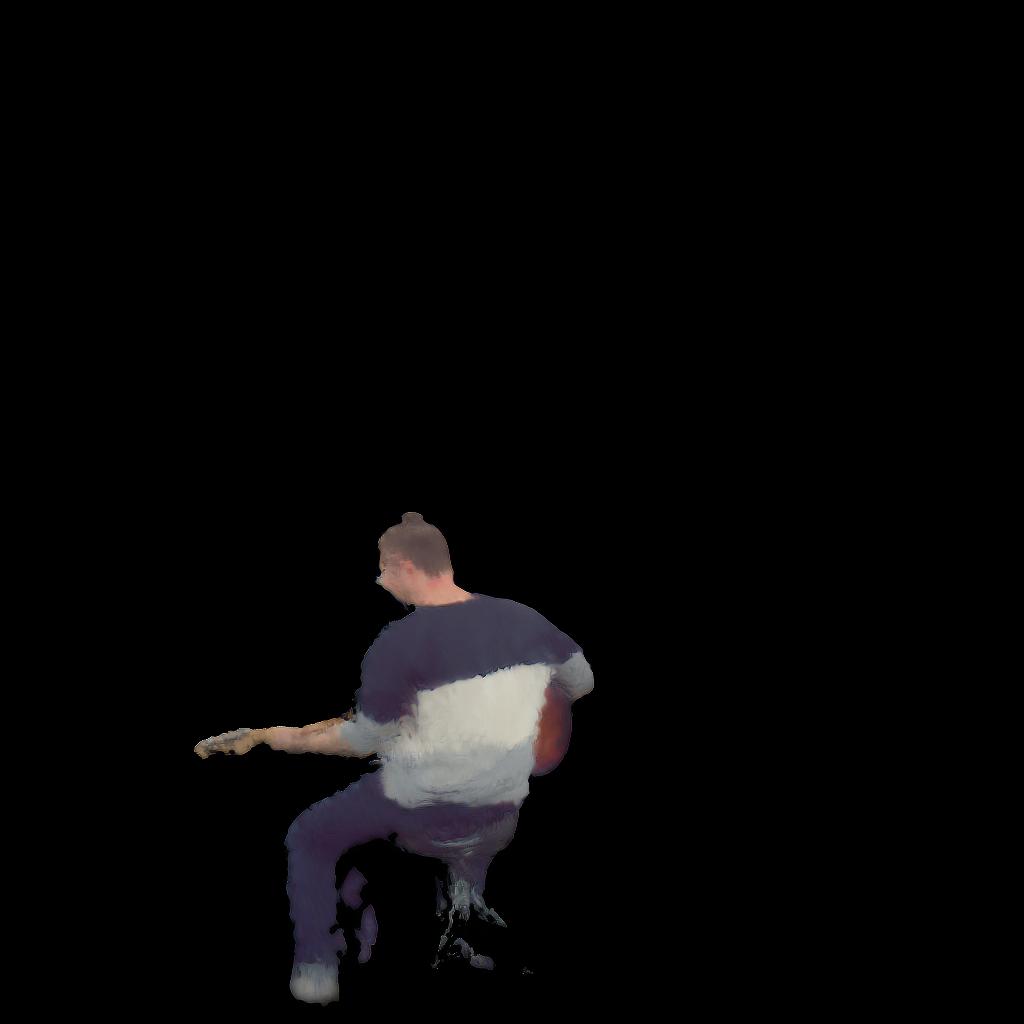} \\

\rotatebox{90}{\makecell{w/o prior}} &
\includegraphics[width=\linewidth]{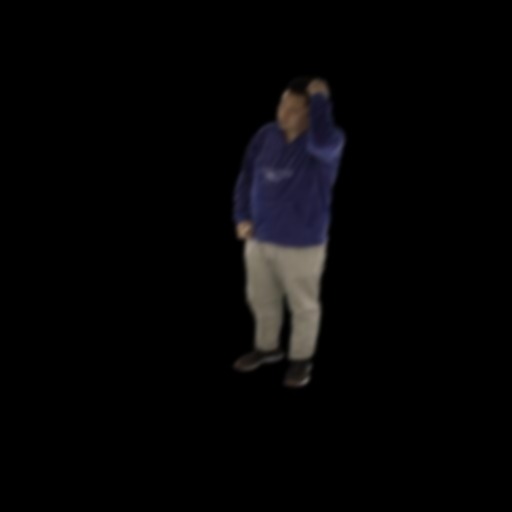} &
\includegraphics[width=\linewidth]{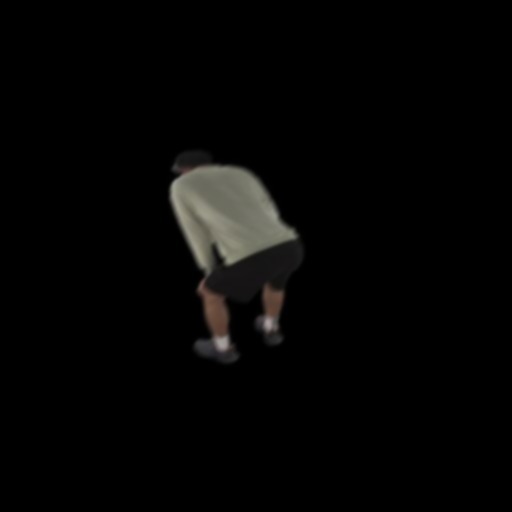} &
\includegraphics[width=\linewidth]{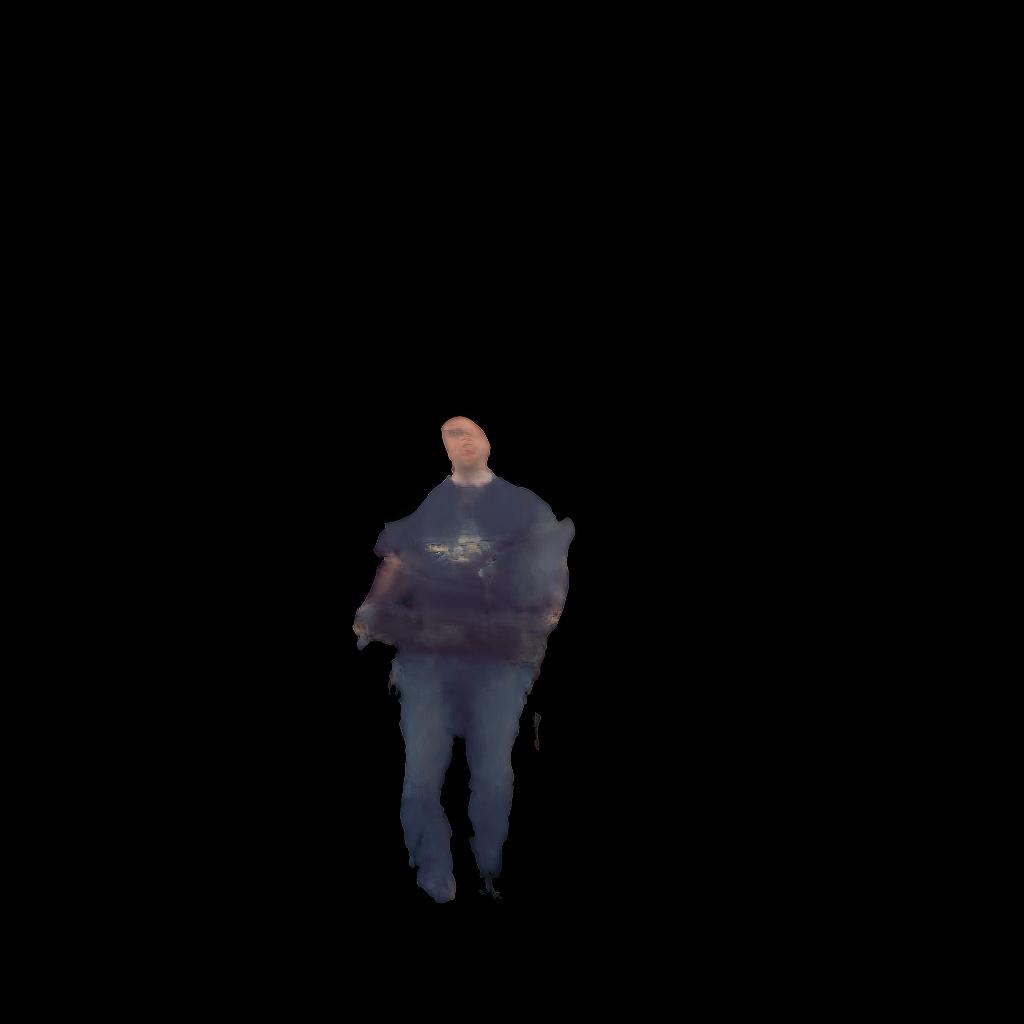} &
\includegraphics[width=\linewidth]{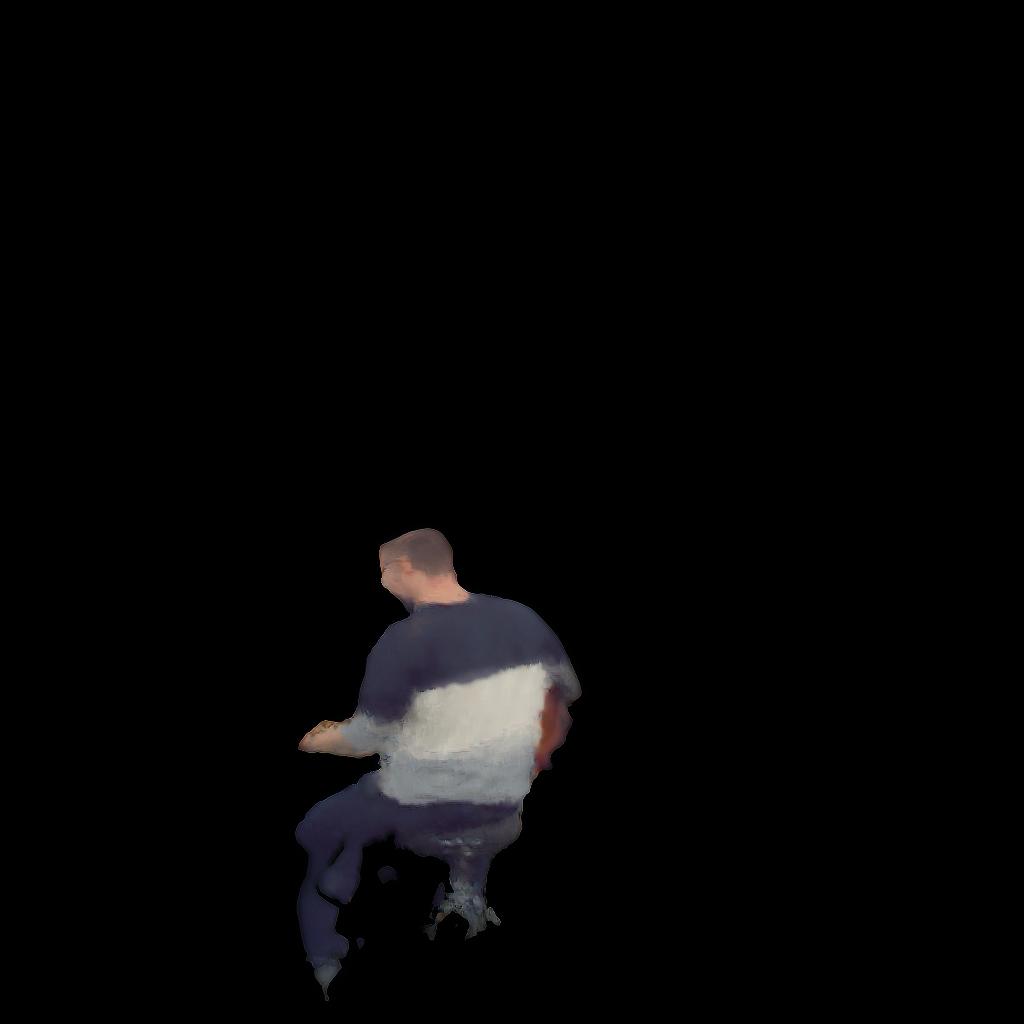} \\

\end{tabular}

\caption{Qualitative evaluation (Part 2) of our model and the different influences of individual parts to our overall network.}

\label{fig:visual-comparison-ablations-1}
\end{figure*}

%% file: suppl/6_tables.tex
\section{Additional Quantitative Evaluations}
\label{sec:quantitative}

In the main paper, we reported ablation studies on the DNA Rendering~\cite{cheng2023dnarenderingdiverseneuralactor} dataset, as it is a widely established benchmark in the community. For completeness, we provide the corresponding quantitative results for the RIFTCast~\cite{zingsheim2025riftcast} dataset in Table~\ref{tab:ablation-riftcast}. 

These results demonstrate consistent performance across different datasets, highlighting the individual contributions of our architectural modules. Notably, the exclusion of residual depth or prior-guided depth results in a significant drop in PSNR and LPIPS, underscoring their importance for maintaining high-fidelity results at a $1024 \times 1024$ resolution.

\begin{table*}[t]
  \centering
  \caption{\textbf{Quantitative Ablation on RIFTCast~\cite{zingsheim2025riftcast}.} Evaluation of model variants with TensorRT optimization at $1024 \times 1024$ resolution. The results highlight the performance trade-offs regarding channel depth, pyramid levels, and our proposed depth modules.}
  \begin{tabular}{@{}lrrr@{}}
    \toprule
    Method & PSNR $\uparrow$ & SSIM $\uparrow$ & LPIPS $\downarrow$ \\
    \midrule
    Full model (3 views) & 25.7 & 0.941 & 0.073 \\
    \specialrule{.01em}{0.1em}{0.15em} 
    2 views & 23.0 & 0.909 & 0.093 \\
    \specialrule{.01em}{0.1em}{0.15em}
    \# channels halved & 24.8 & 0.915 & 0.094 \\
    \# channels doubled & 26.1 & 0.950 & 0.073 \\
    \specialrule{.01em}{0.1em}{0.15em} 
    3-level pyramid & 22.4 & 0.888 & 0.118 \\
    6-level pyramid & 25.5 & 0.939 & 0.080 \\
    \specialrule{.01em}{0.1em}{0.15em} 
    w/o residual depth & 21.1 & 0.881 & 0.135 \\
    w/o residual projector addition & 25.1 & 0.940 & 0.076 \\
    w/o prior-guided depth & 23.4 & 0.933 & 0.087 \\
    \bottomrule
  \end{tabular}
  \label{tab:ablation-riftcast}
\end{table*}